%% file: main-frame.tex
\documentclass[sigconf,nonacm]{acmart}
\AtBeginDocument{
  \providecommand\BibTeX{{
    \normalfont B\kern-0.5em{\scshape i\kern-0.25em b}\kern-0.8em\TeX}}}

\usepackage[caption=false]{subfig}
\usepackage[nolist]{acronym}
\usepackage{listings}
\usepackage{array}
\usepackage{bm}
\usepackage{enumitem}
\usepackage[toc,page]{appendix}
\usepackage{amsmath}
\usepackage{float}
\usepackage{algpseudocode}
\usepackage[ruled]{algorithm2e}
\usepackage{tikz,tikz-qtree,tikz-qtree-compat}
\usepackage{caption}
\usepackage{pgfplots}
\usepackage{tabularx}
\usepackage{multirow}
\usepackage{booktabs}
\usepackage{setspace}
\usepackage{makecell}
\usepackage{dsbda}
\usepackage{adjustbox}

\pgfplotsset{compat=newest}

\newcolumntype{L}{>{\raggedright\arraybackslash}X}
\newcolumntype{V}[1]{>{\raggedright\arraybackslash}m{#1}}
\newcolumntype{C}{>{\centering\arraybackslash}X}
\newcolumntype{M}[1]{>{\centering\arraybackslash}m{#1}}

\newcommand\YAMLcolonstyle{\ttfamily\color{red}\mdseries}
\newcommand\YAMLkeystyle{\ttfamily\color{black}\bfseries\small}
\newcommand\YAMLvaluestyle{\ttfamily\color{blue}\mdseries}

\makeatletter
\def\NAT@spacechar{~}
\makeatother

\makeatletter

\newcommand\language@yaml{yaml}

\expandafter\expandafter\expandafter\lstdefinelanguage
\expandafter{\language@yaml}
{
  keywords={true,false,null,y,n},
  keywordstyle=\color{darkgray}\bfseries,
  basicstyle=\YAMLkeystyle,                                 
  sensitive=false,
  comment=[l]{\#},
  morecomment=[s]{/*}{*/},
  commentstyle=\color{purple}\ttfamily,
  stringstyle=\YAMLvaluestyle\ttfamily,
  moredelim=[l][\color{orange}]{\&},
  moredelim=[l][\color{magenta}]{*},
  moredelim=**[il][\YAMLcolonstyle{:}\YAMLvaluestyle]{:},   
  morestring=[b]',
  morestring=[b]",
  literate =    {---}{{\ProcessThreeDashes}}3
                {>}{{\textcolor{red}\textgreater}}1     
                {|}{{\textcolor{red}\textbar}}1 
                {\ -\ }{{\mdseries\ -\ }}3,
}

\lst@AddToHook{EveryLine}{\ifx\lst@language\language@yaml\YAMLkeystyle\fi}
\makeatother

\newcommand\ProcessThreeDashes{\llap{\color{cyan}\mdseries-{-}-}}

\acrodefplural{DGH}[DGHs]{Domain Generalization Hierarchies}

\begin{acronym}
\acro{ASR}[ASR]{Automatic Speech Recognition}
\acro{BiLSTM}[BiLSTM]{Bidirectional Long Short-Term Memory}
\acro{CRF}[CRF]{Conditional Random Fields}
\acro{CORD}[CORD-19]{COVID-19 Open Research Dataset}
\acro{DGH}[DGH]{Domain Generalization Hierarchy}
\acro{DM}[DM]{Discernibility Metric}
\acro{GAN}[GAN]{Generative Adversarial Network}
\acro{GCP}[GCP]{Global Certainty Penalty}
\acro{GDF}[GDF]{Global Document Frequency}
\acro{GDPR}[GDPR]{General Data Protection Regulation}
\acro{HIPAA}[HIPAA]{Health Insurance Portability and Accountability Act}
\acro{i2b2}[i2b2]{Informatics for Integrating Biology and the Bedside}
\acro{IC}[IC]{Information Content}
\acro{LSTM}[LSTM]{Long-Short Term Memory}
\acro{MIST}[MIST]{MITRE Identification Scrubber Toolkit}
\acro{NCP}[NCP]{Normalized Certainty Penalty}
\acro{NER}[NER]{Named Entity Recognition}
\acro{NIST}[NIST]{National Institute of Standards and Technology}
\acro{NLP}[NLP]{Natural Language Processing}
\acro{OLA}[OLA]{Optimal Lattice Anonymization}
\acro{PHI}[PHI]{Protected Health Information}
\acro{PII}[PII]{Personally Identifiable Information}
\acro{PPDP}[PPDP]{Privacy-Preserving Data Publishing}
\acro{POS}[POS]{Part-of-Speech}
\acro{RNN}[RNN]{Recurrent Neural Network}
\end{acronym}

\newlist{questions}{enumerate}{2}
\setlist[questions,1]{label=\textbf{RQ\arabic*.},ref=RQ\arabic*}
\setlist[questions,2]{label=(\alph*),ref=\thequestionsi(\alph*)}
\setlist[questions]{topsep=0pt}

\newcommand{\kanonymity}{$k$-anonymity}
\newcommand{\kmanonymity}{$k^m$-anonymity}
\newcommand{\ldiversity}{$l$-diversity}
\newcommand{\tcloseness}{$t$-closeness}

\newcommand{\edifprivacy}{$\varepsilon$-differential privacy}

\newcommand{\systemname}{rx-anon\xspace}

\copyrightyear{2021}
\acmYear{2021}
\setcopyright{acmlicensed}\acmConference[DocEng '21]{ACM Symposium on Document Engineering 2021}{August 24--27, 2021}{Limerick, Ireland}
\acmBooktitle{ACM Symposium on Document Engineering 2021 (DocEng '21), August 24--27, 2021, Limerick, Ireland}
\acmPrice{15.00}
\acmDOI{10.1145/3469096.3469871}
\acmISBN{978-1-4503-8596-1/21/08}

\begin{document}

\shortorextended{
\title{A Novel Approach on the Joint De-Identification of Textual and Relational Data with a Modified Mondrian Algorithm}
}{\title{\systemname---A Novel Approach on the De-Identification of Heterogeneous Data based on a Modified Mondrian Algorithm}}

\author{F. Singhofer}
\email{fabian.singhofer@uni-ulm.de}
\affiliation{
  \institution{University of Ulm}
  \country{Germany}
}

\author{A. Garifullina, M. Kern}
\email{{aygul.garifullina, mathias.kern}@bt.com}
\affiliation{
  \institution{BT}
  \country{United Kingdom}
}

\author{A. Scherp}
\email{ansgar.scherp@uni-ulm.de}
\affiliation{
  \institution{University of Ulm}
  \country{Germany}
}

\renewcommand{\shortauthors}{Singhofer, Garifullina, Kern, and Scherp}

\begin{abstract}
Traditional approaches for data anonymization consider relational data and textual data independently. We propose \systemname, an anony\-mi\-zation approach for heterogeneous semi-structured documents composed of relational and textual attributes. We map sensitive terms extracted from the text to the structured data. 
This allows us to use concepts like \kanonymity{} to generate a joint, privacy-preserved version of the heterogeneous data input. We introduce the concept of redundant sensitive information to consistently anonymize the heterogeneous data. To control the influence of anonymization over unstructured textual data versus structured data attributes, we introduce a modified, parameterized Mondrian algorithm. 
\extended{The parameter $\lambda$ allows to give different weights on the relational and textual attributes during the anonymization process.}
We evaluate our approach with two real-world datasets using a Normalized Certainty Penalty score, adapted to the problem of jointly anonymizing relational and textual data. The results show that our approach is capable of reducing information loss by using the tuning parameter to control the Mondrian partitioning while guaranteeing \kanonymity{}\extended{~for relational attributes as well as for sensitive terms}.
As \systemname is a framework approach, it can be reused and extended by other anonymization algorithms, privacy models, and textual similarity metrics.

\extended{
\textbf{Extended Technical Report}. To cite this work please refer to:
Fabian Singhofer, Aygul Garifullina, Mathias Kern, Ansgar Scherp:
\href{https://dblp.org/rec/conf/doceng/SinghoferGKS21.html}{\textit{A novel approach on the joint de-identification of textual and relational data with a modified Mondrian algorithm}}. DocEng 2021: 14:1-14:10.
\href{https://dblp.org/rec/conf/doceng/SinghoferGKS21.html?view=bibtex}{Access the BibTeX}.

}
\end{abstract} 

\ccsdesc[500]{Security and privacy~Data anonymization and sanitization}
\ccsdesc[100]{Computing methodologies~Information extraction}

\keywords{data~anonymization, heterogeneous~data, k-anonymity}

\maketitle

\section{Introduction}
\label{sec:introduction}

\begin{table*}[th]
    \centering
    \caption{Running example of a de-normalized dataset $\mathbf{D}$ with relational and textual attributes. 
    $\mathbf{A^{*}}$ is an attribute directly identifying an individual. $\mathbf{A_1,...,A_5}$ are considered quasi-identifiers and do not directly reveal an individual. 
    $\mathbf{X}$ is the textual attribute. 
    \extended{See Table~\ref{tab:notation} for details on notations.}
    }
    \begin{tabularx}{\linewidth}{c|c|c|c|c|c|L}
        \multicolumn{1}{c}{$A^*$}  & \multicolumn{5}{c}{Relational Attributes $A_1,...,A_5$} & \multicolumn{1}{c}{Textual Attribute $X$}\\
        \multicolumn{1}{c}{$\overbrace{\rule{0.5cm}{0pt}}$} & \multicolumn{5}{c}{$\overbrace{\rule{6.5cm}{0pt}}$} & \multicolumn{1}{c}{$\overbrace{\rule{8.9cm}{0pt}}$} \\
        \thead{id} & \thead{gender} & \thead{age} & \thead{topic} & \thead{sign} & \thead{date} & \thead{text} \\\hline
         1 &    male &   36 &  Education &   Aries & 2004-05-14 &  My name is Pedro, I’m a 36 years old engineer from Mexico. \\
 1 &    male &   36 &  Education &   Aries & 2004-05-15 &  A quick follow up: I will post updates about my education in more detail. \\
 2 &    male &   24 &    Student &     Leo & 2005-08-18 &  I will start working for a big tech company as an engineer. \\
 3 &    male &   37 &    Banking &  Pisces & 2004-05-27 &  During my last business trip to Canada I met my friend Ben from college. \\
 4 &  female &   24 &    Science &   Aries & 2004-01-13 &      As a scientist from the UK, you can be proud! \\
 4 &  female &   24 &    Science &   Aries & 2004-01-17 &  Four days ago, I started my blog. Stay tuned for more content. \\
 4 &  female &   24 &    Science &   Aries & 2004-01-19 &  2004 will be a great year for science and for my career as a biologist.\\
 5 &    male &   29 &     indUnk &  Pisces & 2004-05-15 &  Did you know that Pisces is the last constellation of the zodiac. \\
 6 &  female &   27 &    Science &   Aries & 2004-05-15 &  Rainy weather again here in the UK. I hope you all have a good day! \\
    \end{tabularx}
    \label{tab:example_init}
\end{table*}

Researchers benefit from companies, hospitals, or other research institutions, who share and publish their data.
It can be used for predictions, analytics, or visualizations.
However, often data to be shared contains \ac{PII} which does require measures in order to comply with privacy regulations\extended{~like the \ac{HIPAA} for medical records in the United States or the \ac{GDPR} in the European Union}. 
One possible measure to protect \ac{PII} is to anonymize all personal identifiers. Prior work considered such personal data to be name, age, email address, gender, sex, ZIP, any other identifying numbers, among others~\cite{ElEmam2009,Gardner2008,Sweeney2002,Liu2017,LeFevre2006}. 
\extended{Therefore, the field of \ac{PPDP} has been established which makes the assumption that a data recipient could be an attacker, who might also have additional knowledge (\eg by accessing public datasets or observing individuals).

Data to be shared can be structured in the form of relational data or unstructured like free texts.} 
Research in data mining and predictive models shows that a combination of structured and unstructured data leads to more valuable insights. 
\extended{One successful example involves data mining on COVID-19 datasets containing full texts of scientific literature and structured information about viral genomes.} 
\citet{Zhao2020} showed that linking the mining results can provide valuable answers to complex questions related to\extended{~genetics, tests, and prevention of} SARS-CoV-2. \extended{Moreover, the combination of structured and unstructured data can also be used to improve predictions of machine learning models.} 
\citet{Teinemaa2016} developed a model for predictive process monitoring that benefits from adding unstructured data to structured data. 
\extended{Therefore, links within heterogeneous data should be preserved, even if anonymized.}

However, state of the art methods focus either on anonymizing structured relational data~\cite{Sweeney2002k,Machanavajjhala2006,Li2007,Nergiz2007,Terrovitis2008,He2009} or anonymizing unstructured textual data~\cite{Sanchez2013,Sweeney1996,Ruch2000,Gardner2008,Dernoncourt2017,Liu2017}, but not jointly anonymizing on both.\extended{~In example, for structured data the work by \citet{Sweeney2002k} introduced the privacy concept \kanonymity{}, which provides a framework for categorizing attributes with respect to their risk of re-identification, attack models on structured data, as well as algorithms to optimize the anonymization process by reducing the information loss within the released version. 
For unstructured data like texts, high effort has been conducted to develop systems which can automatically recognize \ac{PII} within free texts using rule based approaches~\cite{Sweeney1996,Ruch2000,Neamatullah2008}, or machine learning methods~\cite{Gardner2008,Dernoncourt2017,Liu2017,Johnson2020} to allow for replacement in the next step.}
To the best of our knowledge, the only work that aimed to exploit synergies between anonymizing texts and structured data is by \citet{Gardner2008}.
The authors transferred textual attributes to structural attributes and subsequently applied a standard anonymization approach.
However, there is no recoding of the original text, \ie there is no transfer back of the anonymized sensitive terms.
Thus, essentially \citet{Gardner2008} only anonymize structured data.
Furthermore, there is no concept of information redundancy, which is needed for a joint de-anonymization, and there is no weighting parameter to control the influence of relational  versus textual attributes as splitting criterion for the data anonymization.
Our experiments show that such a weighting is important as otherwise it may lead to a skewed splitting favoring to retain relational attributes over textual attributes.

To illustrate the problem of a joint anonymization of textual and structured data, we consider an  example from a blog dataset~\cite{Schler2006}.\extended{\footnote{Note, we used the schema of the Blog Authorship Corpus throughout our running examples in Tables~\ref{tab:example_init}, \ref{tab:example_preprocessed}, and~\ref{tab:example_anonymized}.}}
As Table~\ref{tab:example_init} indicates, a combined analysis relies on links between the structured and unstructured data.
Therefore, it is important to generate a privacy-preserved, but also consistently anonymized release of heterogeneous datasets consisting of structured and unstructured data. 
Due to the nature of natural language, textual attributes might contain redundant information which is already available in a structured attribute. 
Anonymizing structured and unstructured parts individually neglects redundant information and leads to inconsistencies in data, since the same information might be anonymized differently. 
Moreover, for privacy-preserving releases, assumptions on the knowledge of an attacker are made. 
Privacy might be at risk if the anonymization tasks are conducted individually and without sharing all information about an individual.

We provide a formal problem definition and software framework \textit{\systemname} on a joint anonymization of relational data with free text fields. 
We experiment with two real-world datasets to demonstrate the benefits of the \systemname framework.
As a baseline, we consider a scenario where relational and textual attributes are anonymized alone\extended{, as it is done by the traditional approaches}. 
We show that we can reduce the information loss in texts under the \kanonymity{} model. 
Furthermore, we demonstrate the influence of the $\lambda$ parameter that influences the weight between relational and textual information and optimize the trade-off between relational and textual information loss.

In summary, our work makes the following contributions:
\begin{itemize}
    \item We formalize the problem of anonymizing\extended{~heterogeneous datasets composed of traditional} relational and textual attributes under the \kanonymity{} model and introduce the concept of redundant information.

    \item We present an anonymization framework based on Mondrian~\cite{LeFevre2006} with an adapted partitioning strategy and recoding scheme for sensitive terms in textual data.
    To this end, we introduce the tuning parameter $\lambda$ to control the share of information loss in relational and textual attributes\extended{~in Mondrian}.
    
    \item We evaluate our approach by measuring statistics on partitions and information loss on two real-world datasets.
    We adapt the Normalized Certainty Penalty score to the problem of a joint anonymization of relational and textual data.
\end{itemize}

Below, we discuss related work on data anonymization.
We provide a problem formalization in Section~\ref{sec:problem-statement} and introduce our joint de-anonymization approach \systemname in Section~\ref{sec:anonymization_approach}.
The experimental apparatus is described in Section~\ref{sec:experiments}.
We report our results in Section~\ref{sec:results}.
We discuss the results in Section~\ref{sec:discussion}, before we conclude.

\section{Related Work}
\label{sec:related_work}

\extended{Research in the field of anonymization can be differentiated according to the type of data to be anonymized.
We present related work on the de-identification of structured data\extended{, \ie traditional relational and transactional data,} and unstructured texts. 
We present works that aim to exploit synergies between these two tasks and contrast them with our\extended{~\systemname}  approach.}

\subsection{Anonymization of Structured Data}

Early work of \citet{Sweeney2000} showed that individuals\extended{, even if obvious identifiers are removed,} can be identified by using publicly available data sources\extended{~and link them to the apparently anonymized datasets}. 
Such attempts to reveal individuals using available linkable data are called record linkage attacks. 
Her work introduced explicit identifiers and quasi-identifiers. 
The former category is also called direct identifier and poses information which directly reveals an identity. 
Attributes of the latter category do not reveal an identity directly, but can\extended{~reveal an identity} if used in combination with other attributes. 

This observation led to extensive research on privacy frameworks. 
An important and very influential approach is \kanonymity{}, which prevents re-identification attacks relying on record linkage using additional data~\cite{Sweeney2002k}. 
\kanonymity{} describes a privacy model where records are grouped and each group is transformed such that their quasi-identifiers are equal. 
To achieve \kanonymity{}, \citet{Samarati2001} studied suppression and generalization as efficient techniques to enforce privacy.
\extended{In addition, \citet{Meyerson2004} and \citet{LeFevre2006} have shown that optimal \kanonymity{} in terms of information loss both in the suppression model and for the multidimensional case is $NP$-hard.} 
Several algorithms have been developed to efficiently compute a $k$-anonymous version of a dataset while keeping the information loss minimal. 
\citet{Sweeney2002} proposed a greedy approach with tuple suppression to achieve \kanonymity{}. 
\citet{LeFevre2006} suggested a top-down greedy algorithm Mondrian which implements multidimensional \kanonymity{} using local recoding models. 
\citet{Ghinita2007} showed how optimal multidimensional \kanonymity{} can be achieved by reducing the problem to a one-dimensional problem which improves performance while reducing information loss.
\extended{Based on the \kanonymity{} model, several extensions have been introduced and studied, where \ldiversity{} and \tcloseness{} are most popular.~}
\citet{Machanavajjhala2006} introduced the model of \ldiversity{} to prevent homogeneity and background knowledge attacks on the \kanonymity{} model. 
\ldiversity{} uses the concept of sensitive attributes to guarantee diversity of sensitive information within groups of records. 
\citet{Li2007} introduced \tcloseness{}, which extends the idea of diversity by guaranteeing that the distribution within groups does not differ more than a threshold $t$ from the global distribution of sensitive attributes.

\extended{While \kanonymity{} was initially designed to be applied for a single table containing personal data (also called microdata), it has been transferred to different settings.}
\citet{Nergiz2007} investigated the problem of anonymizing multi-relational datasets. 
\extended{They state that \kanonymity{} in its original form cannot prevent identity disclosure neither on the universal view nor on the local view and therefore modified \kanonymity{} to be applicable on multiple relations.}
\citet{Gong2017}\extended{~showed that regular \kanonymity{} fails on datasets containing multiple entries for one individual (also called 1:M). 
To anonymize such data, they} introduced $(k,l)$-diversity as a privacy model which is capable of anonymizing 1:M datasets.
\citet{Terrovitis2008} applied \kanonymity{} to transactional data.
Given a set of items within a transaction, they treated each item to be a quasi-identifier as well as a sensitive attribute simultaneously. 
\extended{The solution introduces \kmanonymity{} which adapts the original concept of \kanonymity{} and extends it by modeling the number of known items of the adversary in the transaction as $m$.}
\extended{\citet{He2009} proposed an alternative definition of \kanonymity{} for transactional data where instead of guaranteeing that subsets are equal in at least $k$ transactions, they require that at least $k$ transactions have to be equal.}
Finally, \citet{Poulis2013} showed how \kanonymity{} can be applied to data consisting of relational and transactional data and stated that a combined approach is necessary to ensure privacy.

\subsection{Anonymization of Unstructured Data}

\extended{In order for textual data to be anonymized, information in texts that may reveal individuals and therefore considered sensitive must be recognized.
In recent work, two approaches have been used to extract so called sensitive terms in text.
First,~}\citet{Sanchez2013} proposed an anonymization method which makes use of the \ac{IC} of terms. 
The \ac{IC} states the amount of information a term provides and can be calculated as the probability that a term appears in a corpus. 
The reasoning behind using the \ac{IC} of terms to detect sensitive information is that terms which provide high information tend to be also sensitive\extended{~in a sense that an attacker will gain high amounts of information if those terms are disclosed}. 
\extended{The advances in the field of \acf{NLP} have been used to detect sensitive terms by treating them as named entities. \ac{NER} describes the task of detecting entities within texts and assigning types to them. 
Named entities reflect instances or objects of the real world, like persons, locations, organizations, or products among others and provide a good foundation for detecting sensitive information in texts. 
Therefore, recent work formulated and solved the detection of sensitive information as a \ac{NER} problem~\cite{Johnson2020,Gardner2008,Liu2017,Trienes2020,Eder2019}.}
Early work on \ac{NER} to identify sensitive terms was based on rules and dictionaries~\cite{Sweeney1996,Ruch2000}.
\extended{\citet{Sweeney1996} suggested a rule-based approach using dictionaries with specialized knowledge of the medical domain to detect \ac{PHI}. 
\citet{Ruch2000} introduced a system for locating and removing \ac{PHI} within patient records using a semantic lexicon specialized for medical terms.}
\extended{Advances in machine learning led to new approaches on the de-identification of textual data.} 
\citet{Gardner2008} introduced an integrated system which uses \ac{CRF} to identify \ac{PII}. 
\citet{Dernoncourt2017} implemented a de-identification system with \acp{RNN} achieving high scores in the 2014 \ac{i2b2} challenge. 
\citet{Liu2017} proposed a hybrid automatic de-identification system which incorporates subsystems using rules as well as \acp{CRF} and \ac{BiLSTM} networks. 
\extended{They argued that a combined approach is preferable since entities such as phone numbers or email addresses can be detected using simple rules, while other entities such as names or organizations require trained models due to their diversity.}
\extended{Fundamental work on transformer neural networks established by \citet{Vaswani2017} raises the question, whether transformers can also lead to advances in anonymizing free texts.
\citet{Yan2019} suggested to use transformers for \ac{NER} tasks as an improvement to \ac{BiLSTM} networks. 
In addition, \citet{Khan2020} showed that transformer encoders can be used for multiple \ac{NLP} tasks and for specific domains such as the biomedical domain. Finally,} \citet{Johnson2020} were first to propose a de-identification system using transformer models~\cite{Vaswani2017}. 
Their results indicate that transformers are competitive to modern baseline models for anonymization of free texts. 

In addition to the detection of sensitive information using \ac{NER}, important related work is also on replacement strategies for such information in text. 
Simple strategies involve suppressing sensitive terms with case-sensitive placeholders~\cite{Ruch2000} or with their types~\cite{Neamatullah2008}. 
\extended{While those strategies are straightforward to implement, a disadvantage is loss of utility and semantics in the anonymized texts.}
More complex strategies use surrogates as consistent and grammatically acceptable replacements for sensitive terms~\cite{Trienes2020,Eder2019}. 
\extended{In contrast to the generation of surrogates,~}\citet{Sanchez2013} used generalization to transform sensitive terms to a more general version in order to reduce the loss of utility\extended{~while still hiding sensitive information}.

\subsection{Work Using Synergies Between Both Fields}

\extended{Anonymization of structured and unstructured data has mostly been considered in isolation.}
There were few works using synergies between both fields.
\citet{Chakaravarthy2008} brought a replacement technique for structured data to the field of unstructured texts. 
They used properties from \kanonymity{} to determine the sensitive terms to be anonymized within a single document by investigating their contexts.
Moreover, to the best of our knowledge, only \citet{Gardner2008} studied the task of anonymizing heterogeneous datasets consisting of texts and structured data. 
They provided a conceptual framework including details on data linking, sensitive information extraction, and anonymization.
However, their work has no concept of redundant information between structured and textual data, as we introduce in \systemname.
Furthermore, they have no weighting parameter to balance anonymization based on structural versus textual data like we do.
Basically, \citet{Gardner2008} transfer the problem of text anonymization to the structured world and then their approach ``forgets'' about where the attributes came from.
They do not transfer back the anonymized sensitive terms to recode the original text.
So the output of \citet{Gardner2008}'s anonymization approach is just structured data, which lacks its original heterogeneous form.

\shortorextended{
\section{Problem Formalization}
}{
\section{Problem Statement}
}
\label{sec:problem-statement}

\extended{We propose a method on automatically anonymizing datasets which are composed of relational attributes and textual attributes. 
Our approach is unsupervised and applicable across different domains. 
In order to achieve this task, we need to explain the process of anonymization, formalize the problem of anonymizing heterogeneous data, and describe our anonymization algorithm which is based on \kanonymity{}.

\subsection{Anonymization Process and Data Attributes}

For anonymizing a given dataset, multiple steps are necessary to provide a privacy-preserved release. 
We refer to release as the anonymized version of a given dataset, but a release does not necessarily have to be made publicly available.

In general, the process of anonymization can be divided into three parts, namely preparation, anonymization, and verification~\cite{InformationandPrivacyCommissionerofOntario2016}. 
In the preparation phase, the intended audience is assessed, attributes with their types are named, risks of re-identification attacks are analyzed, and the amount of anonymization is calculated based on the results of the prior steps. The next step involves the anonymization itself, where a dataset and determined parameters are taken as an input, and an anonymized dataset depicts the output. 
Finally, the verification step requires to assess that the required level of anonymization has been achieved (\eg by removing all \acs{PII}) while remaining the utility of the anonymized dataset. 

Depending on the dataset to be anonymized, there exist several different attributes which need to be anonymized. 
We have analyzed the literature and categorize the attributes with respect to the scale of the data (\ie nominal, ordinal, ratio) and their cardinality of relation (\ie one-to-one, one-to-many, and many-to-many). 
While the scale is important to know how attributes can be manipulated in order to achieve anonymity, the cardinality of relation provides information how attributes and individuals relate to each other. 

Table~\ref{tab:attributes} contains a non-exhaustive list of attributes, which typically appear in datasets and are critical with respect to re-identification attacks. 
For the attributes listed in Table~\ref{tab:attributes} we use four scales, namely nominal, ordinal, interval, and ratio. 
However, interval and ratio can be grouped together as numerical for the anonymization task.
Moreover, the cardinality of a relation between an individual and the attribute is to be interpreted as follows: 
A one-to-many relation means that one individual can have multiple instances of an attribute (\eg multiple credit card numbers), whereas many-to-one depicts a scenario where many individuals have one property in common (\eg place of birth). 
One-to-one and one-to-many attributes directly point to an individual and therefore are considered direct identifiers and must be removed prior to releasing a dataset. 
However, many-to-one and many-to-many attributes do not reveal an individual directly and therefore are called quasi-identifiers and might remain in an anonymized form in the released version of the dataset.

Even though in Table~\ref{tab:attributes} we present one exclusive cardinality of relation for each attribute, there are always cases where the cardinality of relation depends on context of attributes or whole datasets. 
An example is home address, where we state that it is a one-to-one attribute. 
However, this only holds if only one person of a household appears in the dataset. 
If multiple persons of a household appear in a dataset, we would need to consider it many-to-one. Moreover, if one individual might appear twice with different addresses (\eg having two delivery addresses in a shop), it would be an one-to-many attribute.

\begin{table}[ht]
    \caption
    {Non-exhaustive list of attributes to anonymize with scale and cardinality of relations (sorted by cardinality). 
    Note, for the anonymization task the interval and ratio data can be grouped together as numerical.}
    \centering
    \begin{tabular}{l|c|c}
        \toprule
        \thead{Attribute} & \thead{Scale} & \thead{Cardinality of\\Relation}\\
        \midrule
        Name \citep{HIPAA,McCallister2010,GDPR,Poulis2013} & nominal & one-to-one\\
        Social Security Number \citep{HIPAA,McCallister2010} & nominal & one-to-one\\
        Online identifier \citep{GDPR} & nominal & one-to-one\\
        Passport Numbers \citep{McCallister2010,GDPR} & nominal & one-to-one\\
        Home Address \citep{HIPAA,McCallister2010,GDPR} & nominal & one-to-one\\
        Credit Card Number \citep{McCallister2010} & nominal & one-to-many\\
        Phone \citep{HIPAA,McCallister2010}& nominal & one-to-many\\
        Email Address \citep{HIPAA,McCallister2010,GDPR} & nominal & one-to-many\\
        License Plate Number \citep{HIPAA} & nominal & one-to-many\\
        IP Address \citep{HIPAA,McCallister2010,GDPR} & nominal & one-to-many\\
        Order Reference & nominal & one-to-many\\
        Age \citep{HIPAA,Poulis2013} & ratio & many-to-one\\
        Sex / Gender \citep{GDPR,Poulis2013} & nominal & many-to-one\\
        ZIP / Postcode \citep{HIPAA} & nominal & many-to-one\\
        Date of Birth \citep{HIPAA,McCallister2010} & interval & many-to-one\\
        Zodiac Sign \citep{Schler2006} & ordinal & many-to-one\\
        Weight \citep{McCallister2010,GDPR} & ratio & many-to-one\\
        Race \citep{McCallister2010,GDPR} & nominal & many-to-one\\
        Country & nominal & many-to-one\\
        City \citep{HIPAA} & nominal & many-to-one\\
        Salary Figures \citep{Fung2010} & ratio & many-to-one\\
        Religion \citep{McCallister2010,GDPR} & nominal & many-to-one\\
        Ethnicity \citep{GDPR} & nominal & many-to-one\\
        Employment Information \citep{McCallister2010} & nominal & many-to-one\\
        Place of Birth \citep{McCallister2010} & nominal & many-to-one\\
        Skill & nominal & many-to-many\\
        Activities \citep{McCallister2010} & nominal & many-to-many\\
        Diagnosis / Diseases \citep{McCallister2010,Fung2010} & nominal & many-to-many\\
        Origin / Nationality \citep{Poulis2013} & nominal & many-to-many\\
        Purchased Products \citep{Poulis2013} & nominal & many-to-many\\
        Work Shift Schedules & nominal & many-to-many\\
        \bottomrule
    \end{tabular}
    \label{tab:attributes}
\end{table}

\subsection{Problem Formalization}

} 

\label{sec:problem_statement}

\extended{In order to provide a method for anonymizing heterogeneous data composed of relational and textual attributes, we first formalize this problem. 
In the remainder of this work, we will focus on the task of de-identification of heterogeneous datasets containing traditional relational as well as textual data.}
We aim to anonymize a dataset by hiding directly identifying attributes. 
To prevent classical record linkage attacks using quasi-identifying attributes, we use \kanonymity{} as our privacy model~\cite{Sweeney2002k}. 
\extended{In general, identification threats based on information within textual documents can be categorized into two categories, where the former poses explicit and the latter poses implicit information leakage~\cite{SayginHT09}.}
Within texts, we adapt \kanonymity{} to prevent explicit information leakage, while keeping the structure of the texts as best as possible intact to allow for text mining on implicit information. 
In other words, using our privacy model, an attacker shall not be able to identify an individual based on attributes, their values, or sensitive terms in texts.
\extended{However, obfuscating personal writing style as discussed in~\cite{McDonald2012,Fernandes2018} exceeds this work and is therefore not considered.}
Table~\ref{tab:notation} provides an overview of the notations used.

\begin{table}[th]
    \caption[Notation for a given $RX$-Dataset $D$.]{Notation for a given $\mathbf{RX}$-Dataset $\mathbf{D}$.}
    \small
    \begin{tabularx}{\linewidth}{p{0.1\linewidth}X}
    \toprule
      $D$ & original dataset, $D = R_1 \bowtie ... \bowtie R_n$\\
      $R_i$ & relation of $D$\\
      $A^{*}$ & attribute of $D$ identifying an individual directly\\
      $A_i$ & attribute of a relation $R_j$\\
      $X$ & textual attribute\\
      $t$ & tuple in $D$\\
      
      $D^*$ & person centric view on $D$\\
      $r$ & record (tuple) in $D^*$\\
      
      $D'$ & anonymized dataset\\
      
      $X'$ & set of all non-redundant sensitive terms of $X$\\
      $T$ & some text in the form of a sequence of tokens\\
      $\mathbf{F}$ & set of aggregation functions, $\mathbf{F} = \{\mathbf{F_1},...,\mathbf{F_l},\mathbf{F_{X'}}\}$\\
      $E$ & set of sensitive entity types\\
      $er$ & entity recognition function, $er: T \to E$\\
      $emap$ & mapping function, $emap: \{A_1,...,A_p\} \to E$\\
      \bottomrule
     \end{tabularx}
    \label{tab:notation}
\end{table}

\textit{Heterogeneous $RX$-dataset.}
Given a dataset $D$ in form of $n$ relations $R_1,...,R_n$, containing both relational and textual attributes.
$D$ contains all data we want to anonymize.
We pre-process $D$ for the anonymization process following Nergiz et al.~\cite{Nergiz2007} by using the natural join, \ie $D = R_1 \bowtie ... \bowtie R_n$, \ie we ``flatten'' the relational structures.
Table~\ref{tab:example_init} shows an example of a dataset composed of two joined relations, where the first relation describes the individuals (\textit{id}, \textit{gender}, \textit{age}, \textit{topic}, \textit{sign}), while the latter relation (\textit{id}, \textit{date}, \textit{text}) contains the posts and links them to an individual with \textit{id} being the foreign key. 
We call $D$ an $RX$-dataset, if one attribute $A^*$ directly identifies an individual, one or more traditional relational attributes\footnote{By traditional relational attributes we refer to numerical, date, or categorical attributes.\extended{~Categorical attributes might even be composed of multiple terms (\eg names or full addresses).}} $A_i$ contain single-valued data, and one textual attribute\footnote{For the ease of reading we explore only one textual attribute. 
However, our approach can be extended for multiple textual attributes $X_1,...,X_m$.} $X$ is in $D$. 
\extended{In other words, an $RX$-dataset is any dataset, which contains at least one directly identifying attribute, one or more quasi-identifying attributes, and one or more textual attributes.}
\extended{For the remainder of this work, we will use relational attributes for attributes we consider traditional relational and textual attributes for attributes with textual values composed of multiple words or even sentences. 
In the example in Table~\ref{tab:example_init}, the relational attributes are the direct identifier \textit{id} as well as the quasi-identifiers \textit{gender}, \textit{age}, \textit{topic}, and \textit{date}. 
The textual attribute is \textit{text}.}
We call a row in $D$ a tuple $t$. 
Relational attributes $A_i$ are single-valued and can be categorized into being nominal, ordinal, or numerical (\ie ratio or interval, which are treated equally in the anonymization process). 
A textual attribute $X$ is any attribute, where its domain is some form of free text. 
Therefore, we can state that $t.X$ consists of an arbitrary sequence of tokens $T = <t_1,...,t_j>$. 

Note, in the case of partial information, the approach would still work by grouping the tuples with missing attribute values.
For example, if we are missing the attribute \textit{age} for an individual tuple, we can group all people together which  have age missing by assigning a placeholder (\eg ``na'') to the missing age fields.

\textit{Sensitive Entity Types.}
We define $E$ to be a set of entity types, where each value $e \in E$ represents a distinct entity type (\eg person or location) and each entity type is critical for the anonymization task. 
We then define a recognition function $er$ on texts as $er \colon T \to E$. 
The recognition function detects sensitive terms in the text $T$ and assigns a sensitive entity type $e \in E$ to each token $t \in T$. 
Moreover, we define a mapping function $emap$ on the set of structural attributes as $emap \colon \{A_1,...,A_p\} \to E$. 
The mapping function $emap$ maps attributes $A_1,...,A_p$ to a sensitive entity type in $E$, which is used to match redundant sensitive information with the text.

\textit{Redundant Sensitive and Non-redundant Sensitive Terms.} 
Some sensitive information might appear in a textual as well as in a relational attribute.
In order to consistently deal with those occurrences, we introduce the \textit{concept of redundant sensitive information}. 
Redundant sensitive information is any sensitive term $x \in t.X$ with $er(x) = e_j$ for which a relational value $v \in \{ t.A_i \mid \forall t \in D \}$ with $emap(A_i) = e_j$ exists and where $x = v$. 
In other words, redundant sensitive information is duplicated information, \ie has the same value which appears under the same sensitive entity type $e_j$ in a relational attribute $t.A_i$ and a sensitive term $x$ in $t.X$.

We introduce the attribute $X'$, which contains all \textit{non-redundant} sensitive information of $X$. For the remainder of this work, attribute names with apostrophes indicate that these attributes contain the extracted sensitive entities with their types (see $\textit{text}'$ in Table~\ref{tab:example_preprocessed}). We model $X'$ as a set-valued attribute since in texts of $t.X$, zero or more sensitive terms can appear. Therefore, we explicitly allow empty sets to appear in $t.X'$ if no sensitive information appears in $t.X$. We then replace $X$ in $D$ with $X'$, so that the schema of $D$ becomes $\{A^*,A_1,...,A_p,X'\}$.

\textit{Person Centric view $D^*$ on the Dataset $D$.}
If a dataset $D$ is composed of multiple relations, there might be multiple tuples $t$ which correspond to a single individual. In order to apply anonymization approaches on this dataset, we need to group the data in a person centric view similar to \citet{Gong2017}, where one record $r$ (\ie one row) corresponds to one individual. 
Therefore, we define $D^*$ being a grouped and aggregated version of $D$.
This means, that we can retrieve $D^*$ from $D$ as
$D^* = _{A^{*}} \mbox{\Large$G$}_{\mathbf{F_1}(A_1),...,\mathbf{F_l}(A_p),\mathbf{F_{X'}(X')}}(D)$,
where $A^*$ denotes a directly identifying attribute related to an individual used to group rows of individuals together,
{\Large$G$} concurrently applies a set of aggregation functions $\mathbf{F_i}$ and $\mathbf{F_{X'}}$ defined on relational attributes $A_i$ as well as sensitive textual terms $X'$. 
\extended{This aggregation operation should create a person centric view of $D$ by using appropriate aggregation functions $\mathbf{F} = \{\mathbf{F_1},...,\mathbf{F_n},\mathbf{F_{X'}}\}$ on the attributes.}
For relational attributes $A_i$, we use $\mathbf{set}$ as a suitable aggregation function, where two or more distinct values in $A_i$ for one individual result in a set containing all distinct values. For set-based attributes like $X'$, we use the aggregation function $\mathbf{union}$, which performs an element-wise union of all sets in $X'$ related to one individual.
Table~\ref{tab:example_preprocessed} presents a person centric view of our initial example where each record $r$ represents one individual.
Dates as well as any non-redundant sensitive terms have been aggregated, as discussed.

\textit{\kanonymity{} in $D^*$.}
Based on the notion of equivalence classes~\cite{Poulis2013} and the definition of equality of set-based attributes~\cite{He2009}, an equivalence class for $D^*$ can be defined as a partition of records $P$ where for any two records $r,s \in P$ holds $(r.A_1,...,r.A_p) = (s.A_1,...,s.A_p)$ and $r.X'=s.X'$. 
Thus, within an equivalence class each record has the same values for the relational attributes and their sets of sensitive terms have the same values, too.
Given our definition of equivalence classes, a person centric dataset $D^*$ is said to be $k$-anonymous if all equivalence classes of $D^*$ have at least the size $k$. 
We refer to the $k$-anonymous version of $D^*$ as $D'$. 
\extended{$D'$ protects privacy by hiding direct identifiers. Moreover, since each of the quasi-identifying attributes and sensitive terms in texts appear at least $k$ times, $D'$ also protects against record linkage attacks.}

\section{\systemname Approach}
\label{sec:anonymization_approach}

Using the definitions from Section~\ref{sec:problem_statement}, we present our anonymization approach \systemname. 
We present how we preprocess our data to generate a person centric view. 
We show how Mondrian~\cite{LeFevre2006}, a recursive greedy anonymization algorithm, can be used to anonymize $RX$-datasets. 
Mondrian transforms a dataset into a $k$-anonymous version by partitioning the dataset into partitions with sizes greater than $k$ and afterwards recodes each partition individually.
We introduce an alternative partitioning strategy called \ac{GDF}  as a baseline for partitioning a dataset with sensitive terms. 
We use the running example (Table~\ref{tab:example_init}) to show how an $RX$-dataset is transformed to a privacy-preserved version.

\subsection{Pre-processing $D$ to Person-centric View $D^{*}$}

\begin{table*}[ht]
    \centering
    \caption[Preprocessed version of the illustrative example.]{Preprocessed version of the illustrative example. The attribute \textit{date} has been aggregated as \textbf{set}. The attribute \textit{$\text{text}'$} contains sensitive terms of the attribute \textit{text} for all blog posts published by a single individual.}
    \begin{tabularx}{\linewidth}{c|c|c|c|c|c|L}
        \thead{id} & \thead{gender} & \thead{age} & \thead{topic} & \thead{sign} & \thead{date} & \thead{$\text{text}'$} \\\hline
 1 &    male &   36 &  Education &   Aries &         2004-05-14, 2004-05-15 & Pedro\textsubscript{person}, engineer\textsubscript{job}, Mexico\textsubscript{location}\\
 2 &    male &   24 &    Student &     Leo &                      2005-08-18 & engineer\textsubscript{job}\\
 3 &    male &   37 &    Banking &  Pisces &                      2004-05-27 & Ben\textsubscript{person}, Canada\textsubscript{location}\\
 4 &  female &   24 &    Science &   Aries &  2004-01-13, 2004-01-17, 2004-01-19 & Four days ago\textsubscript{date}, scientist\textsubscript{job}, biologist\textsubscript{job}, UK\textsubscript{location}\\
 5 &    male &   29 &     indUnk &  Pisces &                      2004-05-15 &  \\
 6 &  female &   27 &    Science &   Aries &                      2004-05-15 & UK\textsubscript{location}\\
    \end{tabularx}
    \label{tab:example_preprocessed}
\end{table*}

\extended{Prior to anonymizing an $RX$-dataset, it needs to be transformed into a person specific view in order to apply \kanonymity{}.}
Using the running example from Table~\ref{tab:example_init}, we demonstrate the steps involved to create the person centric view shown in Table~\ref{tab:example_preprocessed}. 
First, we identify sensitive terms in the texts and assign sensitive entity types to them. In the remainder of this work, we will use subscripts to indicate the entity type assigned to a sensitive term. Given the first row of the example in Table~\ref{tab:example_init}, the text is ``My name is Pedro, I'm a 36 years old engineer from Mexico''. 
The sensitive terms are \textit{Pedro}\textsubscript{person}, \textit{36 years old}\textsubscript{age}, \textit{engineer}\textsubscript{job}, and \textit{Mexico}\textsubscript{location}. 
This analysis of texts is executed for all tuples $t$ in $D$, while there can be multiple sensitive terms from the same entity type within a text, or even no sensitive terms at all.
In the next step, we find and mark redundant sensitive information using the results of the prior steps. 
Therefore, we perform row-wise analyses of relational values with sensitive terms to find links, which actually represent the same information. 
In our example above, the sensitive term \text{36 years old}\textsubscript{age} depicts the same information as the value \textit{36} in the relational attribute \textit{age}.
Therefore, this sensitive term in the textual attribute is marked as redundant and is not considered as new sensitive information during the anonymization algorithm. Non-redundant sensitive information is stored in the attribute $\textit{text}'$.

Finally, we  build a person-centric view to have a condensed representation of all information available for each individual. 
Therefore, as described in Section~\ref{sec:problem_statement}, we group the data on a directly identifying attribute to get an aggregated dataset. In the example in Table~\ref{tab:example_init}, the directly identifying attribute $A^*$ is \textit{id}. 
We use $\mathbf{set}$ as the aggregation function for the relational attributes. 
Moreover, we collect all sensitive terms mentioned in texts of one individual by performing $\mathbf{union}$ on the sets of sensitive terms. 

Table~\ref{tab:example_preprocessed} shows the person-centric view $D^*$ of our dataset $D$, which has been achieved by aggregating on the attribute \textit{id}. 
Since the individuals with the \textit{ids} \textit{1} and \textit{4} have blogged more than once on different dates, multiple dates have been aggregated as sets. 
Moreover, since those people also have blogged different texts on different days, all sensitive terms across all blog posts have been collected in the attribute $\textit{text}'$.

\subsection{Compute Anonymized Dataset $D'$ from $D^*$}

\label{sec:algorithm}
Given a person centric dataset $D^*$, we want to build a $k$-anonymous version $D'$ by using the definitions of the previous section. 
In order to achieve anonymization, we adapt the two step anonymization algorithm of Mondrian by \citet{LeFevre2006}, which first decides on $m$ partitions $P_1,...,P_m$ (refer to Algorithm~\ref{alg:mondrian}), and afterwards recodes the values of each partition to achieve \kanonymity{}. 
We use \acf{GDF} partitioning as baseline partitioning algorithm (see Algorithm~\ref{alg:partition_tokens}), which uses sensitive terms and their frequencies to create a greedy partitioning using presence and absence of sensitive terms.

\paragraph{Modified Mondrian Partitioning with Weight Parameter $\lambda$}

The first step of the algorithm is to find partitions of records with a partition size of at least $k$. 
\citet{LeFevre2006} introduced multi-dimensional strict top-down partitioning where non-overlapping partitions are found based on all relational attributes. 
Moreover, they introduced a greedy strict top-down partitioning algorithm Mondrian. 
Starting with the complete dataset $D^*$ as an input, the partitioning algorithm chooses an attribute to split on and then splits the partition by median-partitioning. 
The authors suggest using the attribute which provides the widest normalized range given a sub-partition. 
For numerical attributes, the normalized range is defined as minimum to maximum. 
For categorical attributes, the range is the number of distinct categories observed in a partition. 
Sensitive, textual terms are treated as categorical attributes.

In order to properly treat textual terms in this heuristic algorithm, we introduce a weight parameter $\lambda$ to the modified Mondrian algorithm shown in Algorithm~\ref{alg:mondrian}. 
\extended{$\lambda$ can be a value between $0$ and $1$.}
It describes the priority to split partitions on relational attributes. 
$\lambda = 1$ means that the algorithm always favors to split on relational attributes. $\lambda = 0$ leads to splits only based on sensitive terms in textual attributes. $\lambda = 0.5$ does not influence the splitting decisions and therefore is considered as default.
The partitioning algorithm stops if no allowable cut can be made such that the criteria of \kanonymity{} holds for both sub-partitions. 
Therefore, we can stop splitting partitions if $|P| < 2k$.
 
\begin{algorithm}[ht]
\caption{Modified Mondrian partitioning with weight parameter $\lambda$ (adapted from \citet{LeFevre2006}). 
\extended{It applies a greedy strict top-down partitioning for relational attributes.}
}
\small
\label{alg:mondrian}
\LinesNumbered
\SetKwInOut{Input}{Input}\SetKwInOut{Output}{Output}
\SetKwFunction{next_term}{next\_term}
\SetKwFunction{nextattribute}{next\_attribute}
\SetKwFunction{mondrianpartitioning}{mondrian\_partitioning}
\SetKwFunction{frequencyset}{frequency\_set}
\SetKwFunction{findmedian}{find\_median}
\SetKwProg{Fn}{Function}{:}{}
\Input{Partition $P$, weight $\lambda$}
\Output{Set of partitions with size of at least $k$}
\Fn{\mondrianpartitioning{$P$, $\lambda$}}{
\If(\tcp*[h]{no allowable cut}){$|P| < 2k$}{ 
    \KwRet{$P$}
}
\Else{
    $A = \nextattribute{$\lambda$}$\\
    $F = \frequencyset{P, A}$\\
    $P_l = (r \in P | r.A < \findmedian{F})$\\
    $P_r = P \setminus P_l$\\
    \KwRet{\mondrianpartitioning{$P_l$} $\cup$ \mondrianpartitioning{$P_r$}}\\
}
}
\end{algorithm}

\paragraph{\acl{GDF} (\ac{GDF}) Partitioning}
Using the idea of a top-down strict partitioning algorithm, we propose with \ac{GDF} a greedy partitioning algorithm using the presence and absence of sensitive terms. 
The main goal is to keep the same sensitive terms within the same partition. 
This is achieved by creating partitions with records which have sensitive terms in common. 
Algorithm~\ref{alg:partition_tokens} presents the \ac{GDF} partitioning algorithm\extended{, which is based on sensitive terms and their frequencies}.
Similar to Algorithm~\ref{alg:mondrian}, we start with the whole dataset as a single partition. 
Instead of splitting the partition using the median of a relational attribute (Mondrian partitioning), \textit{we split partitions on a chosen sensitive term}. 
While the first sub-partition contains only records, where the chosen sensitive term appears, the second sub-partition contains the remaining records. 
For choosing the next term to split on, multiple heuristics are possible. 
We propose to use the most frequently apparent sensitive term for the remaining texts in the partition as the term to split on. 
Taking the most frequent term allows us to keep the most frequently appearing term in a majority of texts while suppressing less frequently used terms. 
The term used to split is then removed and similar to Algorithm~\ref{alg:mondrian} the algorithm is recursively called using the first and second partition, respectively.
\extended{\ac{GDF} partitioning guarantees that records are partitioned such that sensitive terms in texts are tried to be kept by grouping records with same terms. 
Moreover, records with no or less frequently used sensitive terms are also included in one partition.
Therefore, we build partitions with records which would prevent other partitions from being $k$-anonymous.}

\begin{algorithm}[ht]
\caption{Top-down document-frequency-based (\ac{GDF}) partitioning on sensitive terms in $X'$.}
\small
\label{alg:partition_tokens}
\LinesNumbered
\DontPrintSemicolon
\SetKwInOut{Input}{Input}\SetKwInOut{Output}{Output}
\SetKwFunction{nextterm}{next\_term}
\SetKwFunction{gdfpartitioning}{gdf\_partitioning}
\SetKwProg{Fn}{Function}{:}{}
\Input{Partition $P$, terms with their frequencies $F$}
\Output{Set of partitions with size of at least $k$}
\Fn{\gdfpartitioning{$P, F$}}{
\If(\tcp*[h]{no allowable cut}){$|P| < 2k$}{ 
    \KwRet{$P$}
}
\Else{
    $(x,f) = \nextterm{P, F}$\\ 
    $P_l = (r \in P | x \in r.X')$\\
    $P_r = P \setminus P_l$\\
    $F = F \setminus \{(x,f)\}$ \tcp*[h]{remove considered term}\\
    \KwRet{\gdfpartitioning($P_l, F$) $\cup$ \gdfpartitioning($P_r, F$)}
}
}
\end{algorithm}

\extended{
\paragraph{Example} 
Using the running example in Table~\ref{tab:example_preprocessed} with $k = 2$ and the \ac{GDF} partitioning scheme, partitioning is achieved as follows. 
Starting with the initial complete dataset $D^*$ (person-centric view) as the initial partition $P$, we determine the most frequent term, which is either \textit{UK} or \textit{engineer}, both appearing twice. 
Without loss of generality, we assume \textit{engineer} is chosen as the term to split on. 
Then we split $P = \{1,2,3,4,5,6\}$ in $P_l = \{1,2\}$ containing all records where \textit{engineer} appears and $P_r = \{3,4,5,6\}$ containing the remaining records. 
For $P_l$, no allowable cut can be made since $|P_l| = 2$. 
However, the algorithm continues with $P_r$ since $|P_r| = 4$, and splits $P_r$ on \textit{UK} as the most frequent term appearing twice in records within $P_r$. 
This will lead to two new partitions Using $P_{rl} = \{4,6\}$ containing records where \textit{UK} appears in the texts and $P_{rr} = \{3,5\}$ containing the remaining records. 
Finally, the algorithm results in an optimal partitioning with three partitions, each consisting of two records. 
In our case, we refer to optimal as a partition layout with the least amount of information loss within the textual attribute.
}

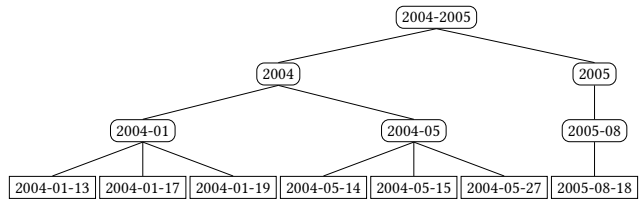
\begin{figure}[ht]
    \centering
    \resizebox{\columnwidth}{!}{
        \begin{tikzpicture}[sibling distance=0.2em]
            \tikzset{every internal node/.style={shape=rectangle, rounded corners, draw, align=center}}
            \tikzset{every leaf node/.style={shape=rectangle, draw, align=center}}
            \Tree
            [.2004-2005
                [.2004
                    [.2004-01
                        2004-01-13
                        2004-01-17
                        2004-01-19
                    ]
                    [.2004-05
                        2004-05-14
                        2004-05-15
                        2004-05-27
                    ]
                ]
                [.2005
                    [.2005-08
                        2005-08-18
                    ]
                ]
            ]
        \end{tikzpicture}
    }
    \caption[\acl{DGH} for date attributes.]{\acl{DGH} for date attributes. 
    \shortorextended{Leaves depict values in the dataset $D$. First level of generalization suppresses day, followed by month, and year.}{The leaves depict actual values appearing in the dataset $D$. The first level of generalization involves suppressing the day. 
    The second level of generalization suppresses the month. 
    Finally, the root can be automatically generated as a range of years.}}
    \label{fig:date_hierarchy}
\end{figure}

\paragraph{Recoding}
In the next step, each partition is transformed such that values of quasi-identifiers of records are indistinguishable. 
This process is called recoding. 
\extended{Recoding can either be global~\cite{Bayardo2005} or local~\cite{Xu2006,LeFevre2006}. 
Local recoding generalizes values per equivalence class, but equal values from two equivalence classes might be recoded differently. 
In contrast, global recoding enforces that the same values are recoded equally throughout the entire dataset. 
Since global recoding requires a global replacement of values with appropriate recoded values, the search space for appropriate replacements may be limited~\cite{LeFevre2005}. 
Therefore, even though global recoding might result in more consistent releases of data, local recoding appears to be more powerful due to its variability in finding good replacements.} 
There are different recoding schemes for the different scales of the attribute.
Nominal and ordinal values are usually recoded using \acp{DGH} as introduced by \citet{Sweeney2002} and used in multiple other works~\cite{Nergiz2007,Ghinita2007,Xu2006,Poulis2013}. 
A \ac{DGH} describes a hierarchy which is used to generalize distinct values to a more general form such that within a partition all values transform to a single value in the \ac{DGH}. 
\extended{Generating \acp{DGH} is usually considered a manual effort, while there already exist approaches on automatically generating concept hierarchies as introduced by \citet{Lee2008}, which have also been used in work on anonymization~\cite{He2009}.} Alternatively, nominal and ordinal attributes can also be recoded as sets containing all distinct items of one partition. 
For numerical attributes, \citet{LeFevre2006} propose to use either mean or range as a summary statistic. 
Additionally, numerical attributes can also be recoded using ranges from minimum to maximum. 
Moreover, for dates \citet{ElEmam2009} propose an automated hierarchical recoding based on suppressing some information of a date value shown in Figure~\ref{fig:date_hierarchy}. 
The leaf nodes represent actual dates appearing in the dataset $D$ (ref. to Table~\ref{tab:example_init}). 
Non-leaf nodes represent automatically generated values by suppressing information on each level.

Since we use a strict-multidimensional partitioning scheme, we apply local recoding as suggested in Mondrian~\cite{LeFevre2006}. For numerical attributes, we use range as a summary statistic. For date attributes we use the automatically generated \ac{DGH} by \citet{ElEmam2009} as shown in Figure~\ref{fig:date_hierarchy}. Moreover, since generalization hierarchies for gender, topic, and sign are flat, we recode nominal and ordinal values as sets of distinct values.

\paragraph{Example}
After equivalence classes have been determined, relational attributes can be recoded. 
Table~\ref{tab:example_anonymized} shows how those recoding schemes are applied to the relational attributes of our running example. 
In addition, a $k$-anonymous representation of the text attribute $X'$ has to be created. 
Terms, which are marked as redundant sensitive information, are replaced by the recoded value of its relational representatives. 
Using the anonymized version of our example in Table~\ref{tab:example_anonymized}, the age appearing in the text of the first row is recoded using the value of the attribute \textit{age} of the same row. 
Moreover, non-redundant sensitive information is recoded using suppression with its entity type. 
If a sensitive information appears within all records of an equivalence class, retaining this information complies with our definition of \kanonymity{} for set-valued attributes from Section~\ref{sec:problem_statement}.
Therefore, it does not need to be suppressed (see sensitive term \textit{engineer} in Table~\ref{tab:example_anonymized}). 
However, if the same sensitive information is not appearing in every record within an equivalence class, this sensitive information (or the lack of it) violates our definition of \kanonymity{} and must be suppressed. 
An example for such a violation in Table~\ref{tab:example_anonymized} is \textit{Mexico}, which appears in the first record, but in no other record of its equivalence class.
The result is the $k$-anonymized dataset $D^*$.

\begin{table*}[ht]
    \centering
    \caption{Anonymized dataset $D'$\extended{~of the running example} for $k=2$. Redundant information and remaining sensitive terms are marked bold.}
    \begin{tabularx}{\linewidth}{c|c|c|c|c|c|L}
        \thead{id} & \thead{gender} & \thead{age} & \thead{topic} & \thead{sign} & \thead{date} & \thead{text} \\\hline
 1 &    male &  \textbf{[24-36]} &  (Student,Education) &  (Leo,Aries) &  [2004-2005] & My name is \textbf{person}, I’m a \textbf{[24-36]} years old \textbf{engineer} from \textbf{location}.  \\
 1 &    male &  [24-36] &  (Student,Education) &  (Leo,Aries) &  [2004-2005] &  A quick follow up: I will post updates about my education in more detail. \\
 2 &    male &  [24-36] &  (Student,Education) &  (Leo,Aries) &  [2004-2005] &  I will start working for a big tech company as an \textbf{engineer}. \\
 3 &    male &  [29-37] &     (indUnk,Banking) &       Pisces &      2004-05 &  During my last business trip to \textbf{location} I met my friend \textbf{person} from college. \\
 4 &  female &  [24-27] &              Science &        Aries &         2004 &            As a \textbf{job} from the \textbf{UK}, you can be proud! \\
 4 &  female &  [24-27] &              Science &        Aries &         2004 &  \textbf{Date}, I started my blog. Stay tuned for more content. \\
 4 &  female &  [24-27] &              \textbf{Science} &        Aries &         \textbf{2004} &  \textbf{2004} will be a great year for \textbf{science} and for my career as a \textbf{job}. \\
 5 &    male &  [29-37] &     (indUnk,Banking) &       \textbf{Pisces} &      2004-05 &  Did you know that \textbf{Pisces} is the last constellation of the zodiac. \\
 6 &  female &  [24-27] &              Science &        Aries &         2004 &  Rainy weather again here in the \textbf{UK}. I hope you all have a good day! \\
    \end{tabularx}
    \label{tab:example_anonymized}
\end{table*}

\section{Experimental Apparatus}
\label{sec:experiments}
We evaluate our framework on two real-world datasets using the modified Mondrian partitioning algorithm with weighting parameter $\lambda$ as well as the \ac{GDF} partitioning baseline.
We use $\lambda$ to manipulate the splitting decisions in Mondrian as discussed in Section~\ref{sec:anonymization_approach} and measure the resulting partitions as well as information loss. 

\subsection{Datasets}
\extended{We require datasets that include a directly identifying attribute $A^*$, one or more quasi-identifying relational attributes $A_i$, and one or more textual attributes $X$ containing sensitive information about individuals (refer to the definition of an $RX$-dataset in Section~\ref{sec:problem_statement}).
We use the publicly available \textit{Blog Authorship Corpus} and \textit{515K Hotel Reviews Data in Europe} datasets.}

\paragraph{Blog Authorship Corpus}
The Blog Authorship Corpus\footnote{\url{https://www.kaggle.com/rtatman/blog-authorship-corpus}} was originally used to create profiles from authors~\cite{Schler2006} but has also been used in privacy research for author re-identification~\cite{Koppel2006}.
After cleaning the input data from unreadable characters and others, the corpus contains $681,260$ blog posts from $19,319$ bloggers, which have been written by a single individual on or before 2006 and published on blogger.com. 
\extended{While the vast majority of blog posts are written in English language, the corpus contains some posts written in other languages. 
However, non-English blog posts are the minority and therefore do not have a significant impact on the experiment results.}
A row in the corpus consists of the \textit{id}, \textit{gender}, \textit{age}, \textit{topic}, and \textit{zodiac sign} of a blogger as well as the \textit{date} and the \textit{text} of the published blog entry. 
Each row corresponds to one blog post written by one individual, but one individual might have written multiple blog posts. 
On average, one blogger has published $35$ blog posts.
We treat \textit{id} as a direct identifier, \textit{gender}, \textit{topic}, and \textit{sign} as categorical attributes, while \textit{age} is treated as a numerical attribute. The attribute \textit{date} is treated as a special case of categorical attribute where we recode dates using the automatically generated \ac{DGH} shown in Figure~\ref{fig:date_hierarchy}. The attribute \textit{text} is used as the textual attribute. The attribute \textit{topic} contains $40$ different topics, including industry-unknown (indUnk). \textit{Age} ranges from $13$ to $48$. \textit{Gender} can be male or female. \textit{Sign} can be one of the twelve astrological signs. 

\extended{In addition to the Blog Authorship Corpus, we run experiments on a second dataset to verify our observations. 
We chose to use a dataset containing reviews of European hotels. 
We refer to this dataset as the \textit{Hotel Reviews Dataset}.}

\paragraph{Hotel Reviews Dataset}
We use the 515K Hotel Reviews Data in Europe dataset\footnote{\url{https://www.kaggle.com/jiashenliu/515k-hotel-reviews-data-in-europe}}, called in the following briefly the \textit{Hotel Reviews Dataset}, which contains $17$ attributes, of which $15$ attributes are relational and two attributes are textual.
The textual attributes are \textit{positive} and \textit{negative reviews} of users. 
Among the relational attributes, we treat \textit{hotel name} and \textit{hotel address} as direct identifiers. 
\extended{The textual attributes are pre-processed and cleaned as described for the Blog Authorship Corpus.}
\textit{Negative} and \textit{positive word count} as well as \textit{tags} are ignored and therefore considered insensitive attributes. 
The remaining attributes are treated as quasi-identifiers, with seven numerical, one date, and two nominal attributes. 
\extended{We recode all quasi-identifying attributes similar to the Blog Authorship Corpus.}
After preparing the Hotel Reviews Dataset, we have $512,126$ reviews for $1,475$ hotels remaining.

\extended{\vspace{1cm}}

\subsection{Procedure}

\extended{
As baselines, we consider the scenario where relational and textual attributes are anonymized independently. 
Usually, sensitive terms within a textual
attribute are suppressed completely, which leads to total loss of utility of sensitive terms. 
With our experiments we want to show that we can improve, \ie reduce the information loss in texts under the \kanonymity{} model. 
Moreover, we want to optimize the trade-off between relational and textual information loss.
}

Similar to experiments conducted in prior work \citep{Gong2017,Ghinita2007,Poulis2013}, we run our anonymization tool for different values of $k=2$, $3$, $4$, $5$, $10$, $20$, and $50$.
Regarding our new weighting parameter $\lambda$, we used values between $0.0$ and $1.0$ in steps of $0.1$. 
Sensitive entity types in texts are those detected by spaCy's English models trained on the OntoNotes5 corpus\footnote{\url{https://spacy.io/api/data-formats\#named-entities}}.
We added rule-based detectors for the entities MAIL, URL, PHONE, and POSTCODE.
We treat all sensitive terms appearing under those entity types as quasi-identifiers. 
For each value of $k$, we conduct experiments using different partitioning strategies and parameter settings. 
In particular, we vary the weight parameter $\lambda$ to tune Mondrian. 
\extended{To speed up experiment execution times, we ignore redundant sensitive information. Ignoring redundant sensitive information does not influence the experiment results, since both datasets do not provide a relevant amount of overlap between relational attributes and textual attributes.}
We use the same recoding scheme, namely local recoding, for all experiments to make partitioning results comparable. 
For the evaluation, we analyze the anonymized dataset with respect to the corresponding partitioning sizes and information loss.

\extended{In addition, we repeat the experiments by just considering location entities with entity type GPE (geopolitical entity). 
We use those experiments to showcase an anonymization task with reduced complexity. 
We chose location-based entities since they are present in blog posts as well as in hotel reviews.
Therefore, they allow for comparison of both datasets.}

\subsection{Measures}\label{sec:metrics}

\extended{In order to evaluate our anonymization approach and compare results of partitioning, we introduce the following measures. 
In particular, we compare statistics on partitions as well as relational and textual information loss.}

\paragraph{Statistics on Partitions}

\extended{We are interested in the resulting partitions of the anonymized dataset. 

\textit{Number of splits (based on relational versus textual attributes):~}}
We evaluate how partitions are created, based on relational attributes versus textual attributes, and how $\lambda$ influences splitting decisions. 
\extended{We expect that for $\lambda < 0.5$ we observe more splits on textual attributes and for $\lambda > 0.5$ more splits on relational attributes.

\textit{Number of partitions and Partition sizes:~}}
In addition to the number of splits, we want to evaluate the size of the resulting partitions since they are closely related to information loss.
By the nature of \kanonymity{}, all partitions need to be at least of size $k$. Relatively large partitions with respect to $k$ will tend to produce more information loss. 
Therefore, partition sizes closer to $k$ will be favorable and increase utility. 
\extended{We evaluate resulting partitions by counting the number of partitions, as well as calculating the mean and standard deviation of partition sizes. }

\paragraph{Information Loss (Adapted to Heterogeneous Datasets)}
Measuring the information loss of an an\-on\-ymi\-zed dataset is well-known practice for evaluating the amount of utility remaining for a published dataset.
We use \acf{NCP}~\cite{Xu2006} to determine how much information loss has been introduced by the anonymization process. 
\extended{In particular, the \ac{NCP} assigns a penalty to each data item in a dataset according to the amount of uncertainty introduced.}
We extend the definitions of \ac{NCP} 
to the problem of anonymizing relational and textual data such that for one record $r$, the information loss is calculated as 
$ NCP(r) = ( w_R \cdot NCP_A(r) + w_X \cdot NCP_X(r) ) / ( w_A + w_X )$, 
where $w_A$ is the importance assigned to the relational attributes, and $NCP_A(r)$ denotes the information loss for relational attributes of record $r$. 
Analogously, we define $w_X$ and $NCP_X(r)$ for the textual attribute. 
For our evaluation, we set $w_A$ and $w_X$ to $1$, \ie weigh the loss stemming from relational data and textual data equally.
\extended{Note, that this decision is independent of the $\lambda$ parameter, which decides which attribute or term is actually used for the splitting of the partitions.}

For \textit{relational attributes} $A = \{A_1,...,A_p\}$ we define the information loss 
$ NCP_A(r) = ( \sum_{A_i \in A} NCP_{A_i}(r) ) / |A| $, 
where $|A|$ denotes the number of relational attributes.
$NCP_{A_i}$ is the information loss for a single attribute and depends on the type of attribute.
It can be calculated either using $NCP_{num}$ for numerical attributes or $NCP_{cat}$ for categorical attributes.
$NCP_{num}$ for numerical values is defined as 
$ NCP_{num}(r) = ( z_{i}-y_{i} ) / |A_{i}| $,
with $z_i$ being the upper and $y_i$ being the lower bound of the recoded numerical interval and $|A_i| = max_{r \in D^*}(r.A_i) - min_{r \in D^*}(r.A_i)$. 
For categorical values, $NCP_{cat}$ is defined as 
 $  NCP_{cat}(r) = \begin{cases}
        0 & |u| = 1\\
        \frac{|u|}{|A_i|} & \text{otherwise}
    \end{cases}$,
where $|u|$ denotes the number of distinct values which the recoded value $u$ describes. 
For categorical values other than dates, $|u|$ will be the number of distinct values appearing in the recoded set. 
For date attributes, $|u|$ denotes the number of leaves of the subtree below the recoded value (see Figure~\ref{fig:date_hierarchy}).

For \textit{textual attributes}, we define 
$NCP_{X}(r) = ( \sum_{x \in r.X'} NCP_x(x) ) /$ $|r.X'|$,
where for each sensitive information $x$, we calculate the individual information loss $NCP_x(x)$ and normalize it by the number of sensitive terms $|r.X'|$. 
We define the individual information loss for one sensitive term as
$ NCP_{x} (x) = 1$ if $x$ is suppressed, and $0$ otherwise.

Finally, we can calculate the total information loss for an entire $RX$-Dataset $D^*$ as 
$ NCP(D^*) = ( \sum_{r \in D^*} NCP(r) ) / |D^*|$,
where for each record $r$ the information loss $NCP(r)$ is calculated and divided by the number of records $|D^*|$.

\section{Results}
\label{sec:results}
We present the results regarding partition statistics and information loss.
For detailed experimental results with plots and tables for all parameter values, we refer to the supplementary material.
\extended{In particular, details on the influence of $\lambda$ and $k$ on the splitting decision for both datasets can be found in the supplementary material in Appendix~\ref{appendix:partition_splits}.
Detailed tables on the influence of $\lambda$ and $k$ on the partition count and size can be found in Appendix~\ref{appendix:partition_count_and_size}.
Detailed figures of the information loss, including zoomed plots, for the experiments run on the Blog Authorship Corpus and Hotel Reviews Dataset can be found in Appendix~\ref{appendix:information_loss}.
Figures with a detailed comparison of information loss per entity type in the Blog Authorship Corpus and Hotel Reviews Dataset for different values of $\lambda$ can also be found in Appendix~\ref{appendix:information_loss}.
}

\subsection{Partitions Splits, Counts, and Size}
\extended{To modify splitting decisions and therefore the distribution of information loss between relational and textual attributes, we introduced the tuning parameter $\lambda$ to Mondrian partitioning. 
Thus, first, we verify how $\lambda$ impacts splitting decisions.
We count for a particular $\lambda$ how often partitions are effectively split on a relational attribute and compare this metric to the number of splits on sensitive terms of textual attributes. 
We also evaluate the count of the resulting partitions and the partitions' sizes.}

\begin{figure}[ht]
    \begin{center}
    \centering
    \subfloat[][Textual vs. relational splits on the Blog Authorship Corpus]{
        \resizebox{1\columnwidth}{!}{
            \begin{adjustbox}{clip,trim=0cm 2.9cm 0cm 0.5cm}
            \input{images/splits_blogs_all_5_shrinked.pgf}
            \end{adjustbox}
        }
        \label{fig:partition_splits_blogs}
      }
    \end{center}

    \begin{center}
    \subfloat[][Textual vs. relational splits on the Hotel Reviews Dataset]{
        \resizebox{1\columnwidth}{!}{
            \begin{adjustbox}{clip,trim=0cm 1.8cm 0cm 0.5cm}
            \input{images/splits_reviews_all_5_shrinked.pgf}
            \end{adjustbox}
        }
        \label{fig:partition_splits_hotels}
    }
    \end{center}

    \caption{Number of splits based on textual attributes (orange) versus relational attributes (blue) using Mondrian partitioning ($k = 5$) with varying weights $\lambda$\extended{~on the (a)~Blog Authorship Corpus and (b)~Hotel Reviews Dataset}.}
    \label{fig:partition_splits}
\end{figure}
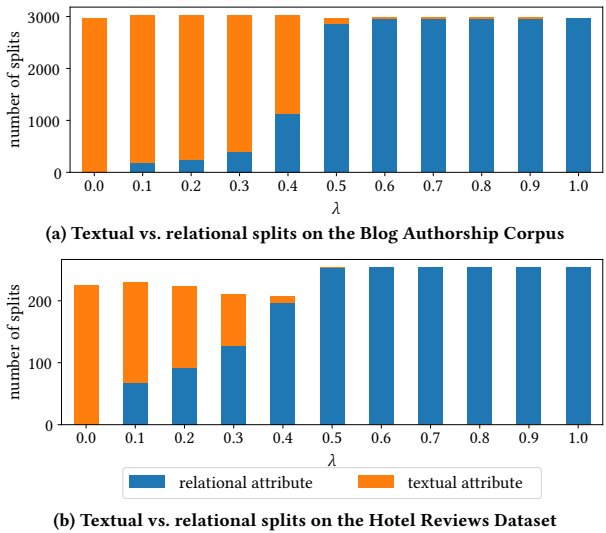

\paragraph{Partition Splits}
Figure~\ref{fig:partition_splits_blogs} shows the distribution of splitting decisions for experiments run on the Blog Authorship Corpus for $k = 5$ and $\lambda = 0.0$ to $1.0$.
\extended{As $\lambda$ was designed, $\lambda = 1$ results in only splits on relational attributes whereas $\lambda = 0$ results in splits only on sensitive terms.}
As our results show, an unbiased run of Mondrian with $\lambda = 0.5$ causes partitions to be split mostly on relational attributes. 
Since the span of relational attributes is lower compared to sensitive terms, relational attributes provide the widest normalized span and are therefore favored to split on.
For $\lambda > 0.5$, the majority of the weight for splitting is given to the relational attributes.
\extended{Thus, there is no relevant change since relational attributes are considered almost every time throughout the partitioning phase.} 
However, for $\lambda < 0.5$, we observe that an increasing number of splits are made based on textual attributes.
For $\lambda = 0.4$, already more than half of the splits are based on textual terms.
\extended{
If only locations, \ie entities of type GPE, are considered, $\lambda$ is not in all cases able to control the share of splits between relational and textual attributes, since low values for $\lambda$ do not result in more splits on textual attributes.
}
\extended{Plots are omitted for brevity here and can be found in the supplementary materials.

}
For the Hotel Reviews Dataset, shown in Figure~\ref{fig:partition_splits_hotels}, the number of splits is generally lower (see also partition sizes, below), since it contains fewer records. 
Also the splitting on textual attributes is less likely for hotel reviews compared to blog posts. 
\extended{
In the case of experiments considering only location entities, the impact of $\lambda$ is even smaller and splits are mostly performed on relational attributes.
Details are provided in the supplementary materials.
}

\paragraph{Partition Count and Size}
Regarding the number of partitions and their size, Table~\ref{tab:part_stat_blogs_all-brief} provides statistics on partitions using Mondrian partitioning with varying $\lambda$ as well as \ac{GDF} partitioning for the Blog Authorship Corpus. 
\extended{In the table, \textit{count} refers to the number of partitions produced under the specific values of $k$ and $\lambda$, while \textit{size} refers to the average number of records per partition.}
The Mondrian partitioning algorithm produces the same partitioning layout for $\lambda$ between $0.6$ and $0.9$. 
This observation matches statistics on partition splits, since for these values of $\lambda$ the Mondrian algorithm decides to use the same attributes to split on.
Furthermore, \ac{GDF} partitioning is not able to generate partition sizes close to $k$, compared to Mondrian partitioning. 
\extended{Table~\ref{tab:part_stat_blogs_gpe-brief} shows the results when only location entities are considered.
Here, $\lambda = 0$ leads to bigger and fewer partitions compared to other settings for $\lambda$. 
Comparing \ac{GDF} to Mondrian with $\lambda = 0$, we observe that for low numbers of $k$, \ac{GDF} partitioning achieves in general smaller, but more variable partitions with regard to size. 
However, for larger values of $k$, Mondrian partitioning achieves better distribution of partitions and therefore better distribution of sensitive terms.

}
The results for the Hotel Reviews Dataset are shown in Table~\ref{tab:part_stat_hotels_all-brief}.
\extended{Table~\ref{tab:part_stat_hotels_gpe-brief} shows the results for the location type only.}
We make the same observations as for the Blog Authorship Dataset.
However, due to the lower number of records in the Hotel Reviews Dataset, the total count of partitions is comparatively smaller.

\begin{table}[th]
\centering
\caption{Statistics on partitions for Blog Authorship Corpus\extended{~considering all entity types}. 
Count refers to the number of partitions found, while size refers to the average number of records per partition.}
\label{tab:part_stat_blogs_all-brief}
\small
\begin{tabular}{lllrrrrrrr}
\midrule[\heavyrulewidth]
                & $\lambda$  &    k & 3  &     4  &     5  &     10 &      20 \\
        \midrule[\heavyrulewidth]
\parbox[t]{2mm}{\multirow{2}{*}{\rotatebox[origin=c]{90}{\textbf{GDF}}}} & \multirow{2}{*}{-} & count &  2479 &   1512 &   1078 &    352 &      92 \\
                   & & size & 7.79 &  12.78 &  17.92 &  54.88 &  209.99 \\
        \midrule[\heavyrulewidth]
\parbox[t]{2mm}{\multirow{10}{*}{\rotatebox[origin=c]{90}{\textbf{Mondrian}}}} & \multirow{2}{*}{0} & count &  5162 &   3795 &   2971 &   1412 &     692 \\
                   & & size & 3.74 &   5.09 &   6.50 &  13.68 &   27.92 \\
        \cmidrule{2-8}
& \multirow{2}{*}{0.3} & count &  5236 &   3841 &   3023 &   1462 &     707 \\
                   & & size &   3.69 &   5.03 &   6.39 &  13.21 &   27.33  \\
        \cmidrule{2-8}

& \multirow{2}{*}{0.5} & count & 5198 &   3800 &   2979 &   1441 &     711 \\
                   & & size &  3.72 &   5.08 &   6.49 &  13.41 &   27.17  \\
        \cmidrule{2-8}
& \multirow{2}{*}{0.6 - 0.9} & count &  5180 &   3781 &   2987 &   1441 &  711\\
                   & & size &  3.73 &   5.11 &   6.47 &  13.41 &   27.17 \\
        \cmidrule{2-8}
& \multirow{2}{*}{1} & count &  5128 &   3749 &   2964 &   1431 &     703 \\
                   & & size &  3.77 &   5.15 &   6.52 &  13.50 &   27.48 \\
\bottomrule
\end{tabular}
\end{table}

\extended{
\begin{table}[th]
\centering
\caption{Statistics on resulting partitions for Blog Authorship Corpus considering only GPE entities}
\label{tab:part_stat_blogs_gpe-brief}
\small
\begin{tabular}{lllrrrrrrr}
\midrule[\heavyrulewidth]
    & $\lambda$  &    k &    3  &     4  &     5  &     10 &      20\\
        \midrule[\heavyrulewidth]
\parbox[t]{2mm}{\multirow{2}{*}{\rotatebox[origin=c]{90}{\textbf{GDF}}}} & \multirow{2}{*}{-} & count & 737 &     301 &     129 &        2 &      1  \\
                   & & size & 26.21 &   64.18 &  149.76 &  9659.50 &  19319  \\
        \midrule[\heavyrulewidth]
\parbox[t]{2mm}{\multirow{10}{*}{\rotatebox[origin=c]{90}{\textbf{Mondrian}}}} & \multirow{2}{*}{0} & count &  1164 &     828 &     703 &      392 &    242  \\
                   & & size & 16.60 &   23.33 &   27.48 &    49.28 &  79.83  \\
        \cmidrule{2-8}
& \multirow{2}{*}{0.3} & count & 5248 &    3780 &    2994 &     1443 &    700 \\
                   & & size & 3.68 &    5.11 &    6.45 &    13.39 &  27.60  \\
        \cmidrule{2-8}

& \multirow{2}{*}{0.5} & count & 5154 &    3762 &    2970 &     1433 &    703 \\
                   & & size &  3.75 &    5.14 &    6.50 &    13.48 &  27.48 \\
        \cmidrule{2-8}
& \multirow{2}{*}{0.6 - 0.9} & count &  5153 &    3761 &    2970 &     1433 &    703 \\
                   & & size &   3.75 &    5.14 &    6.50 &    13.48 &  27.48  \\
        \cmidrule{2-8}
& \multirow{2}{*}{1} & count &  5128 &    3749 &    2964 &     1431 &    703  \\
                   & & size &  3.77 &    5.15 &    6.52 &    13.50 &  27.48 \\
\bottomrule
\end{tabular}
\end{table}
}

\begin{table}[th]
\centering
\caption{Statistics on partitions for Hotel Reviews Dataset\extended{~ considering all entity types}.
Count refers to the number of partitions found, while size refers to the average number of records per partition.}
\label{tab:part_stat_hotels_all-brief}
\small
\begin{tabular}{lllrrrrrrr}
\midrule[\heavyrulewidth]
    & $\lambda$  &    k &    3  &     4  &     5  &     10 &      20\\
        \midrule[\heavyrulewidth]
\parbox[t]{2mm}{\multirow{2}{*}{\rotatebox[origin=c]{90}{\textbf{GDF}}}} & \multirow{2}{*}{-} & count &  272 &   163 &   127 &     43 &     16  \\
                   & & size &  5.42 &  9.05 & 11.61 &  34.30 &  92.19 \\
        \midrule[\heavyrulewidth]
\parbox[t]{2mm}{\multirow{10}{*}{\rotatebox[origin=c]{90}{\textbf{Mondrian}}}} & \multirow{2}{*}{0} & count &  398 &   293 &   226 &    117 &     54  \\
                   & & size & 3.71 &  5.03 &  6.53 &  12.61 &  27.31 \\
        \cmidrule{2-8}
& \multirow{2}{*}{0.3} & count &  404 &   285 &   212 &    106 &     50  \\
                   & & size &  3.65 &  5.18 &  6.96 &  13.92 &  29.50  \\
        \cmidrule{2-8}
& \multirow{2}{*}{0.5} & count &  415 &   256 &   255 &    128 &     64 \\
                   & & size & 3.55 &  5.76 &  5.78 &  11.52 &  23.05 \\
        \cmidrule{2-8}
& \multirow{2}{*}{0.6 - 1} & count &  417 &   256 &   255 &    128 &     64 \\
                   & & size &   3.54 &  5.76 &  5.78 &  11.52 &  23.05 \\
\bottomrule
\end{tabular}
\end{table}

\extended{
\begin{table}[th]
\centering
\caption{Statistics on resulting partitions for Hotel Reviews Dataset considering only GPE entities.}
\label{tab:part_stat_hotels_gpe-brief}
\small
\begin{tabular}{lllrrrrrrr}
\midrule[\heavyrulewidth]
    &   $\lambda$ &    k &   3  &     4  &     5  &     10 &      20 \\
        \midrule[\heavyrulewidth]
\parbox[t]{2mm}{\multirow{2}{*}{\rotatebox[origin=c]{90}{\textbf{GDF}}}} & \multirow{2}{*}{-} & count &  107 &     47 &     15 &      1 &      1 \\
                   & & size &  13.79 &  31.38 &  98.33 &   1475 &   1475 \\
        \midrule[\heavyrulewidth]
\parbox[t]{2mm}{\multirow{8}{*}{\rotatebox[origin=c]{90}{\textbf{Mondrian}}}} & \multirow{2}{*}{0} & count &  49 &     35 &     32 &     17 &     13  \\
                   & & size & 30.10 &  42.14 &  46.09 &  86.76 & 113.46  \\
        \cmidrule{2-8}

& \multirow{2}{*}{0.3} & count &  376 &    312 &    211 &    101 &     50  \\
                   & & size &  3.92 &   4.73 &   6.99 &  14.60 &  29.50  \\
        \cmidrule{2-8}

& \multirow{2}{*}{0.5 - 1} & count &  417 &    256 &    255 &    128 &  64  \\
                   & & size &  3.54 &   5.76 &   5.78 &  11.52 &  23.05 \\
\bottomrule
\end{tabular}
\end{table}
}

\subsection{Information Loss}

\extended{In addition to the statistics on partition splits, counts, and sizes, we are interested in how the partitioning performs with respect to the introduced information loss measure.}
Figure~\ref{fig:relational_information_loss_blog} provides an overview on relational information loss $NCP_A$ (y-axis) for different values of $k$ between $2$ and $50$ (x-axis) for the Blog Authorship Corpus.
Figure~\ref{fig:textual_information_loss_blog} shows the textual information loss $NCP_X$. \extended{Results for $\lambda$ between $0.6$ and $0.9$ are not plotted, since they are almost identical to the run using $\lambda = 0.5$.}
Figures~\ref{fig:relational_information_loss_hotels} and \ref{fig:textual_information_loss_hotels} provide the information loss for experiments run on the Hotel Reviews dataset. 

\begin{figure}[ht]
    \centering
    \subfloat[][Relational information loss $\mathbf{NCP_A}$]{
      \resizebox{0.49\linewidth}{!}{
      \begin{adjustbox}{clip,trim=0.6cm 0.8cm 0.8cm 0.5cm}
      \input{images/ril_blogs_all.pgf}
      \end{adjustbox}
      }
      \label{fig:relational_information_loss_blog}
    }
    \hfill
    \subfloat[][Textual information loss $\mathbf{NCP_X}$]{
      \resizebox{0.49\linewidth}{!}{
       \begin{adjustbox}{clip,trim=0.6cm 0.8cm 0.8cm 0.5cm}
       \input{images/xil_blogs_all.pgf}
       \end{adjustbox}
      \label{fig:textual_information_loss_blog}
      }
    }
    \caption{Information loss for relational attributes \protect\subref{fig:relational_information_loss_blog}, and textual attributes \protect\subref{fig:textual_information_loss_blog} on the Blog Authorship Corpus.}
    \label{fig:information_loss_blog}
\end{figure}

\begin{figure}[ht]
    \centering
    \subfloat[][Relational information loss $\mathbf{NCP_A}$]{
      \resizebox{0.49\linewidth}{!}{
       \begin{adjustbox}{clip,trim=0.6cm 0.8cm 0.8cm 0.5cm}
      \input{images/ril_hotels_all.pgf}
      \end{adjustbox}
      \label{fig:relational_information_loss_hotels}
      }    
    }
    \hfill
    \subfloat[][Textual information loss $\mathbf{NCP_X}$]{
      \resizebox{0.49\linewidth}{!}{
       \begin{adjustbox}{clip,trim=0.6cm 0.8cm 0.8cm 0.5cm}
      \input{images/xil_hotels_all.pgf}
      \end{adjustbox}
      \label{fig:textual_information_loss_hotels}
      }
    }
    \caption{Information loss for relational attributes \protect\subref{fig:relational_information_loss_hotels}, and textual attributes \protect\subref{fig:textual_information_loss_hotels} on the Hotel Reviews Dataset.}
    \label{fig:information_loss_hotels}
\end{figure}

\paragraph{Relational Information Loss}
The information loss increases with larger $k$ throughout all experiments. 
Higher information loss is caused by having larger partitions and therefore higher efforts in recoding. 
Furthermore, we can state that information loss in the relational attributes increases if the tuning parameter $\lambda$ decreases (see Figure~\ref{fig:relational_information_loss_blog}). 
This observation agrees with statistics on splitting decisions, since for lower values of $\lambda$, Mondrian more frequently decides to split on sensitive terms in textual attributes. 
This leads to more variations in relational values of partitions, which ultimately increases the relational information loss.
\extended{
In experiments where only locations are considered, GDF partitioning as well as Mondrian partitioning with $\lambda = 0$ result in relatively high relational information loss compared to other experiment runs (see figures in the supplementary material). 
In both cases, the high relational information loss is caused by having partitions split only based on one option, namely the recognized sensitive locations appearing in the textual attribute (cf. previous section).
}
Comparing with Figure~\ref{fig:relational_information_loss_hotels}, we can state that relational information loss appears to be higher for the Hotel Reviews Dataset compared to the Blog Authorship Corpus. 
However, we still observe the same behavior where higher values of $\lambda$ result in relatively lower relational information loss. 

\paragraph{Textual Information Loss}
Analyzing the information loss in the textual attribute, see Figure~\ref{fig:textual_information_loss_blog}, one observation is that for values of $k \geq 10$ the information loss in texts tends to become $1$. 
This equals suppressing all sensitive terms in texts. 
Moreover, our modified Mondrian partitioning performs better compared to the naive partitioning strategy GDF. 
GDF partitioning results in partitions with unequal and larger sizes and therefore ends up with large partitions, which significantly increase information loss. 
\extended{Moreover, GDF partitioning decides on splitting partitions taking a single global maximum (most frequent term) ignoring the multi-dimensionality and diversity of sensitive terms in texts.}
We make the same observations on the Hotel Reviews Dataset plotted in Figure~\ref{fig:textual_information_loss_hotels}. 
However, information loss for $k \leq 5$ tends to be slightly lower. 
\extended{
If only locations are considered, textual information loss in hotel reviews can significantly be reduced (see figures in the supplementary material). 
Since the Hotel Reviews Dataset only contains reviews for hotels in Europe, there is a limited number of locations that are included. 
This leads to significant preservation of sensitive terms even for values of $k \leq 10$.
}

\subsection{Attribute-level Textual Information Loss}

\begin{figure}[ht]
\begin{center}
\subfloat[][Entity-level textual loss on the Blog Authorship Corpus]{
    \resizebox{0.8\columnwidth}{!}{
        \begin{adjustbox}{clip,trim=0cm 0cm 0cm 0.5cm}
            \input{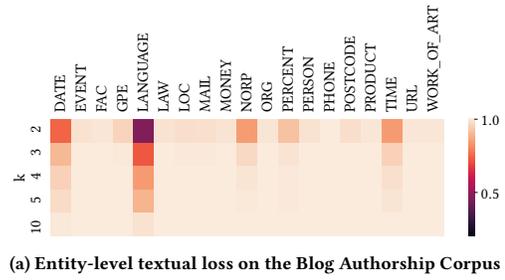}
        \end{adjustbox}
    }
    \label{fig:detailed_xil_blogs_0_2}
}

\subfloat[][Entity-level textual loss on positive reviews of Hotel Review Dataset ]{
    \resizebox{0.8\columnwidth}{!}{
        \begin{adjustbox}{clip,trim=0cm 0cm 0cm 0.5cm}
            \input{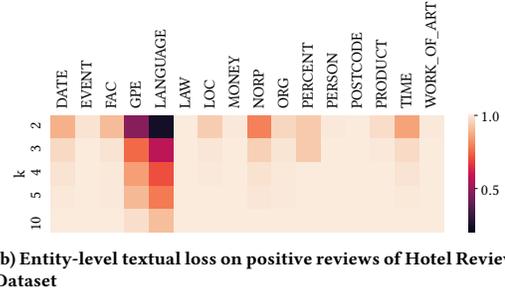}
        \end{adjustbox}
     }
    \label{fig:detailed_xil_hotels_positive_review_0_2}
}

\subfloat[][Entity-level textual loss on negative reviews of Hotel Review Dataset ]{
        \resizebox{0.8\columnwidth}{!}{
        \begin{adjustbox}{clip,trim=0cm 0cm 0cm 0.5cm}
            \input{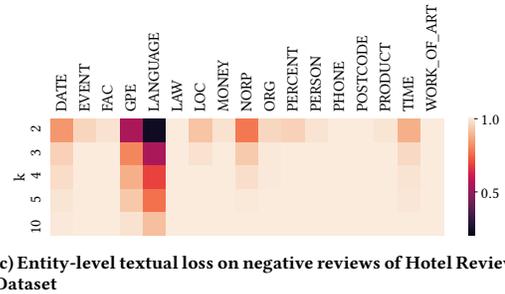}
        \end{adjustbox}
        }
    \label{fig:detailed_xil_hotels_negative_review_0_2}
}

\end{center}

\caption{Textual information loss per entity type with Mondrian ($\lambda = 0.2$). 
Not all entity types appear in all datasets.}
\label{fig:detailed_xil_attributed_based}
\end{figure}

To get a deeper understanding of textual attributes on the anonymization process, we analyzed textual information loss on entity type level. 
Figure~\ref{fig:detailed_xil_blogs_0_2} provides an overview of information loss per different entity type extracted from text in the Blog Authorship Corpus for $k$ is $2$ to $50$ and a fixed $\lambda = 0.2$. 
It shows that there is a high information loss for most attributes, even for small $k$.
However, information loss of sensitive terms of type LANGUAGE may be reduced for values of $k \leq 5$. 
Since the number of distinct entities of type LANGUAGE is much lower compared to other entity types in the Blog Authorship Corpus like EVENT and PERSON, the entities (\ie number of sensitive terms) of type LANGUAGE can be better preserved. 
We obtain similar results for Mondrian partitioning with  other values of $\lambda \leq 0.4$. 
We make the same observations on the Hotel Reviews dataset for both textual attributes, the positive reviews and negative reviews (see Figures~\ref{fig:detailed_xil_hotels_positive_review_0_2} and \ref{fig:detailed_xil_hotels_negative_review_0_2}). 
In addition to LANGUAGE entities, sensitive locations (GPE) can also be preserved for both textual attributes.

\section{Discussion}
\label{sec:discussion}

\extended{\subsection{Key Results}}
Due to heterogeneity of sensitive terms in texts, by default, they are less likely to be considered to split on. 
By introducing the tuning parameter $\lambda$ in our framework, we were able to control the Mondrian algorithm to preserve more information in either relational or textual attributes. 
\extended{Our experiments show that the partitioning parameter may be tuned in order to favor information preservation in textual attributes over relational attributes.~}
We observe that a value of $\lambda$ between $0.4$ and $0.5$ results in balanced splits, \ie about the same number of splits are based on relational attributes versus textual terms.
Our anonymization approach allows us to reduce the information loss in texts under the \kanonymity{} privacy model. 
In contrast, in the related work~\cite{Liu2017,Dernoncourt2017,Sweeney1996} sensitive terms have been completely suppressed. 
Furthermore, our experiments show that for $k \leq 5$, not all sensitive terms need to be suppressed. 
In case of entities of type LANGUAGE, our approach could preserve about $60 \%$ for $k=2$ in the Blog Authorship Corpus (see Figure~\ref{fig:detailed_xil_blogs_0_2}) and up to $80\%$ of terms for $k=2$ in the Hotel Reviews Dataset (see Figures~\ref{fig:detailed_xil_hotels_positive_review_0_2} and \ref{fig:detailed_xil_hotels_negative_review_0_2}).
Generally, when applying \kanonymity{} on sensitive terms, it works better for texts from a specific domain (\eg hotels) than cross-domain datasets (\eg blogs), as the latter have a higher diversity.

\extended{
\subsection{Threads to Validity}
While our approach presents a general framework to anonymize heterogeneous data, our choices on detecting and comparing sensitive terms may have an impact on the experiments' outcomes. 
We consider all sensitive terms in the texts to be quasi-identifiers. 
However, in certain situations, sensitive entity types should---similar to relational attributes---also be distinguished in direct and quasi-identifying attributes. 
Having a distinction between direct and quasi-identifiers is necessary in cases where texts include many names, or other identifiers appearing for multiple records.}
There may be a possible over-anonymization or under-anonymi\-za\-tion in our \systemname approach, influenced by the accuracy of the detected sensitive terms.
Over-anonymization resembles the case where sensitive terms are falsely suppressed.
It is caused by low precision and reduces utility of the anonymized data.
\extended{This happens, when terms, which do not pose any risk of identity disclosure, are  anonymized and the text loses important structures due to the missing terms. 
If sensitive terms are labeled with false entity types, they might also falsely be anonymized, since our strict definition of \kanonymity{} requires also entity types to be equal.}
Under-anonymization describes a case where sensitive terms are falsely kept.
This case is generally considered more critical than falsely suppressing terms and is related to low recall. 
If entities which should have been anonymized are not detected at all, the information they provide will appear in the released dataset and might reveal information which should not have been disclosed. 
We address this thread of validity and use a state-of-the-art \ac{NLP} library spaCy to extract named entities from text.
We use spaCy's recent transformer-based language model~\cite{bert} for English (Version 3.0.0a0)\extended{\footnote{Available at \url{https://github.com/explosion/spacy-models/releases//tag/en_core_web_trf-3.0.0a0}}}, which has an F1-score for \ac{NER} tasks of 89.41.
However, there are cases such as misspellings, jargon, foreign words, and others that we are yet missing.
There can be different ways to express the same sensitive information which leads to over-anonymization.
For example, the capital city of Germany may be referred to simply by its actual name ``Berlin'' or indirectly referred to as ``Germany's capital''.
It is not possible for our current system to resolve such linkage. 
\extended{We refer to such cases as false negative matches. }
There may also be identical terms which actually have different semantics, which leads to under-anonymization.
In example, consider the phrases ``I live in Berlin'' and ``I love Berlin''\extended{, which appear in two different records and would happen to be grouped into the same partition}. 
Our approach would treat both appearances of ``Berlin'' the same way\extended{~even though in the first case it is referring to a place of residence while in the second case it is an expression of preference}. 
\extended{We refer to such a scenario as false positive matches. }
In order to mitigate such \extended{false positive and negative~}cases, one can integrate more advanced text matching functions to our \systemname{} framework, potentially depending on the requirements of a specific use case.
\extended{For false negative matches, one may introduce synonym tables, semantic rules, and metrics such as Levenshtein distance to cope with spelling mistakes.
To cope with false positive matches, one suggestion is to consider the surrounding context by comparing Part-of-Speech-Tags and dependencies of terms within and across sentences.}
For example, one could use contextualized word vectors~\cite{DBLP:conf/nips/MikolovSCCD13,bert}.
Note, in this work, we focus on showing that heterogeneous data can be anonymized using our \systemname approach\extended{~and demonstrate the influence of the splitting parameter $\lambda$ on the creation of the partitions}. 
\extended{Using different extensions to \systemname such as word matching functions is prepared by proposing a framework approach and can be integrated and evaluated as required by a different use case or dataset.}

\extended{Finally, false negative matches and false positive matches can also occur on redundant sensitive information. 
While false negative matches result in inconsistencies in the released data, false positive matches obfuscate semantic meaning of sensitive terms in texts.}

\extended{\subsection{Generalizability}
Our work has multiple implications which can be beneficial for other work.
We showed that anonymizing unstructured text data can be achieved by extracting sensitive terms and casting the task into a structured anonymization problem.
One may generalize the concept also for semi-structured data such as JSON documents.}
The idea of linking relational fields to attributes of other data types could be extended in order to retrieve a consistent, and privacy-preserved version of heterogeneous JSON or XML documents~\cite{DBLP:journals/tkde/GkountounaT15}. 
In addition, tuning the partitioning using a parameter like $\lambda$ is not only relevant in the context of anonymizing heterogeneous data, but could also be adapted to an attribute level to favor distinct attributes over others. 
\extended{An adjustable attribute-level bias within the partitioning phase of Mondrian would allow users to prioritize preservation of information in specific attributes. 
Suppose that one department within an organization shares data with a second department, which should do an age-based market analysis of sold products, but should not get access to raw data and therefore receive an anonymized version. 
As a consequence, the department providing data could adjust the anonymization using a bias to preserve more information in relevant attributes (\ie age), and less information in others.}

\section{Conclusion}
We introduced \systemname as a step towards a framework for anonymizing hybrid documents consisting of relational as well as textual attributes.
We have formally defined the problem of joint anonymizing heterogeneous datasets.
This is achieved by transferring sensitive terms in texts to an anonymization task of structured data, introducing the concept of redundant sensitive information, and establishing the tuning parameter $\lambda$ to control and prioritize information loss in relational as well as textual attributes.
We have demonstrated the usefulness of \systemname at the example of two real-world datasets using the privacy model \kanonymity{}~\cite{Sweeney2002k}. 

\textbf{Data Availability and Reproducibility}: 
\extended{Although extensive success has been achieved in anonymizing different types of data, there is limited work in the field of anonymizing heterogeneous data. Therefore, we would like to emphasize the importance and encourage researchers to investigate combined anonymization approaches for heterogeneous data to receive a consistent and privacy-preserved release of data.}
The source code is available at \url{https://github.com/Serpinex3/rx-anon}\shortorextended{~as well as more technical details at \url{https://arxiv.org/abs/2105.08842}}

\extended{As a framework approach, \systemname can be extended in all aspects of the anonymization pipeline, namely the \textit{partitioning, string matching, recoding, privacy model, and supported entity types}. 
Particularly, we are interested to see how anonymizing heterogeneous data can be achieved using other anonymization techniques than \kanonymity{} and use contextualized text similarity functions~\cite{DBLP:conf/nips/MikolovSCCD13,bert}. 
A detailed discussion of the extensibility of our framework is provided in Appendix~\ref{appendix:future-work}.}

\bibliographystyle{ACM-Reference-Format}
\bibliography{library}

\newpage

\renewcommand{\appendixpagename}{Supplementary Materials}
\begin{appendices}

\section{Extended Experimental Results}
\label{sec:extended_experiments}

The following sections contain extended experiment results. In particular, we provide numbers of distinct entities, give information about the performance of our framework, share statistics relevant for partitioning, and present details on information loss.

\subsection{Distinct Terms}
Table~\ref{tab:distinct_terms} provides an overview of the number of distinct terms appearing in textual attributes. In general, the texts of the Blog Authorship Corpus contain significantly more distinct entities.

\begin{table}[H]
    \centering
    \caption[Numbers of distinct terms per entity type.]{Numbers of distinct terms per entity type. The Blog Authorship Corpus contains one textual attribute \textit{text}. The Hotel Reviews Dataset contains two textual attributes, namely \textit{negative review} and \textit{positive review}.}
    \label{tab:distinct_terms}
    \begin{tabular}{l|r|r|r}
        \toprule
         & \multicolumn{1}{M{1.5cm}|}{\bfseries Blog Authorship Corpus} & \multicolumn{2}{M{3.15cm}}{\bfseries Hotel Reviews Dataset}\\
        entity type & \multicolumn{1}{M{1.5cm}|}{text} & \multicolumn{1}{M{1.5cm}|}{negative review} & \multicolumn{1}{M{1.5cm}}{positive review}\\
        \midrule
        DATE & 83,972 & 2,672 & 1,993\\
        EVENT & 13,883 & 161 & 244\\
        FAC & 32,864 & 3,452 & 14,070\\
        GPE & 34,639 & 1,058 & 2,512\\
        LANGUAGE & 761 & 30 & 30\\
        LAW & 5,153 & 12 & 3\\
        LOC & 13,635 & 480 & 1,370\\
        MAIL & 3,225 & 0 & 0\\
        MONEY & 16,050 & 2,089 & 625\\
        NORP & 9,676 & 293 & 344\\
        ORG & 162,555 & 3,887 & 9,444\\
        PERCENT & 4,104 & 30 & 25\\
        PERSON & 245,667 & 2,273 & 5,728\\
        PHONE & 442 & 1 & 0\\
        POSTCODE & 739 & 7 & 8\\
        PRODUCT & 48,207 & 842 & 892\\
        TIME & 61,669 & 4,311 & 2,366\\
        URL & 29,297 & 0 & 0\\
        WORK\_OF\_ART & 145,421 & 290 & 349\\
        \bottomrule
    \end{tabular}
\end{table}

\subsection{Performance}
Table~\ref{tab:performance} provides valuable insights in execution times of the experiments. Each experiment was executed on a single CPU core and did not require to analyze the texts, since the processed \ac{NLP} state is read from cached results. In the case of experiments run on the Blog Authorship Corpus, execution times were significantly higher compared to the Hotel Reviews dataset. One observation is that if only relational attributes are considered (Mondrian, $\lambda = 1$), execution times come down to a fraction of experiments where sensitive terms are considered during the partitioning phase.

Considering memory consumption, running a single experiment on the Blog Authorship Corpus required 25.2~GB for all entities and 13.4~GB in the case of just considering GPE entities (locations). In the case of the Hotel Review dataset, 5.4~GB and 4.2~GB were required respectively.

\begin{table}[H]
    \centering
    \caption[Execution times of experiments.]{Execution times of experiments in hh:mm:ss.}
    \label{tab:performance}
    \begin{tabular}{ll|c|c|c|c}
        \toprule
        \multicolumn{2}{c|}{} & \multicolumn{2}{c|}{\thead{Blog Authorship\\Corpus}} & \multicolumn{2}{c}{\thead{Hotel Reviews\\Dataset}}\\
         & $\lambda$ & all & GPE & all & GPE\\
        \midrule
        \parbox[t]{2mm}{\multirow{2}{*}{\rotatebox[origin=c]{90}{\textbf{GDF}}}} & \multirow{2}{*}{-}                    & \multirow{2}{*}{10:25:12} & \multirow{2}{*}{03:50:13} & \multirow{2}{*}{00:12:55} & \multirow{2}{*}{00:10:26}\\[0.4 cm]
        \midrule
        \parbox[t]{2mm}{\multirow{11}{*}{\rotatebox[origin=c]{90}{\textbf{Mondrian}}}} & 0 & 15:19:31 & 02:46:14 & 00:29:26 & 00:11:44\\
                                   & 0.1 & 15:01:34 & 01:32:25 & 00:29:56 & 00:13:16\\
                                   & 0.2 & 14:43:37 & 01:30:11 & 00:29:44 & 00:12:58\\
                                   & 0.3 & 14:32:23 & 01:24:31 & 00:29:17 & 00:12:49\\
                                   & 0.4 & 13:59:02 & 01:15:58 & 00:27:59 & 00:12:44\\
                                   & 0.5 & 11:03:43 & 01:04:59 & 00:25:13 & 00:12:18\\
                                   & 0.6 & 11:05:29 & 01:05:04 & 00:25:12 & 00:12:20\\
                                   & 0.7 & 11:06:26 & 01:04:47 & 00:25:14 & 00:12:20\\
                                   & 0.8 & 11:03:33 & 01:04:50 & 00:25:01 & 00:12:23\\
                                   & 0.9 & 11:02:57 & 01:04:32 & 00:25:12 & 00:12:19\\
                                   & 1 & 01:48:41 & 00:47:22 & 00:12:48 & 00:11:17\\
        \bottomrule
    \end{tabular}
\end{table}

\subsection{Partitions}
In our experiments, we evaluate statistics on partition splits to gain insights how $\lambda$ influences splitting decisions of Mondrian partitioning. Moreover, we also share statistics on resulting partitions.

\subsubsection{Partition Splits}
\label{appendix:partition_splits}

\begin{figure*}[p]
    \centering
    \resizebox{\linewidth}{!}{
    \input{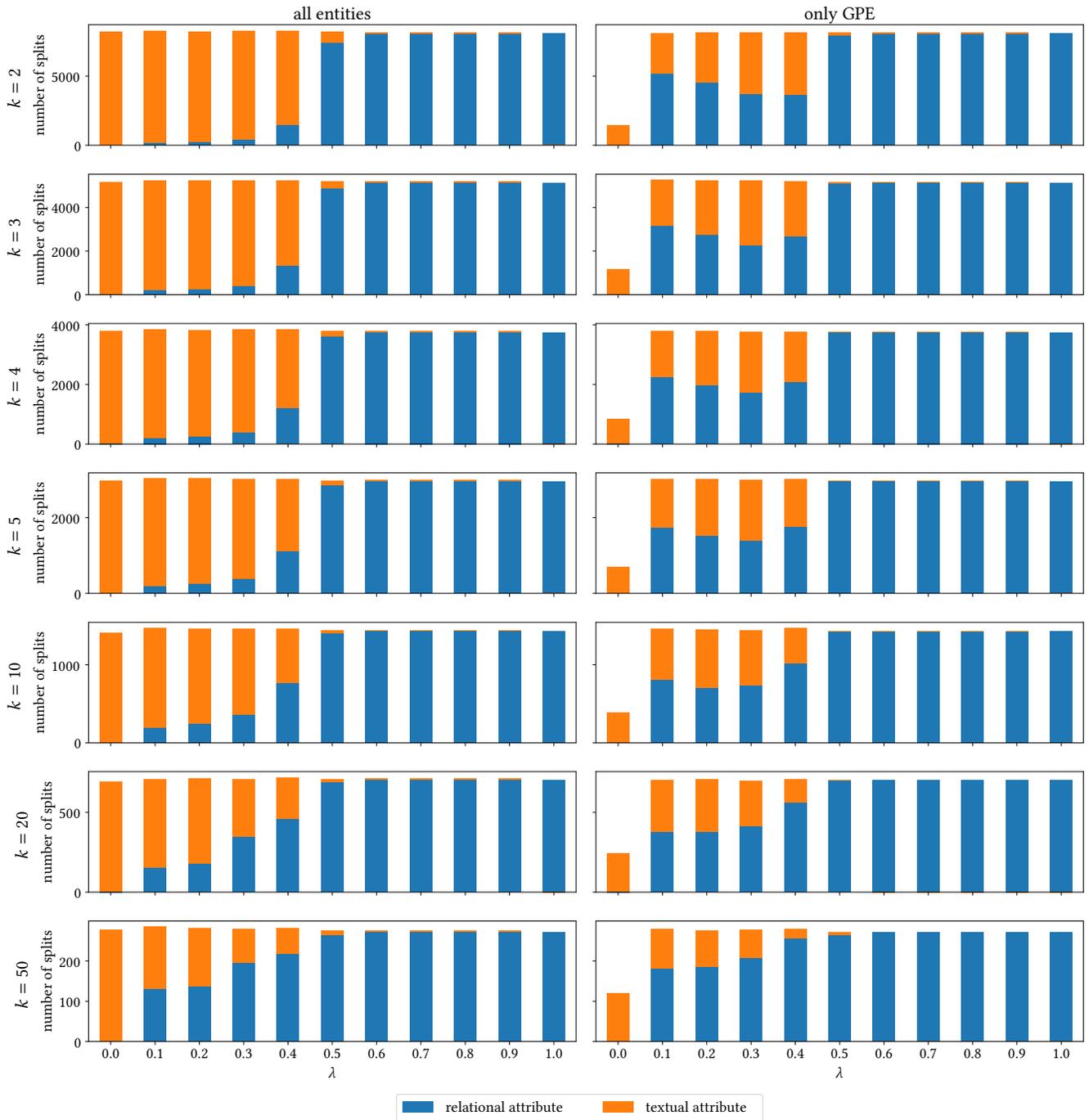}
    }
    \caption{Splitting statistics for the Blog Authorship Corpus. Left plots are results for experiments run considering all entities. 
    Right plots represent statistics for experiments run only considering GPE entities.}
    \label{fig:partition_splits_combined_blogs}
\end{figure*}

Figure~\ref{fig:partition_splits_combined_blogs} provides an overview of the distribution of splitting decisions between relational and textual attributes for experiments run on the Blog Authorship Corpus for all values of $k$.
The left column includes experiments considering all entities, while the right column presents results for experiments run only considering location (GPE) entities. 

A noteworthy observation is that for a fixed $\lambda$, the number of splits on textual attributes decreases if $k$ increases.
Since we are only considering valid splits, sensitive terms have to appear at least $2k$ times within a partition to be split on. 
Therefore, in case of $k = 50$, sensitive terms are required to appear $100$ times, which is less likely due to heterogeneity of blog post texts. 
 
If only locations, \ie entities of type GPE, are considered, $\lambda$ is not in all cases able to control the share of splits between relational and textual attributes, since low values for $\lambda$ do not result in more splits on textual attributes.
This effect is caused by the lack of multi-dimensionality. 
Since only one category of sensitive entity types is considered, Mondrian has only one option (namely split on sensitive terms with type GPE) to split on textual attributes. 
If splits on GPE terms fail (\eg if there are none), Mondrian will ultimately continue to split on a relational attribute.

\begin{figure*}[p]
    \centering
    \resizebox{\linewidth}{!}{
    \input{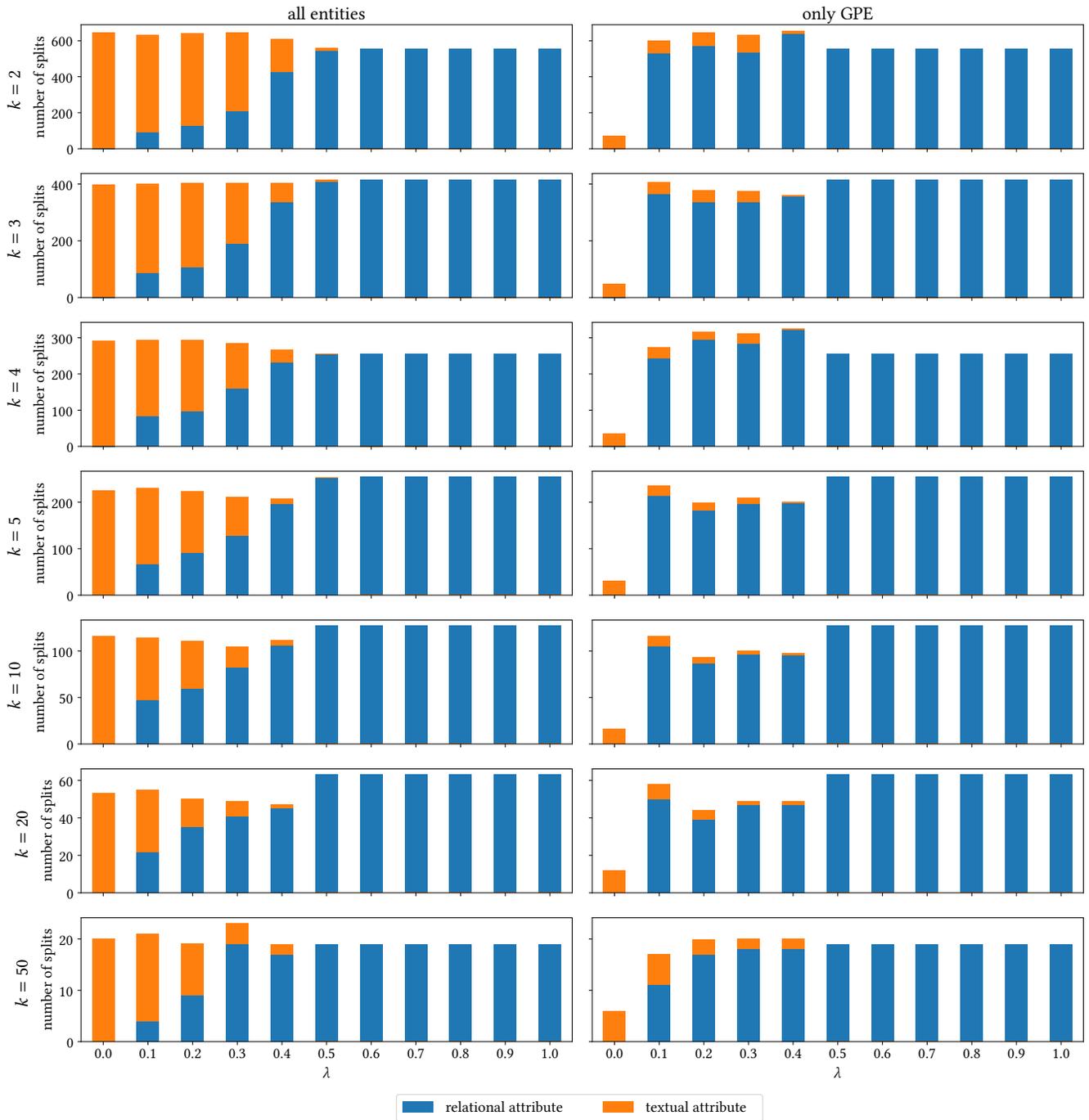}
    }
    \caption{Splitting statistics for Hotel Reviews Dataset. 
    Left plots are results for experiments run considering all entities. 
    Right plots represent statistics for experiments run only considering GPE entities.}
    \label{fig:partition_splits_combined_hotels}
\end{figure*}

Similarly, Figure~\ref{fig:partition_splits_combined_hotels} highlights the impact of $\lambda$ on partition splits for experiments run on the Hotel Reviews Dataset for all values of~$k$.

\subsubsection{Partition Count and Size}
\label{appendix:partition_count_and_size}

We present the results regarding partition statistics. Table~\ref{tab:part_stat_blogs_all} provides valuable insights on the number of partitions as well as the size and standard deviation regarding partition sizes for the experiments on the Blog Authorship Corpus considering all entities. Similarly, Table~\ref{tab:part_stat_blogs_gpe} provides an overview of the same metrics for the Blog Authorship Corpus only considering GPE entities (locations). Tables~\ref{tab:part_stat_hotels_all} and~\ref{tab:part_stat_hotels_gpe} share insights on partition statistics for the Hotel Reviews dataset.

\begin{table*}[p]
\centering
\small
\caption{Statistics on resulting partitions for Blog Authorship Corpus considering all entity types.}
\label{tab:part_stat_blogs_all}
\begin{tabular}{lllrrrrrrr}
\midrule[\heavyrulewidth]
                &   &    k &    2  &     3  &     4  &     5  &     10 &      20 &      50 \\
         & $\lambda$ &  &  &  & & & & & \\
        \midrule[\heavyrulewidth]
\parbox[t]{2mm}{\multirow{3}{*}{\rotatebox[origin=c]{90}{\textbf{GDF}}}} & \multirow{3}{*}{-} & count &  5810 &   2479 &   1512 &   1078 &    352 &      92 &       7 \\
                   & & size &  3.33 &   7.79 &  12.78 &  17.92 &  54.88 &  209.99 & 2759.86 \\
                   & & std & 26.78 & 103.95 & 199.22 & 292.95 & 726.15 & 1742.66 & 7146.68 \\
        \midrule[\heavyrulewidth]
\parbox[t]{2mm}{\multirow{27}{*}{\rotatebox[origin=c]{90}{\textbf{Mondrian}}}} & \multirow{3}{*}{0} & count &  8219 &   5162 &   3795 &   2971 &   1412 &     692 &     278 \\
                   & & size &  2.35 &   3.74 &   5.09 &   6.50 &  13.68 &   27.92 &   69.49 \\
                   & & std &  0.76 &   1.15 &   1.48 &   1.91 &   3.56 &    6.76 &   18.28 \\
        \cmidrule{2-10}
& \multirow{3}{*}{0.1} & count &  8277 &   5226 &   3841 &   3028 &   1471 &     710 &     286 \\
                   & & size &  2.33 &   3.70 &   5.03 &   6.38 &  13.13 &   27.21 &   67.55 \\
                   & & std &  0.47 &   0.78 &   1.06 &   1.35 &   2.74 &    5.77 &   14.40 \\
        \cmidrule{2-10}
& \multirow{3}{*}{0.2} & count &  8243 &   5234 &   3829 &   3031 &   1460 &     715 &     283 \\
                   & & size &  2.34 &   3.69 &   5.05 &   6.37 &  13.23 &   27.02 &   68.27 \\
                   & & std &  0.47 &   0.79 &   1.06 &   1.36 &   2.79 &    5.67 &   14.64 \\
        \cmidrule{2-10}
& \multirow{3}{*}{0.3} & count &  8302 &   5236 &   3841 &   3023 &   1462 &     707 &     280 \\
                   & & size &  2.33 &   3.69 &   5.03 &   6.39 &  13.21 &   27.33 &   69.00 \\
                   & & std &  0.47 &   0.78 &   1.06 &   1.34 &   2.82 &    5.81 &   14.47 \\
        \cmidrule{2-10}
& \multirow{3}{*}{0.4} & count &  8307 &   5238 &   3855 &   3024 &   1466 &     720 &     283 \\
                   & & size &  2.33 &   3.69 &   5.01 &   6.39 &  13.18 &   26.83 &   68.27 \\
                   & & std &  0.47 &   0.78 &   1.06 &   1.36 &   2.77 &    5.57 &   14.50 \\
        \cmidrule{2-10}
& \multirow{3}{*}{0.5} & count &  8186 &   5198 &   3800 &   2979 &   1441 &     711 &     278 \\
                   & & size &  2.36 &   3.72 &   5.08 &   6.49 &  13.41 &   27.17 &   69.49 \\
                   & & std &  0.48 &   0.77 &   1.07 &   1.34 &   2.76 &    5.68 &   14.59 \\
        \cmidrule{2-10}
& \multirow{3}{*}{0.6 - 0.9} & count &  8168 &   5180 &   3781 &   2987 &   1441 &     711 &     276 \\
                   & & size &  2.37 &   3.73 &   5.11 &   6.47 &  13.41 &   27.17 &   70.00 \\
                   & & std &  0.48 &   0.77 &   1.07 &   1.35 &   2.76 &    5.69 &   14.43 \\
        \cmidrule{2-10}
& \multirow{3}{*}{1} & count &  8080 &   5128 &   3749 &   2964 &   1431 &     703 &     273 \\
                   & & size &  2.39 &   3.77 &   5.15 &   6.52 &  13.50 &   27.48 &   70.77 \\
                   & & std &  0.51 &   0.80 &   1.10 &   1.38 &   2.80 &    5.78 &   14.63 \\
\bottomrule
\end{tabular}
\end{table*}

\begin{table*}[p]
\centering
\caption{Statistics on resulting partitions for Blog Authorship Corpus considering only GPE entities}
\label{tab:part_stat_blogs_gpe}
\begin{tabular}{lllrrrrrrr}
\midrule[\heavyrulewidth]
                &   &    k &    2  &     3  &     4  &     5  &     10 &      20 &      50 \\
         & $\lambda$ &  &  &  & & & & & \\
        \midrule[\heavyrulewidth]
\parbox[t]{2mm}{\multirow{3}{*}{\rotatebox[origin=c]{90}{\textbf{GDF}}}} & \multirow{3}{*}{-} & count &  2140 &    737 &     301 &     129 &        2 &      1 &      1 \\
                   & & size &  9.03 &  26.21 &   64.18 &  149.76 &  9659.50 &  19319 &  19319 \\
                   & & std & 575.34 & 1016.03 & 1629.78 & 13645.04 &      0 &      0 \\
        \midrule[\heavyrulewidth]
\parbox[t]{2mm}{\multirow{27}{*}{\rotatebox[origin=c]{90}{\textbf{Mondrian}}}} & \multirow{3}{*}{0} & count &  1447 &   1164 &     828 &     703 &      392 &    242 &    121 \\
                   & & size &  13.35 &  16.60 &   23.33 &   27.48 &    49.28 &  79.83 & 159.66 \\
                   & & std &  113.84 & 126.77 &  150.14 &  162.72 &   217.34 & 275.39 & 383.25 \\
        \cmidrule{2-10}
& \multirow{3}{*}{0.1} & count &  8137 &   5265 &    3790 &    3019 &     1460 &    704 &    281 \\
                   & & size &  2.37 &   3.67 &    5.10 &    6.40 &    13.23 &  27.44 &  68.75 \\
                   & & std &  0.49 &   0.77 &    1.09 &    1.36 &     2.83 &   5.82 &  15.32 \\
        \cmidrule{2-10}
& \multirow{3}{*}{0.2} & count &  8186 &   5227 &    3801 &    3016 &     1451 &    709 &    276 \\
                   & & size &  2.36 &   3.70 &    5.08 &    6.41 &    13.31 &  27.25 &  70.00 \\
                   & & std &  0.48 &   0.79 &    1.08 &    1.37 &     2.85 &   5.87 &  14.48 \\
        \cmidrule{2-10}
& \multirow{3}{*}{0.3} & count &  8167 &   5248 &    3780 &    2994 &     1443 &    700 &    279 \\
                   & & size &  2.37 &   3.68 &    5.11 &    6.45 &    13.39 &  27.60 &  69.24 \\
                   & & std &  0.48 &   0.78 &    1.09 &    1.34 &     2.86 &   6.08 &  14.55 \\
        \cmidrule{2-10}
& \multirow{3}{*}{0.4} & count &  8153 &   5200 &    3780 &    3022 &     1473 &    707 &    280 \\
                   & & size &  2.37 &   3.72 &    5.11 &    6.39 &    13.12 &  27.33 &  69.00 \\
                   & & std &  0.48 &   0.79 &    1.08 &    1.35 &     2.74 &   5.88 &  14.07 \\
        \cmidrule{2-10}
& \multirow{3}{*}{0.5} & count &  8153 &   5154 &    3762 &    2970 &     1433 &    703 &    273 \\
                   & & size &  2.37 &   3.75 &    5.14 &    6.50 &    13.48 &  27.48 &  70.77 \\
                   & & std &  0.49 &   0.79 &    1.09 &    1.37 &     2.79 &   5.82 &  14.88 \\
        \cmidrule{2-10}
& \multirow{3}{*}{0.6 - 0.9} & count &  8146 &   5153 &    3761 &    2970 &     1433 &    703 &    273 \\
                   & & size &  2.37 &   3.75 &    5.14 &    6.50 &    13.48 &  27.48 &  70.77 \\
                   & & std &  0.49 &   0.79 &    1.09 &    1.37 &     2.79 &   5.78 &  14.63 \\
        \cmidrule{2-10}
& \multirow{3}{*}{1} & count &  8080 &   5128 &    3749 &    2964 &     1431 &    703 &    273 \\
                   & & size &  2.39 &   3.77 &    5.15 &    6.52 &    13.50 &  27.48 &  70.77 \\
                   & & std &  0.51 &   0.80 &    1.10 &    1.38 &     2.80 &   5.78 &  14.63 \\
\bottomrule
\end{tabular}
\end{table*}

\begin{table*}[p]
\centering
\caption{Statistics on resulting partitions for Hotel Reviews Dataset considering all entity types.}
\label{tab:part_stat_hotels_all}
\begin{tabular}{lllrrrrrrr}
\midrule[\heavyrulewidth]
                &   &    k &    2  &     3  &     4  &     5  &     10 &      20 &      50 \\
         & $\lambda$ &  &  &  & & & & & \\
        \midrule[\heavyrulewidth]
\parbox[t]{2mm}{\multirow{3}{*}{\rotatebox[origin=c]{90}{\textbf{GDF}}}} & \multirow{3}{*}{-} & count &  523 &   272 &   163 &   127 &     43 &     16 &     1 \\
                   & & size &  2.82 &  5.42 &  9.05 & 11.61 &  34.30 &  92.19 &  1475 \\
                   & & std & 3.45 & 11.51 & 24.56 & 35.47 & 112.32 & 264.30 &     0 \\
        \midrule[\heavyrulewidth]
\parbox[t]{2mm}{\multirow{23}{*}{\rotatebox[origin=c]{90}{\textbf{Mondrian}}}} & \multirow{3}{*}{0} & count &  645 &   398 &   293 &   226 &    117 &     54 &    21 \\
                   & & size &  2.29 &  3.71 &  5.03 &  6.53 &  12.61 &  27.31 & 70.24 \\
                   & & std &  0.47 &  0.77 &  1.11 &  1.45 &   2.60 &   5.59 & 15.67 \\
        \cmidrule{2-10}
& \multirow{3}{*}{0.1} & count &  635 &   401 &   294 &   231 &    115 &     56 &    22 \\
                   & & size &  2.32 &  3.68 &  5.02 &  6.39 &  12.83 &  26.34 & 67.05 \\
                   & & std &  0.47 &  0.76 &  1.04 &  1.39 &   2.78 &   5.22 & 14.78 \\
        \cmidrule{2-10}
& \multirow{3}{*}{0.2} & count &  641 &   404 &   295 &   224 &    112 &     51 &    20 \\
                   & & size &  2.30 &  3.65 &     5 &  6.58 &  13.17 &  28.92 & 73.75 \\
                   & & std &  0.46 &  0.75 &  1.08 &  1.41 &   2.71 &   5.03 & 15.00 \\
        \cmidrule{2-10}
& \multirow{3}{*}{0.3} & count &  645 &   404 &   285 &   212 &    106 &     50 &    24 \\
                   & & size &  2.29 &  3.65 &  5.18 &  6.96 &  13.92 &  29.50 & 61.46 \\
                   & & std &  0.45 &  0.79 &  1.03 &  1.41 &   2.43 &   3.65 &  6.47 \\
        \cmidrule{2-10}
& \multirow{3}{*}{0.4} & count &  610 &   406 &   268 &   209 &    113 &     48 &    20 \\
                   & & size &  2.42 &  3.63 &  5.50 &  7.06 &  13.05 &  30.73 & 73.75 \\
                   & & std &  0.49 &  0.84 &  0.85 &  1.76 &   2.32 &   6.71 &  6.45 \\
        \cmidrule{2-10}
& \multirow{3}{*}{0.5} & count &  558 &   415 &   256 &   255 &    128 &     64 &    20 \\
                   & & size &  2.64 &  3.55 &  5.76 &  5.78 &  11.52 &  23.05 & 73.75 \\
                   & & std &  0.48 &  0.84 &  0.70 &  0.72 &   1.19 &   2.22 & 19.13 \\
        \cmidrule{2-10}
& \multirow{3}{*}{0.6 - 1} & count &  554 &   417 &   256 &   255 &    128 &     64 &    20 \\
                   & & size &  2.66 &  3.54 &  5.76 &  5.78 &  11.52 &  23.05 & 73.75 \\
                   & & std &  0.47 &  0.83 &  0.70 &  0.71 &   1.19 &   2.22 & 19.13 \\
\bottomrule
\end{tabular}
\end{table*}

\begin{table*}[p]
\centering
\caption{Statistics on resulting partitions for Hotel Reviews Dataset considering only GPE entities.}
\label{tab:part_stat_hotels_gpe}
\begin{tabular}{lllrrrrrrr}
\midrule[\heavyrulewidth]
                &   &    k &    2  &     3  &     4  &     5  &     10 &      20 &      50 \\
         & $\lambda$ &  &  &  & & & & & \\
        \midrule[\heavyrulewidth]
\parbox[t]{2mm}{\multirow{3}{*}{\rotatebox[origin=c]{90}{\textbf{GDF}}}} & \multirow{3}{*}{-} & count &  303 &   107 &     47 &     15 &      1 &      1 &      1 \\
                   & & size &  4.87 & 13.79 &  31.38 &  98.33 &   1475 &   1475 &   1475 \\
                   & & std & 18.59 & 90.92 & 179.98 & 358.71 &      0 &      0 &      0 \\
        \midrule[\heavyrulewidth]
\parbox[t]{2mm}{\multirow{20}{*}{\rotatebox[origin=c]{90}{\textbf{Mondrian}}}} & \multirow{3}{*}{0} & count &  74 &    49 &     35 &     32 &     17 &     13 &      7 \\
                   & & size &  19.93 & 30.10 &  42.14 &  46.09 &  86.76 & 113.46 & 210.71 \\
                   & & std &  61.71 & 74.77 &  86.30 &  89.59 & 112.62 & 121.77 & 135.59 \\
        \cmidrule{2-10}
& \multirow{3}{*}{0.1} & count &  603 &   408 &    276 &    236 &    117 &     59 &     18 \\
                   & & size &  2.45 &  3.62 &   5.34 &   6.25 &  12.61 &     25 &  81.94 \\
                   & & std &  0.50 &  0.72 &   1.10 &   1.06 &   1.82 &   3.03 &  13.70 \\
        \cmidrule{2-10}
& \multirow{3}{*}{0.2} & count &  647 &   380 &    318 &    201 &     95 &     45 &     21 \\
                   & & size &  2.28 &  3.88 &   4.64 &   7.34 &  15.53 &  32.78 &  70.24 \\
                   & & std &  0.45 &  0.70 &   0.89 &   1.45 &   2.70 &   4.46 &   8.09 \\
        \cmidrule{2-10}
& \multirow{3}{*}{0.3} & count &  634 &   376 &    312 &    211 &    101 &     50 &     21 \\
                   & & size &  2.33 &  3.92 &   4.73 &   6.99 &  14.60 &  29.50 &  70.24 \\
                   & & std &  0.47 &  0.75 &   0.89 &   1.55 &   2.91 &   5.75 &   7.42 \\
        \cmidrule{2-10}
& \multirow{3}{*}{0.4} & count &  657 &   362 &    327 &    202 &     99 &     50 &     21 \\
                   & & size &  2.25 &  4.07 &   4.51 &   7.30 &  14.90 &  29.50 &  70.24 \\
                   & & std &  0.43 &  0.66 &   0.75 &   1.51 &   2.82 &   5.75 &   7.42 \\
        \cmidrule{2-10}
& \multirow{3}{*}{0.5 - 1} & count &  554 &   417 &    256 &    255 &    128 &     64 &     20 \\
                   & & size &  2.66 &  3.54 &   5.76 &   5.78 &  11.52 &  23.05 &  73.75 \\
                   & & std &  0.47 &  0.83 &   0.70 &   0.71 &   1.19 &   2.22 &  19.13 \\
\bottomrule
\end{tabular}
\end{table*}

\subsection{Information Loss}
\label{appendix:information_loss}

In addition to evaluating resulting partitions, we are also interested in the actual information loss which is introduced by anonymizing a given dataset. Figure~\ref{fig:il_blogs} provides an overview of information loss for experiments run on the Blog Authorship Corpus. In particular, Figure~\ref{fig:ril_blogs_all} visualizes relational and Figure~\ref{fig:xil_blogs_all} textual information loss for experiments considering all entity types. Similarly, Figure~\ref{fig:ril_blogs_gpe} and Figure~\ref{fig:xil_blogs_gpe} provide an overview of information loss if only GPE entities are considered. Figure~\ref{fig:il_hotels} provides the same statistics for the Hotel Reviews Dataset.

Moreover, Figures~\ref{fig:xil_blogs_all_zoomed} and \ref{fig:xil_blogs_gpe_zoomed} provide a zoomed version of Figure~\ref{fig:xil_blogs_all} and Figure~\ref{fig:xil_blogs_gpe} showing textual information loss for experiments on the Blog Authorship Corpus. 
Similarly, we visualize zoomed textual information loss for the Hotel Reviews Dataset in Figure~\ref{fig:xil_hotels_all_zoomed} and Figure~\ref{fig:xil_hotels_gpe_zoomed}.

\begin{figure*}[p]
    \centering
    \resizebox{\linewidth}{!}{
    \input{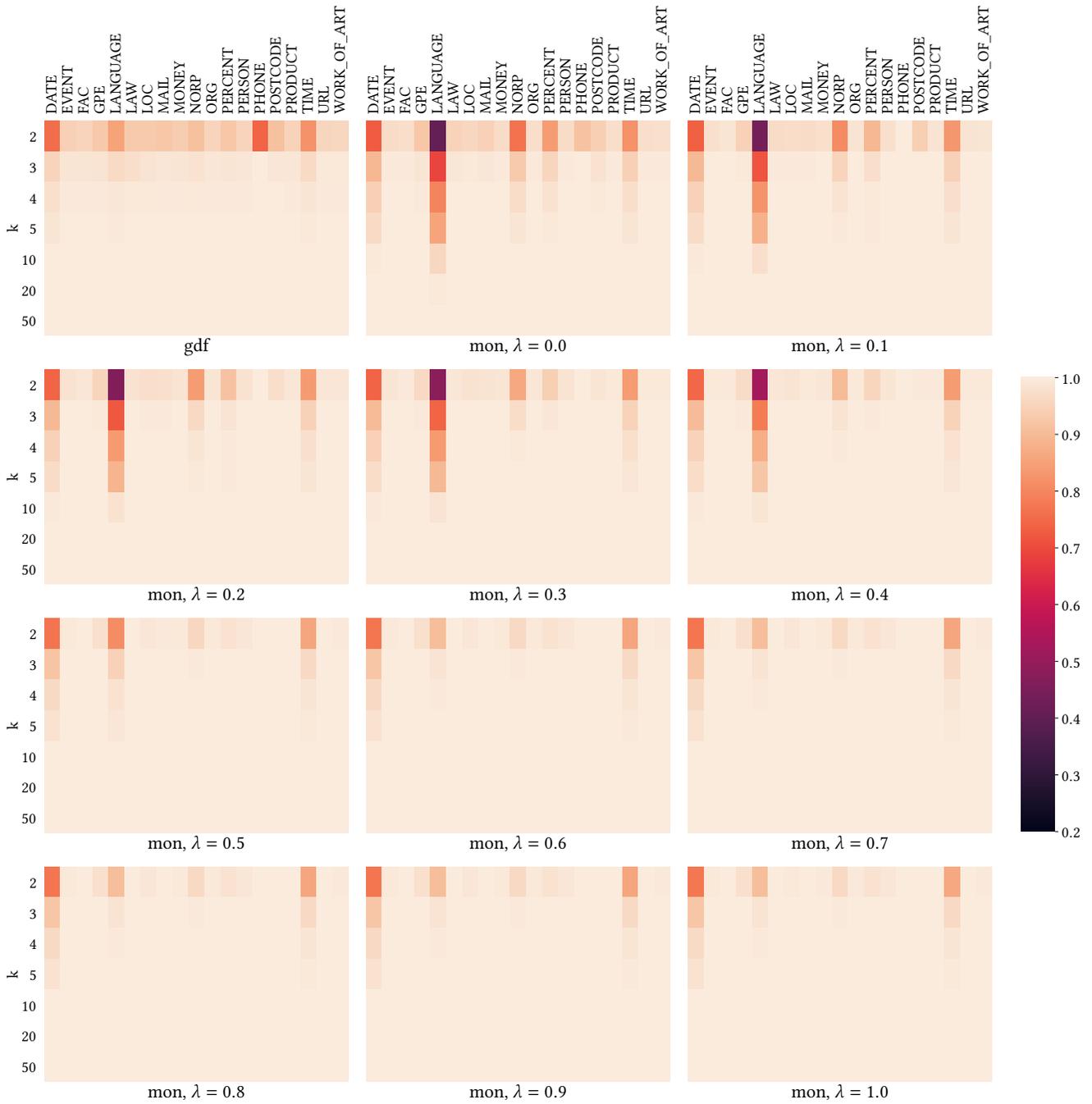}
    }
    \caption[Textual information loss of the attribute text per entity type for experiments run on the Blog Authorship Corpus.]{Textual information loss of the attribute \textit{text} per entity types for experiments run on the Blog Authorship Corpus. Information loss is visualized for GDF and Mondrian partitioning with varying $\mathbf{\bm{\lambda}}$.}
    \label{fig:heatmap_text}
\end{figure*}

In addition to high-level charts on information loss, Figure~\ref{fig:heatmap_text} provides a detailed analysis of information loss per entity type for the attribute \textit{text} in the Blog Authorship Corpus. Additionally, Figure~\ref{fig:heatmap_negative_review} and Figure~\ref{fig:heatmap_positive_review} visualize the textual information loss per entity type for the attributes \textit{negative review} and \textit{positive review} respectively.

\begin{figure*}[p]
    \centering
    \subfloat[][Relational information loss $\mathbf{NCP_A}$, all entity types]{
    \resizebox{0.45\linewidth}{!}{
    \input{images/ril_blogs_all.pgf}
    \label{fig:ril_blogs_all}
    }
    }
    \hfill
    \subfloat[][Textual information loss $\mathbf{NCP_X}$, all entity types]{
    \resizebox{0.45\linewidth}{!}{
    \input{images/xil_blogs_all.pgf}
    \label{fig:xil_blogs_all}
    }
    }
    
    \subfloat[][Relational information loss $\mathbf{NCP_A}$, only GPE entities]{
    \resizebox{0.45\linewidth}{!}{
    \input{images/ril_blogs_gpe.pgf}
    \label{fig:ril_blogs_gpe}
    }
    }
    \hfill
    \subfloat[][Textual information loss $\mathbf{NCP_X}$, only GPE entities]{
    \resizebox{0.45\linewidth}{!}{
    \input{images/xil_blogs_gpe.pgf}
    \label{fig:xil_blogs_gpe}
    }
    }
    \caption[Relational and textual information loss for experiments run on the Blog Authorship Corpus.]{Relational and textual information loss for experiments run on the Blog Authorship Corpus. Results for relational \protect\subref{fig:ril_blogs_all} and textual information loss \protect\subref{fig:xil_blogs_all} for experiments considering all entities. Results for relational \protect\subref{fig:ril_blogs_gpe} and textual information loss \protect\subref{fig:xil_blogs_gpe} for experiments considering only GPE entities.}
    \label{fig:il_blogs}
\end{figure*}

\begin{figure*}[p]
    \centering
    \resizebox{0.6\linewidth}{!}{
    \input{images/xil_blogs_all_zoomed.pgf}
    }
    \caption[Zoomed textual information loss for experiments run on the Blog Authorship Corpus considering all entities.]{Zoomed textual information loss for experiments run on the Blog Authorship Corpus considering all entities.}
    \label{fig:xil_blogs_all_zoomed}
\end{figure*}

\begin{figure*}[p]
    \centering
    \resizebox{0.6\linewidth}{!}{
    \input{images/xil_blogs_gpe_zoomed.pgf}
    }
    \caption[Zoomed textual information loss for experiments run on the Blog Authorship Corpus considering only GPE entities.]{Zoomed textual information loss for experiments run on the Blog Authorship Corpus considering only GPE entities.}
    \label{fig:xil_blogs_gpe_zoomed}
\end{figure*}

\begin{figure*}[p]
    \centering
    \subfloat[][Relational information loss $\mathbf{NCP_A}$, all entity types]{
    \resizebox{0.45\linewidth}{!}{
    \input{images/ril_hotels_all.pgf}
    \label{fig:ril_hotels_all}
    }
    }
    \hfill
    \subfloat[][Textual information loss $\mathbf{NCP_X}$, all entity types]{
    \resizebox{0.45\linewidth}{!}{
    \input{images/xil_hotels_all.pgf}
    \label{fig:xil_hotels_all}
    }
    }
    
    \subfloat[][Relational information loss $\mathbf{NCP_A}$, only GPE entities]{
    \resizebox{0.45\linewidth}{!}{
    \input{images/ril_hotels_gpe.pgf}
    \label{fig:ril_hotels_gpe}
    }
    }
    \hfill
    \subfloat[][Textual information loss $\mathbf{NCP_X}$, only GPE entities]{
    \resizebox{0.45\linewidth}{!}{
    \input{images/xil_hotels_gpe.pgf}
    \label{fig:xil_hotels_gpe}
    }
    }
    \caption[Relational and textual information loss for experiments run on the Hotel Reviews Dataset.]{Relational and textual information loss for experiments run on the Hotel Reviews Dataset. Results for relational \protect\subref{fig:ril_hotels_all} and textual information loss \protect\subref{fig:xil_hotels_all} for experiments considering all entities. Results for relational \protect\subref{fig:ril_hotels_gpe} and textual information loss \protect\subref{fig:xil_hotels_gpe} for experiments considering only GPE entities.}
    \label{fig:il_hotels}
\end{figure*}

\begin{figure*}[p]
    \centering
    \resizebox{0.6\linewidth}{!}{
    \input{images/xil_hotels_all_zoomed.pgf}
    }
    \caption[Zoomed textual information loss for experiments run on the Hotel Reviews Dataset considering all entities.]{Zoomed textual information loss for experiments run on the Hotel Reviews Dataset considering all entities.}
    \label{fig:xil_hotels_all_zoomed}
\end{figure*}

\begin{figure*}[p]
    \centering
    \resizebox{0.6\linewidth}{!}{
    \input{images/xil_hotels_gpe_zoomed.pgf}
    }
    \caption[Zoomed textual information loss for experiments run on the Hotel Reviews Dataset considering only GPE entities.]{Zoomed textual information loss for experiments run on the Hotel Reviews Dataset considering only GPE entities.}
    \label{fig:xil_hotels_gpe_zoomed}
\end{figure*}

\begin{figure*}[p]
    \centering
    \resizebox{\linewidth}{!}{
    \input{images/heatmap_negative_review.pgf}
    }
    \caption[Textual information loss of the attribute negative review per entity type for experiments run on the Hotel Reviews Dataset.]{Textual information loss of the attribute \textit{negative review} per entity type for experiments run on the Hotel Reviews Dataset. Information loss is visualized for GDF and Mondrian partitioning with varying $\mathbf{\bm{\lambda}}$.}
    \label{fig:heatmap_negative_review}
\end{figure*}

\begin{figure*}[p]
    \centering
    \resizebox{\linewidth}{!}{
    \input{images/heatmap_positive_review.pgf}
    }
    \caption[Textual information loss of the attribute positive review per entity type for experiments run on the Hotel Reviews Dataset.]{Textual information loss of the attribute \textit{positive review} per entity type for experiments run on the Hotel Reviews Dataset. Information loss is visualized for GDF and Mondrian partitioning with varying $\mathbf{\bm{\lambda}}$.}
    \label{fig:heatmap_positive_review}
\end{figure*}

\extended{
\subsection{Directions of Extending the \systemname{} Framework}
\label{appendix:future-work}

As a framework approach, \systemname enables several paths for future work. 
These include all aspects of the anonymization pipeline, namely the partitioning, string matching, recording, privacy model, and supported entity types. 
We provide examples below.

\paragraph{Partitioning}
We showed how decisions on partitions significantly influence information loss. 
\extended{While the naive partitioning strategy GDF can deal with sparse, but diverse sets of sensitive terms, there might be partitioning strategies better suited to attributes with such properties. }
It would be interesting to see if clustering algorithms applied on sensitive terms lead to improved partitioning. 
Such clustering algorithms require a minimum lower bound on the partition sizes of at least $k$. 
\citet{Abu-Khzam2018} presents a general framework for clustering algorithms with a lower bound on the cluster size.

\paragraph{String matching}
The matching of relational and textual attributes is currently using an exact string match.
Another interesting research topic to build on our work is to investigate sophisticated methods to find non-trivial links within the dataset. 
Non-trivial links are links which cannot be detected using simple string matching. 
Mechanisms to reveal non-trivial links are discussed by \citet{Hassanzadeh2009}. 
They studied approximations on string matching as well as semantic mechanisms based on ontology and created a declarative framework and specification language to resolve links in relational data. 
Those mechanisms would also be applicable to find links between relational data and sensitive entities.
It would also be interesting to use string matching based on models using word embeddings~\cite{DBLP:conf/nips/MikolovSCCD13} or transformer-based similarity functions~\cite{bert}.

\paragraph{Recoding}
Moreover, our current recoding strategy for sensitive terms in texts uses suppression to generate a $k$-anonymous version of texts. 
However, suppression tends to introduce more information loss compared to generalization. 
Therefore, it would be interesting to introduce an automatic generalization mechanism for sensitive terms and evaluate it. 
One way to automatically generate \acp{DGH} for sensitive terms is to use hypernym-trees as discussed by \citet{Lee2008} and used by \citet{Anandan2012} to anonymize texts.

\paragraph{Privacy model}
We used \kanonymity{} as the privacy model to prevent identity disclosure. 
Even though \kanonymity{} establishes guarantees on privacy, it does not guard against attacks where adversaries have access to background knowledge. 
Differential privacy introduced by \citet{Dwork2006} resists such attacks by adding noise to data.
Our \systemname{} can be extended by using such an alternative privacy model. 
An interesting question to answer would be how differential private methods defined on relational data could be combined with work on creating a differential private representation of texts~\cite{Xu2020, Fernandes2018}.

\paragraph{Entity types}
For the recognition of sensitive entities, we chose to use spaCy and its entity types trained on the OntoNotes5 corpus. 
We chose to use the OntoNotes5 entity types scheme since it provides more distinguishment and therefore more semantics to entities compared to WikiNer annotations. 
However, there are still cases, where more fine-grained entity recognition will reduce false positive matches. One example is the term ``Georgia'' which can refer to the country in the Caucasus, the U.S. state, or to a city in Indiana. 
\citet{Ling2012} presents a fine-grained set of 112 entity-types which would cover the explained example and state that a fine-grained entity recognition system benefits from its accuracy.

}

\section{Extended Related Work}
We present related work for anonymization of other types of data. 
\extended{Moreover, we present an overview of the existing regulations on anonymization and their view on \ac{PII}. 
Finally, we present a non-exhaustive overview of anonymization tools and frameworks available.}

\subsection{Research in Data Anonymization for Audio, Images, and Video}

Even though this work only focuses on structured data and free text, recent work on anonymization of other forms of data is worth mentioning. 
For de-identification of images showing faces, \citet{Gross2006} highlighted that pixelation and blurring offers poor privacy and suggested a model-based approach to protect privacy while preserving data utility. 
In contrast, recent work by \citet{Embedding2019} applied methods from machine learning by implementing a simple \ac{GAN} to generate new faces to preserve privacy while retaining original data distribution.

For audio data, recent work focused either on anonymization of the speaker's identity or the speech content. 
\citet{Justin2015} suggested a framework which automatically transfers speech into a de-identified version using different acoustical models for recognition and synthesis. Moreover, \citet{Cohn2019} investigated the task of de-identifying spoken text by first using \ac{ASR} to transcribe texts, then extracting entities using \ac{NER}, and finally aligning text elements to the audio and suppressing audio segments which should be de-identified.

Additionally, recent work by \citet{Agrawal2011} showed that de-identification of people can also be applied to whole bodies within videos whereas \citet{Gafni2019} focused on live de-identification of faces in video streams.

Finally, \citet{McDonald2012} developed a framework for obfuscating writing styles which can be used by authors to prevent stylometry attacks to retrieve their identities. When it comes to unstructured text, their approach anonymizes writing styles in text documents by analyzing stylographic properties, determining features to be changed, ranking those features with respect to their clusters, and suggesting those changes to the user.

\extended{

\subsection{What is considered \acl{PII}?}
\label{app:identifiers}
In order to understand what fields should be anonymized, a common understanding on what \acf{PII} is needs to be established. Therefore, we provide a broad overview on regulations such as the \acf{HIPAA}, the \acf{GDPR}, and definitions by \ac{NIST} to get an understanding for \ac{PII}.

\subsubsection{\acl{HIPAA}}
First, we want to consider the \acf{HIPAA} providing regulations to ensure privacy within medical data in the USA~\cite{HIPAA}. Even though the \ac{HIPAA} privacy rule uses the terminology \acf{PHI}, in general we can  transfer their identifiers to the domain of \ac{PII}. The \ac{HIPAA} states that any information from the past, present, or future which is linked to an individual is considered \ac{PHI}. In addition to domain experts defining \ac{PHI}, the Safe Harbor Method defined in the \ac{HIPAA} provides an overview of attributes which should be anonymized by removing~\cite{HIPAA}. Those attributes are in particular:
\begin{enumerate}
\item Names
\item Geographic entities smaller than states (street address, city, county, ZIP, etc.)
\item Dates (except year)
\item Phone numbers
\item Vehicle identifiers and serial numbers
\item Fax numbers
\item Device identifiers and serial numbers
\item Email addresses
\item URLs
\item Social security numbers
\item IP addresses
\item Medical record numbers
\item Biometric identifiers, including finger and voice prints
\item Health plan beneficiary numbers
\item Full-face photographs
\item Account numbers
\item Any other unique identifying number, characteristic, code, etc.
\item Certificate and license numbers
\end{enumerate}

\subsubsection{\acl{GDPR}}
In Europe, one important privacy regulation is the \acf{GDPR}~\cite{GDPR}. Instead of using the term \ac{PII}, the \ac{GDPR} refers to the term \textit{personal data}. The regulation states that \textit{"‘Personal data’ means any information relating to an identified or identifiable natural person ..."}~\cite{GDPR}. Even though the \ac{GDPR} does not explicitly state a list of attributes considered personal data, they provide some guidance on which properties are considered personal data. In particular the GDPR states that personal data is any data which can identify an individual directly or indirectly \textit{"by reference to an identifier such as a name, an identification number, location data, an online identifier or to one or more factors specific to the physical, physiological, genetic, mental, economic, cultural or social identity of that natural person"}~\cite{GDPR}.

\subsubsection{Guidelines by \acs{NIST}}
In contrast to the \ac{GDPR}, the \acf{NIST} provides guidance on protecting \ac{PII}~\cite{McCallister2010}. The \ac{NIST} distinguishes \ac{PII} in two categories. The first category includes "... any information that can be used to distinguish or trace an individual‘s identity ..."~\cite{McCallister2010}. In particular, they list the following attributes:
\begin{itemize}
    \item Name
    \item Social Security Number
    \item Date and place of birth
    \item Mother's maiden name
    \item Biometric records
\end{itemize}

Moreover, the \ac{NIST} labels "... any other information that is linked or linkable to an individual ..." also as \ac{PII}~\cite{McCallister2010}. Examples for linked or linkable attributes are:
\begin{itemize}
    \item Medical information
    \item Educational information
    \item Financial information
    \item Employment information
\end{itemize}

\subsection{Existing Anonymization Tools and Frameworks}
Multiple publicly available tools and frameworks for anonymization of data have been released.
\textit{ARX}\footnote{\url{https://arx.deidentifier.org/}} is an open source comprehensive software providing a graphical interface for anonymizing structured datasets~\cite{Prasser2014,Prasser2020}. ARX supports multiple privacy and risk models, methods for transforming data, and concepts for analyzing the output data. Among the privacy models, it supports syntactic privacy models like \kanonymity{}, \ldiversity{}, and \tcloseness{}, but also supports semantic privacy models like \edifprivacy{}.
Moreover, \textit{Amnesia}\footnote{\url{https://amnesia.openaire.eu/}} is a flexible data anonymization tool which allows to ensure privacy on structured data. Amnesia supports \kanonymity{} for relational data as well as \kmanonymity{} for datasets containing set-valued data fields. 
Finally, \textit{Privacy Analytics}\footnote{\url{https://privacy-analytics.com/health-data-privacy/}} offers a commercial Eclipse plugin which can be used to anonymize structured data.

Besides toolings for de-identification of structured data, there also exist frameworks or modules to achieve anonymization. \textit{python-datafly}\footnote{\url{https://github.com/alessiovierti/python-datafly}} is a Python implementation of the Datafly algorithm introduced by \citet{Sweeney2002} as one of the first algorithms to transfer structured data to match \kanonymity{}.
Additionally, \textit{Crowds}\footnote{\url{https://github.com/leo-mazz/crowds}} is an open-source python module developed to de-identify a dataframe using the \ac{OLA} algorithm as proposed by \citet{ElEmam2009} to achieve \kanonymity{}.
Finally, an example for an implementation of the Mondrian algorithm~\cite{LeFevre2006} is available for Python\footnote{\url{https://github.com/Nuclearstar/K-Anonymity}} to show how \kanonymity{}, \ldiversity{}, and \tcloseness{} can be used as privacy models.

There are multiple tools and frameworks for de-identification of free text. \textit{NLM-Scrubber}\footnote{\url{https://scrubber.nlm.nih.gov/}} is a freely available tool for de-identification of clinical texts according to the Safe Harbor Method introduced in the \ac{HIPAA} Privacy Rule. 
Moreover, \textit{\ac{MIST}}\footnote{\url{http://mist-deid.sourceforge.net/}} is a suite of tools for identifying and redacting \ac{PII} in free-text medical records~\cite{Kayaalp2014}.
\textit{deid}\footnote{\url{https://www.physionet.org/content/deid/1.1/}} is a tool which allows anonymization of free texts within the medical domain.
Finally, \textit{deidentify}\footnote{\url{https://github.com/nedap/deidentify}} is a Python library developed especially for de-identification of medical records and comparison of rule-, feature-, and deep-learning-based approaches for de-identification of free texts~\cite{Trienes2020}.

}

\end{appendices}

\end{document}

%% file: images/splits_blogs_all_5_shrinked.pgf
\begingroup%
\makeatletter%
\begin{pgfpicture}%
\pgfpathrectangle{\pgfpointorigin}{\pgfqpoint{8.000000in}{4.000000in}}%
\pgfusepath{use as bounding box, clip}%
\begin{pgfscope}%
\pgfsetbuttcap%
\pgfsetmiterjoin%
\definecolor{currentfill}{rgb}{1.000000,1.000000,1.000000}%
\pgfsetfillcolor{currentfill}%
\pgfsetlinewidth{0.000000pt}%
\definecolor{currentstroke}{rgb}{1.000000,1.000000,1.000000}%
\pgfsetstrokecolor{currentstroke}%
\pgfsetdash{}{0pt}%
\pgfpathmoveto{\pgfqpoint{0.000000in}{0.000000in}}%
\pgfpathlineto{\pgfqpoint{8.000000in}{0.000000in}}%
\pgfpathlineto{\pgfqpoint{8.000000in}{4.000000in}}%
\pgfpathlineto{\pgfqpoint{0.000000in}{4.000000in}}%
\pgfpathclose%
\pgfusepath{fill}%
\end{pgfscope}%
\begin{pgfscope}%
\pgfsetbuttcap%
\pgfsetmiterjoin%
\definecolor{currentfill}{rgb}{1.000000,1.000000,1.000000}%
\pgfsetfillcolor{currentfill}%
\pgfsetlinewidth{0.000000pt}%
\definecolor{currentstroke}{rgb}{0.000000,0.000000,0.000000}%
\pgfsetstrokecolor{currentstroke}%
\pgfsetstrokeopacity{0.000000}%
\pgfsetdash{}{0pt}%
\pgfpathmoveto{\pgfqpoint{1.023125in}{1.668889in}}%
\pgfpathlineto{\pgfqpoint{7.760000in}{1.668889in}}%
\pgfpathlineto{\pgfqpoint{7.760000in}{3.760000in}}%
\pgfpathlineto{\pgfqpoint{1.023125in}{3.760000in}}%
\pgfpathclose%
\pgfusepath{fill}%
\end{pgfscope}%
\begin{pgfscope}%
\pgfpathrectangle{\pgfqpoint{1.023125in}{1.668889in}}{\pgfqpoint{6.736875in}{2.091111in}}%
\pgfusepath{clip}%
\pgfsetbuttcap%
\pgfsetmiterjoin%
\definecolor{currentfill}{rgb}{0.121569,0.466667,0.705882}%
\pgfsetfillcolor{currentfill}%
\pgfsetlinewidth{0.000000pt}%
\definecolor{currentstroke}{rgb}{0.000000,0.000000,0.000000}%
\pgfsetstrokecolor{currentstroke}%
\pgfsetstrokeopacity{0.000000}%
\pgfsetdash{}{0pt}%
\pgfpathmoveto{\pgfqpoint{1.176236in}{1.668889in}}%
\pgfpathlineto{\pgfqpoint{1.482458in}{1.668889in}}%
\pgfpathlineto{\pgfqpoint{1.482458in}{1.668889in}}%
\pgfpathlineto{\pgfqpoint{1.176236in}{1.668889in}}%
\pgfpathclose%
\pgfusepath{fill}%
\end{pgfscope}%
\begin{pgfscope}%
\pgfpathrectangle{\pgfqpoint{1.023125in}{1.668889in}}{\pgfqpoint{6.736875in}{2.091111in}}%
\pgfusepath{clip}%
\pgfsetbuttcap%
\pgfsetmiterjoin%
\definecolor{currentfill}{rgb}{0.121569,0.466667,0.705882}%
\pgfsetfillcolor{currentfill}%
\pgfsetlinewidth{0.000000pt}%
\definecolor{currentstroke}{rgb}{0.000000,0.000000,0.000000}%
\pgfsetstrokecolor{currentstroke}%
\pgfsetstrokeopacity{0.000000}%
\pgfsetdash{}{0pt}%
\pgfpathmoveto{\pgfqpoint{1.788679in}{1.668889in}}%
\pgfpathlineto{\pgfqpoint{2.094901in}{1.668889in}}%
\pgfpathlineto{\pgfqpoint{2.094901in}{1.793113in}}%
\pgfpathlineto{\pgfqpoint{1.788679in}{1.793113in}}%
\pgfpathclose%
\pgfusepath{fill}%
\end{pgfscope}%
\begin{pgfscope}%
\pgfpathrectangle{\pgfqpoint{1.023125in}{1.668889in}}{\pgfqpoint{6.736875in}{2.091111in}}%
\pgfusepath{clip}%
\pgfsetbuttcap%
\pgfsetmiterjoin%
\definecolor{currentfill}{rgb}{0.121569,0.466667,0.705882}%
\pgfsetfillcolor{currentfill}%
\pgfsetlinewidth{0.000000pt}%
\definecolor{currentstroke}{rgb}{0.000000,0.000000,0.000000}%
\pgfsetstrokecolor{currentstroke}%
\pgfsetstrokeopacity{0.000000}%
\pgfsetdash{}{0pt}%
\pgfpathmoveto{\pgfqpoint{2.401122in}{1.668889in}}%
\pgfpathlineto{\pgfqpoint{2.707344in}{1.668889in}}%
\pgfpathlineto{\pgfqpoint{2.707344in}{1.831235in}}%
\pgfpathlineto{\pgfqpoint{2.401122in}{1.831235in}}%
\pgfpathclose%
\pgfusepath{fill}%
\end{pgfscope}%
\begin{pgfscope}%
\pgfpathrectangle{\pgfqpoint{1.023125in}{1.668889in}}{\pgfqpoint{6.736875in}{2.091111in}}%
\pgfusepath{clip}%
\pgfsetbuttcap%
\pgfsetmiterjoin%
\definecolor{currentfill}{rgb}{0.121569,0.466667,0.705882}%
\pgfsetfillcolor{currentfill}%
\pgfsetlinewidth{0.000000pt}%
\definecolor{currentstroke}{rgb}{0.000000,0.000000,0.000000}%
\pgfsetstrokecolor{currentstroke}%
\pgfsetstrokeopacity{0.000000}%
\pgfsetdash{}{0pt}%
\pgfpathmoveto{\pgfqpoint{3.013565in}{1.668889in}}%
\pgfpathlineto{\pgfqpoint{3.319787in}{1.668889in}}%
\pgfpathlineto{\pgfqpoint{3.319787in}{1.925882in}}%
\pgfpathlineto{\pgfqpoint{3.013565in}{1.925882in}}%
\pgfpathclose%
\pgfusepath{fill}%
\end{pgfscope}%
\begin{pgfscope}%
\pgfpathrectangle{\pgfqpoint{1.023125in}{1.668889in}}{\pgfqpoint{6.736875in}{2.091111in}}%
\pgfusepath{clip}%
\pgfsetbuttcap%
\pgfsetmiterjoin%
\definecolor{currentfill}{rgb}{0.121569,0.466667,0.705882}%
\pgfsetfillcolor{currentfill}%
\pgfsetlinewidth{0.000000pt}%
\definecolor{currentstroke}{rgb}{0.000000,0.000000,0.000000}%
\pgfsetstrokecolor{currentstroke}%
\pgfsetstrokeopacity{0.000000}%
\pgfsetdash{}{0pt}%
\pgfpathmoveto{\pgfqpoint{3.626009in}{1.668889in}}%
\pgfpathlineto{\pgfqpoint{3.932230in}{1.668889in}}%
\pgfpathlineto{\pgfqpoint{3.932230in}{2.407663in}}%
\pgfpathlineto{\pgfqpoint{3.626009in}{2.407663in}}%
\pgfpathclose%
\pgfusepath{fill}%
\end{pgfscope}%
\begin{pgfscope}%
\pgfpathrectangle{\pgfqpoint{1.023125in}{1.668889in}}{\pgfqpoint{6.736875in}{2.091111in}}%
\pgfusepath{clip}%
\pgfsetbuttcap%
\pgfsetmiterjoin%
\definecolor{currentfill}{rgb}{0.121569,0.466667,0.705882}%
\pgfsetfillcolor{currentfill}%
\pgfsetlinewidth{0.000000pt}%
\definecolor{currentstroke}{rgb}{0.000000,0.000000,0.000000}%
\pgfsetstrokecolor{currentstroke}%
\pgfsetstrokeopacity{0.000000}%
\pgfsetdash{}{0pt}%
\pgfpathmoveto{\pgfqpoint{4.238452in}{1.668889in}}%
\pgfpathlineto{\pgfqpoint{4.544673in}{1.668889in}}%
\pgfpathlineto{\pgfqpoint{4.544673in}{3.551316in}}%
\pgfpathlineto{\pgfqpoint{4.238452in}{3.551316in}}%
\pgfpathclose%
\pgfusepath{fill}%
\end{pgfscope}%
\begin{pgfscope}%
\pgfpathrectangle{\pgfqpoint{1.023125in}{1.668889in}}{\pgfqpoint{6.736875in}{2.091111in}}%
\pgfusepath{clip}%
\pgfsetbuttcap%
\pgfsetmiterjoin%
\definecolor{currentfill}{rgb}{0.121569,0.466667,0.705882}%
\pgfsetfillcolor{currentfill}%
\pgfsetlinewidth{0.000000pt}%
\definecolor{currentstroke}{rgb}{0.000000,0.000000,0.000000}%
\pgfsetstrokecolor{currentstroke}%
\pgfsetstrokeopacity{0.000000}%
\pgfsetdash{}{0pt}%
\pgfpathmoveto{\pgfqpoint{4.850895in}{1.668889in}}%
\pgfpathlineto{\pgfqpoint{5.157117in}{1.668889in}}%
\pgfpathlineto{\pgfqpoint{5.157117in}{3.616386in}}%
\pgfpathlineto{\pgfqpoint{4.850895in}{3.616386in}}%
\pgfpathclose%
\pgfusepath{fill}%
\end{pgfscope}%
\begin{pgfscope}%
\pgfpathrectangle{\pgfqpoint{1.023125in}{1.668889in}}{\pgfqpoint{6.736875in}{2.091111in}}%
\pgfusepath{clip}%
\pgfsetbuttcap%
\pgfsetmiterjoin%
\definecolor{currentfill}{rgb}{0.121569,0.466667,0.705882}%
\pgfsetfillcolor{currentfill}%
\pgfsetlinewidth{0.000000pt}%
\definecolor{currentstroke}{rgb}{0.000000,0.000000,0.000000}%
\pgfsetstrokecolor{currentstroke}%
\pgfsetstrokeopacity{0.000000}%
\pgfsetdash{}{0pt}%
\pgfpathmoveto{\pgfqpoint{5.463338in}{1.668889in}}%
\pgfpathlineto{\pgfqpoint{5.769560in}{1.668889in}}%
\pgfpathlineto{\pgfqpoint{5.769560in}{3.616386in}}%
\pgfpathlineto{\pgfqpoint{5.463338in}{3.616386in}}%
\pgfpathclose%
\pgfusepath{fill}%
\end{pgfscope}%
\begin{pgfscope}%
\pgfpathrectangle{\pgfqpoint{1.023125in}{1.668889in}}{\pgfqpoint{6.736875in}{2.091111in}}%
\pgfusepath{clip}%
\pgfsetbuttcap%
\pgfsetmiterjoin%
\definecolor{currentfill}{rgb}{0.121569,0.466667,0.705882}%
\pgfsetfillcolor{currentfill}%
\pgfsetlinewidth{0.000000pt}%
\definecolor{currentstroke}{rgb}{0.000000,0.000000,0.000000}%
\pgfsetstrokecolor{currentstroke}%
\pgfsetstrokeopacity{0.000000}%
\pgfsetdash{}{0pt}%
\pgfpathmoveto{\pgfqpoint{6.075781in}{1.668889in}}%
\pgfpathlineto{\pgfqpoint{6.382003in}{1.668889in}}%
\pgfpathlineto{\pgfqpoint{6.382003in}{3.616386in}}%
\pgfpathlineto{\pgfqpoint{6.075781in}{3.616386in}}%
\pgfpathclose%
\pgfusepath{fill}%
\end{pgfscope}%
\begin{pgfscope}%
\pgfpathrectangle{\pgfqpoint{1.023125in}{1.668889in}}{\pgfqpoint{6.736875in}{2.091111in}}%
\pgfusepath{clip}%
\pgfsetbuttcap%
\pgfsetmiterjoin%
\definecolor{currentfill}{rgb}{0.121569,0.466667,0.705882}%
\pgfsetfillcolor{currentfill}%
\pgfsetlinewidth{0.000000pt}%
\definecolor{currentstroke}{rgb}{0.000000,0.000000,0.000000}%
\pgfsetstrokecolor{currentstroke}%
\pgfsetstrokeopacity{0.000000}%
\pgfsetdash{}{0pt}%
\pgfpathmoveto{\pgfqpoint{6.688224in}{1.668889in}}%
\pgfpathlineto{\pgfqpoint{6.994446in}{1.668889in}}%
\pgfpathlineto{\pgfqpoint{6.994446in}{3.616386in}}%
\pgfpathlineto{\pgfqpoint{6.688224in}{3.616386in}}%
\pgfpathclose%
\pgfusepath{fill}%
\end{pgfscope}%
\begin{pgfscope}%
\pgfpathrectangle{\pgfqpoint{1.023125in}{1.668889in}}{\pgfqpoint{6.736875in}{2.091111in}}%
\pgfusepath{clip}%
\pgfsetbuttcap%
\pgfsetmiterjoin%
\definecolor{currentfill}{rgb}{0.121569,0.466667,0.705882}%
\pgfsetfillcolor{currentfill}%
\pgfsetlinewidth{0.000000pt}%
\definecolor{currentstroke}{rgb}{0.000000,0.000000,0.000000}%
\pgfsetstrokecolor{currentstroke}%
\pgfsetstrokeopacity{0.000000}%
\pgfsetdash{}{0pt}%
\pgfpathmoveto{\pgfqpoint{7.300668in}{1.668889in}}%
\pgfpathlineto{\pgfqpoint{7.606889in}{1.668889in}}%
\pgfpathlineto{\pgfqpoint{7.606889in}{3.616386in}}%
\pgfpathlineto{\pgfqpoint{7.300668in}{3.616386in}}%
\pgfpathclose%
\pgfusepath{fill}%
\end{pgfscope}%
\begin{pgfscope}%
\pgfpathrectangle{\pgfqpoint{1.023125in}{1.668889in}}{\pgfqpoint{6.736875in}{2.091111in}}%
\pgfusepath{clip}%
\pgfsetbuttcap%
\pgfsetmiterjoin%
\definecolor{currentfill}{rgb}{1.000000,0.498039,0.054902}%
\pgfsetfillcolor{currentfill}%
\pgfsetlinewidth{0.000000pt}%
\definecolor{currentstroke}{rgb}{0.000000,0.000000,0.000000}%
\pgfsetstrokecolor{currentstroke}%
\pgfsetstrokeopacity{0.000000}%
\pgfsetdash{}{0pt}%
\pgfpathmoveto{\pgfqpoint{1.176236in}{1.668889in}}%
\pgfpathlineto{\pgfqpoint{1.482458in}{1.668889in}}%
\pgfpathlineto{\pgfqpoint{1.482458in}{3.620987in}}%
\pgfpathlineto{\pgfqpoint{1.176236in}{3.620987in}}%
\pgfpathclose%
\pgfusepath{fill}%
\end{pgfscope}%
\begin{pgfscope}%
\pgfpathrectangle{\pgfqpoint{1.023125in}{1.668889in}}{\pgfqpoint{6.736875in}{2.091111in}}%
\pgfusepath{clip}%
\pgfsetbuttcap%
\pgfsetmiterjoin%
\definecolor{currentfill}{rgb}{1.000000,0.498039,0.054902}%
\pgfsetfillcolor{currentfill}%
\pgfsetlinewidth{0.000000pt}%
\definecolor{currentstroke}{rgb}{0.000000,0.000000,0.000000}%
\pgfsetstrokecolor{currentstroke}%
\pgfsetstrokeopacity{0.000000}%
\pgfsetdash{}{0pt}%
\pgfpathmoveto{\pgfqpoint{1.788679in}{1.793113in}}%
\pgfpathlineto{\pgfqpoint{2.094901in}{1.793113in}}%
\pgfpathlineto{\pgfqpoint{2.094901in}{3.658451in}}%
\pgfpathlineto{\pgfqpoint{1.788679in}{3.658451in}}%
\pgfpathclose%
\pgfusepath{fill}%
\end{pgfscope}%
\begin{pgfscope}%
\pgfpathrectangle{\pgfqpoint{1.023125in}{1.668889in}}{\pgfqpoint{6.736875in}{2.091111in}}%
\pgfusepath{clip}%
\pgfsetbuttcap%
\pgfsetmiterjoin%
\definecolor{currentfill}{rgb}{1.000000,0.498039,0.054902}%
\pgfsetfillcolor{currentfill}%
\pgfsetlinewidth{0.000000pt}%
\definecolor{currentstroke}{rgb}{0.000000,0.000000,0.000000}%
\pgfsetstrokecolor{currentstroke}%
\pgfsetstrokeopacity{0.000000}%
\pgfsetdash{}{0pt}%
\pgfpathmoveto{\pgfqpoint{2.401122in}{1.831235in}}%
\pgfpathlineto{\pgfqpoint{2.707344in}{1.831235in}}%
\pgfpathlineto{\pgfqpoint{2.707344in}{3.660423in}}%
\pgfpathlineto{\pgfqpoint{2.401122in}{3.660423in}}%
\pgfpathclose%
\pgfusepath{fill}%
\end{pgfscope}%
\begin{pgfscope}%
\pgfpathrectangle{\pgfqpoint{1.023125in}{1.668889in}}{\pgfqpoint{6.736875in}{2.091111in}}%
\pgfusepath{clip}%
\pgfsetbuttcap%
\pgfsetmiterjoin%
\definecolor{currentfill}{rgb}{1.000000,0.498039,0.054902}%
\pgfsetfillcolor{currentfill}%
\pgfsetlinewidth{0.000000pt}%
\definecolor{currentstroke}{rgb}{0.000000,0.000000,0.000000}%
\pgfsetstrokecolor{currentstroke}%
\pgfsetstrokeopacity{0.000000}%
\pgfsetdash{}{0pt}%
\pgfpathmoveto{\pgfqpoint{3.013565in}{1.925882in}}%
\pgfpathlineto{\pgfqpoint{3.319787in}{1.925882in}}%
\pgfpathlineto{\pgfqpoint{3.319787in}{3.655165in}}%
\pgfpathlineto{\pgfqpoint{3.013565in}{3.655165in}}%
\pgfpathclose%
\pgfusepath{fill}%
\end{pgfscope}%
\begin{pgfscope}%
\pgfpathrectangle{\pgfqpoint{1.023125in}{1.668889in}}{\pgfqpoint{6.736875in}{2.091111in}}%
\pgfusepath{clip}%
\pgfsetbuttcap%
\pgfsetmiterjoin%
\definecolor{currentfill}{rgb}{1.000000,0.498039,0.054902}%
\pgfsetfillcolor{currentfill}%
\pgfsetlinewidth{0.000000pt}%
\definecolor{currentstroke}{rgb}{0.000000,0.000000,0.000000}%
\pgfsetstrokecolor{currentstroke}%
\pgfsetstrokeopacity{0.000000}%
\pgfsetdash{}{0pt}%
\pgfpathmoveto{\pgfqpoint{3.626009in}{2.407663in}}%
\pgfpathlineto{\pgfqpoint{3.932230in}{2.407663in}}%
\pgfpathlineto{\pgfqpoint{3.932230in}{3.655822in}}%
\pgfpathlineto{\pgfqpoint{3.626009in}{3.655822in}}%
\pgfpathclose%
\pgfusepath{fill}%
\end{pgfscope}%
\begin{pgfscope}%
\pgfpathrectangle{\pgfqpoint{1.023125in}{1.668889in}}{\pgfqpoint{6.736875in}{2.091111in}}%
\pgfusepath{clip}%
\pgfsetbuttcap%
\pgfsetmiterjoin%
\definecolor{currentfill}{rgb}{1.000000,0.498039,0.054902}%
\pgfsetfillcolor{currentfill}%
\pgfsetlinewidth{0.000000pt}%
\definecolor{currentstroke}{rgb}{0.000000,0.000000,0.000000}%
\pgfsetstrokecolor{currentstroke}%
\pgfsetstrokeopacity{0.000000}%
\pgfsetdash{}{0pt}%
\pgfpathmoveto{\pgfqpoint{4.238452in}{3.551316in}}%
\pgfpathlineto{\pgfqpoint{4.544673in}{3.551316in}}%
\pgfpathlineto{\pgfqpoint{4.544673in}{3.626245in}}%
\pgfpathlineto{\pgfqpoint{4.238452in}{3.626245in}}%
\pgfpathclose%
\pgfusepath{fill}%
\end{pgfscope}%
\begin{pgfscope}%
\pgfpathrectangle{\pgfqpoint{1.023125in}{1.668889in}}{\pgfqpoint{6.736875in}{2.091111in}}%
\pgfusepath{clip}%
\pgfsetbuttcap%
\pgfsetmiterjoin%
\definecolor{currentfill}{rgb}{1.000000,0.498039,0.054902}%
\pgfsetfillcolor{currentfill}%
\pgfsetlinewidth{0.000000pt}%
\definecolor{currentstroke}{rgb}{0.000000,0.000000,0.000000}%
\pgfsetstrokecolor{currentstroke}%
\pgfsetstrokeopacity{0.000000}%
\pgfsetdash{}{0pt}%
\pgfpathmoveto{\pgfqpoint{4.850895in}{3.616386in}}%
\pgfpathlineto{\pgfqpoint{5.157117in}{3.616386in}}%
\pgfpathlineto{\pgfqpoint{5.157117in}{3.631503in}}%
\pgfpathlineto{\pgfqpoint{4.850895in}{3.631503in}}%
\pgfpathclose%
\pgfusepath{fill}%
\end{pgfscope}%
\begin{pgfscope}%
\pgfpathrectangle{\pgfqpoint{1.023125in}{1.668889in}}{\pgfqpoint{6.736875in}{2.091111in}}%
\pgfusepath{clip}%
\pgfsetbuttcap%
\pgfsetmiterjoin%
\definecolor{currentfill}{rgb}{1.000000,0.498039,0.054902}%
\pgfsetfillcolor{currentfill}%
\pgfsetlinewidth{0.000000pt}%
\definecolor{currentstroke}{rgb}{0.000000,0.000000,0.000000}%
\pgfsetstrokecolor{currentstroke}%
\pgfsetstrokeopacity{0.000000}%
\pgfsetdash{}{0pt}%
\pgfpathmoveto{\pgfqpoint{5.463338in}{3.616386in}}%
\pgfpathlineto{\pgfqpoint{5.769560in}{3.616386in}}%
\pgfpathlineto{\pgfqpoint{5.769560in}{3.631503in}}%
\pgfpathlineto{\pgfqpoint{5.463338in}{3.631503in}}%
\pgfpathclose%
\pgfusepath{fill}%
\end{pgfscope}%
\begin{pgfscope}%
\pgfpathrectangle{\pgfqpoint{1.023125in}{1.668889in}}{\pgfqpoint{6.736875in}{2.091111in}}%
\pgfusepath{clip}%
\pgfsetbuttcap%
\pgfsetmiterjoin%
\definecolor{currentfill}{rgb}{1.000000,0.498039,0.054902}%
\pgfsetfillcolor{currentfill}%
\pgfsetlinewidth{0.000000pt}%
\definecolor{currentstroke}{rgb}{0.000000,0.000000,0.000000}%
\pgfsetstrokecolor{currentstroke}%
\pgfsetstrokeopacity{0.000000}%
\pgfsetdash{}{0pt}%
\pgfpathmoveto{\pgfqpoint{6.075781in}{3.616386in}}%
\pgfpathlineto{\pgfqpoint{6.382003in}{3.616386in}}%
\pgfpathlineto{\pgfqpoint{6.382003in}{3.631503in}}%
\pgfpathlineto{\pgfqpoint{6.075781in}{3.631503in}}%
\pgfpathclose%
\pgfusepath{fill}%
\end{pgfscope}%
\begin{pgfscope}%
\pgfpathrectangle{\pgfqpoint{1.023125in}{1.668889in}}{\pgfqpoint{6.736875in}{2.091111in}}%
\pgfusepath{clip}%
\pgfsetbuttcap%
\pgfsetmiterjoin%
\definecolor{currentfill}{rgb}{1.000000,0.498039,0.054902}%
\pgfsetfillcolor{currentfill}%
\pgfsetlinewidth{0.000000pt}%
\definecolor{currentstroke}{rgb}{0.000000,0.000000,0.000000}%
\pgfsetstrokecolor{currentstroke}%
\pgfsetstrokeopacity{0.000000}%
\pgfsetdash{}{0pt}%
\pgfpathmoveto{\pgfqpoint{6.688224in}{3.616386in}}%
\pgfpathlineto{\pgfqpoint{6.994446in}{3.616386in}}%
\pgfpathlineto{\pgfqpoint{6.994446in}{3.631503in}}%
\pgfpathlineto{\pgfqpoint{6.688224in}{3.631503in}}%
\pgfpathclose%
\pgfusepath{fill}%
\end{pgfscope}%
\begin{pgfscope}%
\pgfpathrectangle{\pgfqpoint{1.023125in}{1.668889in}}{\pgfqpoint{6.736875in}{2.091111in}}%
\pgfusepath{clip}%
\pgfsetbuttcap%
\pgfsetmiterjoin%
\definecolor{currentfill}{rgb}{1.000000,0.498039,0.054902}%
\pgfsetfillcolor{currentfill}%
\pgfsetlinewidth{0.000000pt}%
\definecolor{currentstroke}{rgb}{0.000000,0.000000,0.000000}%
\pgfsetstrokecolor{currentstroke}%
\pgfsetstrokeopacity{0.000000}%
\pgfsetdash{}{0pt}%
\pgfpathmoveto{\pgfqpoint{7.300668in}{1.668889in}}%
\pgfpathlineto{\pgfqpoint{7.606889in}{1.668889in}}%
\pgfpathlineto{\pgfqpoint{7.606889in}{1.668889in}}%
\pgfpathlineto{\pgfqpoint{7.300668in}{1.668889in}}%
\pgfpathclose%
\pgfusepath{fill}%
\end{pgfscope}%
\begin{pgfscope}%
\pgfsetbuttcap%
\pgfsetroundjoin%
\definecolor{currentfill}{rgb}{0.000000,0.000000,0.000000}%
\pgfsetfillcolor{currentfill}%
\pgfsetlinewidth{0.803000pt}%
\definecolor{currentstroke}{rgb}{0.000000,0.000000,0.000000}%
\pgfsetstrokecolor{currentstroke}%
\pgfsetdash{}{0pt}%
\pgfsys@defobject{currentmarker}{\pgfqpoint{0.000000in}{-0.048611in}}{\pgfqpoint{0.000000in}{0.000000in}}{%
\pgfpathmoveto{\pgfqpoint{0.000000in}{0.000000in}}%
\pgfpathlineto{\pgfqpoint{0.000000in}{-0.048611in}}%
\pgfusepath{stroke,fill}%
}%
\begin{pgfscope}%
\pgfsys@transformshift{1.329347in}{1.668889in}%
\pgfsys@useobject{currentmarker}{}%
\end{pgfscope}%
\end{pgfscope}%
\begin{pgfscope}%
\definecolor{textcolor}{rgb}{0.000000,0.000000,0.000000}%
\pgfsetstrokecolor{textcolor}%
\pgfsetfillcolor{textcolor}%
\pgftext[x=1.329347in,y=1.571667in,,top]{\color{textcolor}\rmfamily\fontsize{16.000000}{19.200000}\selectfont 0.0}%
\end{pgfscope}%
\begin{pgfscope}%
\pgfsetbuttcap%
\pgfsetroundjoin%
\definecolor{currentfill}{rgb}{0.000000,0.000000,0.000000}%
\pgfsetfillcolor{currentfill}%
\pgfsetlinewidth{0.803000pt}%
\definecolor{currentstroke}{rgb}{0.000000,0.000000,0.000000}%
\pgfsetstrokecolor{currentstroke}%
\pgfsetdash{}{0pt}%
\pgfsys@defobject{currentmarker}{\pgfqpoint{0.000000in}{-0.048611in}}{\pgfqpoint{0.000000in}{0.000000in}}{%
\pgfpathmoveto{\pgfqpoint{0.000000in}{0.000000in}}%
\pgfpathlineto{\pgfqpoint{0.000000in}{-0.048611in}}%
\pgfusepath{stroke,fill}%
}%
\begin{pgfscope}%
\pgfsys@transformshift{1.941790in}{1.668889in}%
\pgfsys@useobject{currentmarker}{}%
\end{pgfscope}%
\end{pgfscope}%
\begin{pgfscope}%
\definecolor{textcolor}{rgb}{0.000000,0.000000,0.000000}%
\pgfsetstrokecolor{textcolor}%
\pgfsetfillcolor{textcolor}%
\pgftext[x=1.941790in,y=1.571667in,,top]{\color{textcolor}\rmfamily\fontsize{16.000000}{19.200000}\selectfont 0.1}%
\end{pgfscope}%
\begin{pgfscope}%
\pgfsetbuttcap%
\pgfsetroundjoin%
\definecolor{currentfill}{rgb}{0.000000,0.000000,0.000000}%
\pgfsetfillcolor{currentfill}%
\pgfsetlinewidth{0.803000pt}%
\definecolor{currentstroke}{rgb}{0.000000,0.000000,0.000000}%
\pgfsetstrokecolor{currentstroke}%
\pgfsetdash{}{0pt}%
\pgfsys@defobject{currentmarker}{\pgfqpoint{0.000000in}{-0.048611in}}{\pgfqpoint{0.000000in}{0.000000in}}{%
\pgfpathmoveto{\pgfqpoint{0.000000in}{0.000000in}}%
\pgfpathlineto{\pgfqpoint{0.000000in}{-0.048611in}}%
\pgfusepath{stroke,fill}%
}%
\begin{pgfscope}%
\pgfsys@transformshift{2.554233in}{1.668889in}%
\pgfsys@useobject{currentmarker}{}%
\end{pgfscope}%
\end{pgfscope}%
\begin{pgfscope}%
\definecolor{textcolor}{rgb}{0.000000,0.000000,0.000000}%
\pgfsetstrokecolor{textcolor}%
\pgfsetfillcolor{textcolor}%
\pgftext[x=2.554233in,y=1.571667in,,top]{\color{textcolor}\rmfamily\fontsize{16.000000}{19.200000}\selectfont 0.2}%
\end{pgfscope}%
\begin{pgfscope}%
\pgfsetbuttcap%
\pgfsetroundjoin%
\definecolor{currentfill}{rgb}{0.000000,0.000000,0.000000}%
\pgfsetfillcolor{currentfill}%
\pgfsetlinewidth{0.803000pt}%
\definecolor{currentstroke}{rgb}{0.000000,0.000000,0.000000}%
\pgfsetstrokecolor{currentstroke}%
\pgfsetdash{}{0pt}%
\pgfsys@defobject{currentmarker}{\pgfqpoint{0.000000in}{-0.048611in}}{\pgfqpoint{0.000000in}{0.000000in}}{%
\pgfpathmoveto{\pgfqpoint{0.000000in}{0.000000in}}%
\pgfpathlineto{\pgfqpoint{0.000000in}{-0.048611in}}%
\pgfusepath{stroke,fill}%
}%
\begin{pgfscope}%
\pgfsys@transformshift{3.166676in}{1.668889in}%
\pgfsys@useobject{currentmarker}{}%
\end{pgfscope}%
\end{pgfscope}%
\begin{pgfscope}%
\definecolor{textcolor}{rgb}{0.000000,0.000000,0.000000}%
\pgfsetstrokecolor{textcolor}%
\pgfsetfillcolor{textcolor}%
\pgftext[x=3.166676in,y=1.571667in,,top]{\color{textcolor}\rmfamily\fontsize{16.000000}{19.200000}\selectfont 0.3}%
\end{pgfscope}%
\begin{pgfscope}%
\pgfsetbuttcap%
\pgfsetroundjoin%
\definecolor{currentfill}{rgb}{0.000000,0.000000,0.000000}%
\pgfsetfillcolor{currentfill}%
\pgfsetlinewidth{0.803000pt}%
\definecolor{currentstroke}{rgb}{0.000000,0.000000,0.000000}%
\pgfsetstrokecolor{currentstroke}%
\pgfsetdash{}{0pt}%
\pgfsys@defobject{currentmarker}{\pgfqpoint{0.000000in}{-0.048611in}}{\pgfqpoint{0.000000in}{0.000000in}}{%
\pgfpathmoveto{\pgfqpoint{0.000000in}{0.000000in}}%
\pgfpathlineto{\pgfqpoint{0.000000in}{-0.048611in}}%
\pgfusepath{stroke,fill}%
}%
\begin{pgfscope}%
\pgfsys@transformshift{3.779119in}{1.668889in}%
\pgfsys@useobject{currentmarker}{}%
\end{pgfscope}%
\end{pgfscope}%
\begin{pgfscope}%
\definecolor{textcolor}{rgb}{0.000000,0.000000,0.000000}%
\pgfsetstrokecolor{textcolor}%
\pgfsetfillcolor{textcolor}%
\pgftext[x=3.779119in,y=1.571667in,,top]{\color{textcolor}\rmfamily\fontsize{16.000000}{19.200000}\selectfont 0.4}%
\end{pgfscope}%
\begin{pgfscope}%
\pgfsetbuttcap%
\pgfsetroundjoin%
\definecolor{currentfill}{rgb}{0.000000,0.000000,0.000000}%
\pgfsetfillcolor{currentfill}%
\pgfsetlinewidth{0.803000pt}%
\definecolor{currentstroke}{rgb}{0.000000,0.000000,0.000000}%
\pgfsetstrokecolor{currentstroke}%
\pgfsetdash{}{0pt}%
\pgfsys@defobject{currentmarker}{\pgfqpoint{0.000000in}{-0.048611in}}{\pgfqpoint{0.000000in}{0.000000in}}{%
\pgfpathmoveto{\pgfqpoint{0.000000in}{0.000000in}}%
\pgfpathlineto{\pgfqpoint{0.000000in}{-0.048611in}}%
\pgfusepath{stroke,fill}%
}%
\begin{pgfscope}%
\pgfsys@transformshift{4.391563in}{1.668889in}%
\pgfsys@useobject{currentmarker}{}%
\end{pgfscope}%
\end{pgfscope}%
\begin{pgfscope}%
\definecolor{textcolor}{rgb}{0.000000,0.000000,0.000000}%
\pgfsetstrokecolor{textcolor}%
\pgfsetfillcolor{textcolor}%
\pgftext[x=4.391563in,y=1.571667in,,top]{\color{textcolor}\rmfamily\fontsize{16.000000}{19.200000}\selectfont 0.5}%
\end{pgfscope}%
\begin{pgfscope}%
\pgfsetbuttcap%
\pgfsetroundjoin%
\definecolor{currentfill}{rgb}{0.000000,0.000000,0.000000}%
\pgfsetfillcolor{currentfill}%
\pgfsetlinewidth{0.803000pt}%
\definecolor{currentstroke}{rgb}{0.000000,0.000000,0.000000}%
\pgfsetstrokecolor{currentstroke}%
\pgfsetdash{}{0pt}%
\pgfsys@defobject{currentmarker}{\pgfqpoint{0.000000in}{-0.048611in}}{\pgfqpoint{0.000000in}{0.000000in}}{%
\pgfpathmoveto{\pgfqpoint{0.000000in}{0.000000in}}%
\pgfpathlineto{\pgfqpoint{0.000000in}{-0.048611in}}%
\pgfusepath{stroke,fill}%
}%
\begin{pgfscope}%
\pgfsys@transformshift{5.004006in}{1.668889in}%
\pgfsys@useobject{currentmarker}{}%
\end{pgfscope}%
\end{pgfscope}%
\begin{pgfscope}%
\definecolor{textcolor}{rgb}{0.000000,0.000000,0.000000}%
\pgfsetstrokecolor{textcolor}%
\pgfsetfillcolor{textcolor}%
\pgftext[x=5.004006in,y=1.571667in,,top]{\color{textcolor}\rmfamily\fontsize{16.000000}{19.200000}\selectfont 0.6}%
\end{pgfscope}%
\begin{pgfscope}%
\pgfsetbuttcap%
\pgfsetroundjoin%
\definecolor{currentfill}{rgb}{0.000000,0.000000,0.000000}%
\pgfsetfillcolor{currentfill}%
\pgfsetlinewidth{0.803000pt}%
\definecolor{currentstroke}{rgb}{0.000000,0.000000,0.000000}%
\pgfsetstrokecolor{currentstroke}%
\pgfsetdash{}{0pt}%
\pgfsys@defobject{currentmarker}{\pgfqpoint{0.000000in}{-0.048611in}}{\pgfqpoint{0.000000in}{0.000000in}}{%
\pgfpathmoveto{\pgfqpoint{0.000000in}{0.000000in}}%
\pgfpathlineto{\pgfqpoint{0.000000in}{-0.048611in}}%
\pgfusepath{stroke,fill}%
}%
\begin{pgfscope}%
\pgfsys@transformshift{5.616449in}{1.668889in}%
\pgfsys@useobject{currentmarker}{}%
\end{pgfscope}%
\end{pgfscope}%
\begin{pgfscope}%
\definecolor{textcolor}{rgb}{0.000000,0.000000,0.000000}%
\pgfsetstrokecolor{textcolor}%
\pgfsetfillcolor{textcolor}%
\pgftext[x=5.616449in,y=1.571667in,,top]{\color{textcolor}\rmfamily\fontsize{16.000000}{19.200000}\selectfont 0.7}%
\end{pgfscope}%
\begin{pgfscope}%
\pgfsetbuttcap%
\pgfsetroundjoin%
\definecolor{currentfill}{rgb}{0.000000,0.000000,0.000000}%
\pgfsetfillcolor{currentfill}%
\pgfsetlinewidth{0.803000pt}%
\definecolor{currentstroke}{rgb}{0.000000,0.000000,0.000000}%
\pgfsetstrokecolor{currentstroke}%
\pgfsetdash{}{0pt}%
\pgfsys@defobject{currentmarker}{\pgfqpoint{0.000000in}{-0.048611in}}{\pgfqpoint{0.000000in}{0.000000in}}{%
\pgfpathmoveto{\pgfqpoint{0.000000in}{0.000000in}}%
\pgfpathlineto{\pgfqpoint{0.000000in}{-0.048611in}}%
\pgfusepath{stroke,fill}%
}%
\begin{pgfscope}%
\pgfsys@transformshift{6.228892in}{1.668889in}%
\pgfsys@useobject{currentmarker}{}%
\end{pgfscope}%
\end{pgfscope}%
\begin{pgfscope}%
\definecolor{textcolor}{rgb}{0.000000,0.000000,0.000000}%
\pgfsetstrokecolor{textcolor}%
\pgfsetfillcolor{textcolor}%
\pgftext[x=6.228892in,y=1.571667in,,top]{\color{textcolor}\rmfamily\fontsize{16.000000}{19.200000}\selectfont 0.8}%
\end{pgfscope}%
\begin{pgfscope}%
\pgfsetbuttcap%
\pgfsetroundjoin%
\definecolor{currentfill}{rgb}{0.000000,0.000000,0.000000}%
\pgfsetfillcolor{currentfill}%
\pgfsetlinewidth{0.803000pt}%
\definecolor{currentstroke}{rgb}{0.000000,0.000000,0.000000}%
\pgfsetstrokecolor{currentstroke}%
\pgfsetdash{}{0pt}%
\pgfsys@defobject{currentmarker}{\pgfqpoint{0.000000in}{-0.048611in}}{\pgfqpoint{0.000000in}{0.000000in}}{%
\pgfpathmoveto{\pgfqpoint{0.000000in}{0.000000in}}%
\pgfpathlineto{\pgfqpoint{0.000000in}{-0.048611in}}%
\pgfusepath{stroke,fill}%
}%
\begin{pgfscope}%
\pgfsys@transformshift{6.841335in}{1.668889in}%
\pgfsys@useobject{currentmarker}{}%
\end{pgfscope}%
\end{pgfscope}%
\begin{pgfscope}%
\definecolor{textcolor}{rgb}{0.000000,0.000000,0.000000}%
\pgfsetstrokecolor{textcolor}%
\pgfsetfillcolor{textcolor}%
\pgftext[x=6.841335in,y=1.571667in,,top]{\color{textcolor}\rmfamily\fontsize{16.000000}{19.200000}\selectfont 0.9}%
\end{pgfscope}%
\begin{pgfscope}%
\pgfsetbuttcap%
\pgfsetroundjoin%
\definecolor{currentfill}{rgb}{0.000000,0.000000,0.000000}%
\pgfsetfillcolor{currentfill}%
\pgfsetlinewidth{0.803000pt}%
\definecolor{currentstroke}{rgb}{0.000000,0.000000,0.000000}%
\pgfsetstrokecolor{currentstroke}%
\pgfsetdash{}{0pt}%
\pgfsys@defobject{currentmarker}{\pgfqpoint{0.000000in}{-0.048611in}}{\pgfqpoint{0.000000in}{0.000000in}}{%
\pgfpathmoveto{\pgfqpoint{0.000000in}{0.000000in}}%
\pgfpathlineto{\pgfqpoint{0.000000in}{-0.048611in}}%
\pgfusepath{stroke,fill}%
}%
\begin{pgfscope}%
\pgfsys@transformshift{7.453778in}{1.668889in}%
\pgfsys@useobject{currentmarker}{}%
\end{pgfscope}%
\end{pgfscope}%
\begin{pgfscope}%
\definecolor{textcolor}{rgb}{0.000000,0.000000,0.000000}%
\pgfsetstrokecolor{textcolor}%
\pgfsetfillcolor{textcolor}%
\pgftext[x=7.453778in,y=1.571667in,,top]{\color{textcolor}\rmfamily\fontsize{16.000000}{19.200000}\selectfont 1.0}%
\end{pgfscope}%
\begin{pgfscope}%
\definecolor{textcolor}{rgb}{0.000000,0.000000,0.000000}%
\pgfsetstrokecolor{textcolor}%
\pgfsetfillcolor{textcolor}%
\pgftext[x=4.391563in,y=1.302762in,,top]{\color{textcolor}\rmfamily\fontsize{16.000000}{19.200000}\selectfont \(\displaystyle \lambda\)}%
\end{pgfscope}%
\begin{pgfscope}%
\pgfsetbuttcap%
\pgfsetroundjoin%
\definecolor{currentfill}{rgb}{0.000000,0.000000,0.000000}%
\pgfsetfillcolor{currentfill}%
\pgfsetlinewidth{0.803000pt}%
\definecolor{currentstroke}{rgb}{0.000000,0.000000,0.000000}%
\pgfsetstrokecolor{currentstroke}%
\pgfsetdash{}{0pt}%
\pgfsys@defobject{currentmarker}{\pgfqpoint{-0.048611in}{0.000000in}}{\pgfqpoint{-0.000000in}{0.000000in}}{%
\pgfpathmoveto{\pgfqpoint{-0.000000in}{0.000000in}}%
\pgfpathlineto{\pgfqpoint{-0.048611in}{0.000000in}}%
\pgfusepath{stroke,fill}%
}%
\begin{pgfscope}%
\pgfsys@transformshift{1.023125in}{1.668889in}%
\pgfsys@useobject{currentmarker}{}%
\end{pgfscope}%
\end{pgfscope}%
\begin{pgfscope}%
\definecolor{textcolor}{rgb}{0.000000,0.000000,0.000000}%
\pgfsetstrokecolor{textcolor}%
\pgfsetfillcolor{textcolor}%
\pgftext[x=0.815835in, y=1.585556in, left, base]{\color{textcolor}\rmfamily\fontsize{16.000000}{19.200000}\selectfont \(\displaystyle {0}\)}%
\end{pgfscope}%
\begin{pgfscope}%
\pgfsetbuttcap%
\pgfsetroundjoin%
\definecolor{currentfill}{rgb}{0.000000,0.000000,0.000000}%
\pgfsetfillcolor{currentfill}%
\pgfsetlinewidth{0.803000pt}%
\definecolor{currentstroke}{rgb}{0.000000,0.000000,0.000000}%
\pgfsetstrokecolor{currentstroke}%
\pgfsetdash{}{0pt}%
\pgfsys@defobject{currentmarker}{\pgfqpoint{-0.048611in}{0.000000in}}{\pgfqpoint{-0.000000in}{0.000000in}}{%
\pgfpathmoveto{\pgfqpoint{-0.000000in}{0.000000in}}%
\pgfpathlineto{\pgfqpoint{-0.048611in}{0.000000in}}%
\pgfusepath{stroke,fill}%
}%
\begin{pgfscope}%
\pgfsys@transformshift{1.023125in}{2.326161in}%
\pgfsys@useobject{currentmarker}{}%
\end{pgfscope}%
\end{pgfscope}%
\begin{pgfscope}%
\definecolor{textcolor}{rgb}{0.000000,0.000000,0.000000}%
\pgfsetstrokecolor{textcolor}%
\pgfsetfillcolor{textcolor}%
\pgftext[x=0.485630in, y=2.242828in, left, base]{\color{textcolor}\rmfamily\fontsize{16.000000}{19.200000}\selectfont \(\displaystyle {1000}\)}%
\end{pgfscope}%
\begin{pgfscope}%
\pgfsetbuttcap%
\pgfsetroundjoin%
\definecolor{currentfill}{rgb}{0.000000,0.000000,0.000000}%
\pgfsetfillcolor{currentfill}%
\pgfsetlinewidth{0.803000pt}%
\definecolor{currentstroke}{rgb}{0.000000,0.000000,0.000000}%
\pgfsetstrokecolor{currentstroke}%
\pgfsetdash{}{0pt}%
\pgfsys@defobject{currentmarker}{\pgfqpoint{-0.048611in}{0.000000in}}{\pgfqpoint{-0.000000in}{0.000000in}}{%
\pgfpathmoveto{\pgfqpoint{-0.000000in}{0.000000in}}%
\pgfpathlineto{\pgfqpoint{-0.048611in}{0.000000in}}%
\pgfusepath{stroke,fill}%
}%
\begin{pgfscope}%
\pgfsys@transformshift{1.023125in}{2.983433in}%
\pgfsys@useobject{currentmarker}{}%
\end{pgfscope}%
\end{pgfscope}%
\begin{pgfscope}%
\definecolor{textcolor}{rgb}{0.000000,0.000000,0.000000}%
\pgfsetstrokecolor{textcolor}%
\pgfsetfillcolor{textcolor}%
\pgftext[x=0.485630in, y=2.900100in, left, base]{\color{textcolor}\rmfamily\fontsize{16.000000}{19.200000}\selectfont \(\displaystyle {2000}\)}%
\end{pgfscope}%
\begin{pgfscope}%
\pgfsetbuttcap%
\pgfsetroundjoin%
\definecolor{currentfill}{rgb}{0.000000,0.000000,0.000000}%
\pgfsetfillcolor{currentfill}%
\pgfsetlinewidth{0.803000pt}%
\definecolor{currentstroke}{rgb}{0.000000,0.000000,0.000000}%
\pgfsetstrokecolor{currentstroke}%
\pgfsetdash{}{0pt}%
\pgfsys@defobject{currentmarker}{\pgfqpoint{-0.048611in}{0.000000in}}{\pgfqpoint{-0.000000in}{0.000000in}}{%
\pgfpathmoveto{\pgfqpoint{-0.000000in}{0.000000in}}%
\pgfpathlineto{\pgfqpoint{-0.048611in}{0.000000in}}%
\pgfusepath{stroke,fill}%
}%
\begin{pgfscope}%
\pgfsys@transformshift{1.023125in}{3.640705in}%
\pgfsys@useobject{currentmarker}{}%
\end{pgfscope}%
\end{pgfscope}%
\begin{pgfscope}%
\definecolor{textcolor}{rgb}{0.000000,0.000000,0.000000}%
\pgfsetstrokecolor{textcolor}%
\pgfsetfillcolor{textcolor}%
\pgftext[x=0.485630in, y=3.557372in, left, base]{\color{textcolor}\rmfamily\fontsize{16.000000}{19.200000}\selectfont \(\displaystyle {3000}\)}%
\end{pgfscope}%
\begin{pgfscope}%
\definecolor{textcolor}{rgb}{0.000000,0.000000,0.000000}%
\pgfsetstrokecolor{textcolor}%
\pgfsetfillcolor{textcolor}%
\pgftext[x=0.430074in,y=2.714444in,,bottom,rotate=90.000000]{\color{textcolor}\rmfamily\fontsize{16.000000}{19.200000}\selectfont number of splits}%
\end{pgfscope}%
\begin{pgfscope}%
\pgfsetrectcap%
\pgfsetmiterjoin%
\pgfsetlinewidth{0.803000pt}%
\definecolor{currentstroke}{rgb}{0.000000,0.000000,0.000000}%
\pgfsetstrokecolor{currentstroke}%
\pgfsetdash{}{0pt}%
\pgfpathmoveto{\pgfqpoint{1.023125in}{1.668889in}}%
\pgfpathlineto{\pgfqpoint{1.023125in}{3.760000in}}%
\pgfusepath{stroke}%
\end{pgfscope}%
\begin{pgfscope}%
\pgfsetrectcap%
\pgfsetmiterjoin%
\pgfsetlinewidth{0.803000pt}%
\definecolor{currentstroke}{rgb}{0.000000,0.000000,0.000000}%
\pgfsetstrokecolor{currentstroke}%
\pgfsetdash{}{0pt}%
\pgfpathmoveto{\pgfqpoint{7.760000in}{1.668889in}}%
\pgfpathlineto{\pgfqpoint{7.760000in}{3.760000in}}%
\pgfusepath{stroke}%
\end{pgfscope}%
\begin{pgfscope}%
\pgfsetrectcap%
\pgfsetmiterjoin%
\pgfsetlinewidth{0.803000pt}%
\definecolor{currentstroke}{rgb}{0.000000,0.000000,0.000000}%
\pgfsetstrokecolor{currentstroke}%
\pgfsetdash{}{0pt}%
\pgfpathmoveto{\pgfqpoint{1.023125in}{1.668889in}}%
\pgfpathlineto{\pgfqpoint{7.760000in}{1.668889in}}%
\pgfusepath{stroke}%
\end{pgfscope}%
\begin{pgfscope}%
\pgfsetrectcap%
\pgfsetmiterjoin%
\pgfsetlinewidth{0.803000pt}%
\definecolor{currentstroke}{rgb}{0.000000,0.000000,0.000000}%
\pgfsetstrokecolor{currentstroke}%
\pgfsetdash{}{0pt}%
\pgfpathmoveto{\pgfqpoint{1.023125in}{3.760000in}}%
\pgfpathlineto{\pgfqpoint{7.760000in}{3.760000in}}%
\pgfusepath{stroke}%
\end{pgfscope}%
\begin{pgfscope}%
\pgfsetbuttcap%
\pgfsetmiterjoin%
\definecolor{currentfill}{rgb}{1.000000,1.000000,1.000000}%
\pgfsetfillcolor{currentfill}%
\pgfsetfillopacity{0.800000}%
\pgfsetlinewidth{1.003750pt}%
\definecolor{currentstroke}{rgb}{0.800000,0.800000,0.800000}%
\pgfsetstrokecolor{currentstroke}%
\pgfsetstrokeopacity{0.800000}%
\pgfsetdash{}{0pt}%
\pgfpathmoveto{\pgfqpoint{1.789276in}{0.734444in}}%
\pgfpathlineto{\pgfqpoint{6.993849in}{0.734444in}}%
\pgfpathquadraticcurveto{\pgfqpoint{7.038293in}{0.734444in}}{\pgfqpoint{7.038293in}{0.778889in}}%
\pgfpathlineto{\pgfqpoint{7.038293in}{1.081126in}}%
\pgfpathquadraticcurveto{\pgfqpoint{7.038293in}{1.125571in}}{\pgfqpoint{6.993849in}{1.125571in}}%
\pgfpathlineto{\pgfqpoint{1.789276in}{1.125571in}}%
\pgfpathquadraticcurveto{\pgfqpoint{1.744832in}{1.125571in}}{\pgfqpoint{1.744832in}{1.081126in}}%
\pgfpathlineto{\pgfqpoint{1.744832in}{0.778889in}}%
\pgfpathquadraticcurveto{\pgfqpoint{1.744832in}{0.734444in}}{\pgfqpoint{1.789276in}{0.734444in}}%
\pgfpathclose%
\pgfusepath{stroke,fill}%
\end{pgfscope}%
\begin{pgfscope}%
\pgfsetbuttcap%
\pgfsetmiterjoin%
\definecolor{currentfill}{rgb}{0.121569,0.466667,0.705882}%
\pgfsetfillcolor{currentfill}%
\pgfsetlinewidth{0.000000pt}%
\definecolor{currentstroke}{rgb}{0.000000,0.000000,0.000000}%
\pgfsetstrokecolor{currentstroke}%
\pgfsetstrokeopacity{0.000000}%
\pgfsetdash{}{0pt}%
\pgfpathmoveto{\pgfqpoint{1.833721in}{0.870015in}}%
\pgfpathlineto{\pgfqpoint{2.278165in}{0.870015in}}%
\pgfpathlineto{\pgfqpoint{2.278165in}{1.025571in}}%
\pgfpathlineto{\pgfqpoint{1.833721in}{1.025571in}}%
\pgfpathclose%
\pgfusepath{fill}%
\end{pgfscope}%
\begin{pgfscope}%
\definecolor{textcolor}{rgb}{0.000000,0.000000,0.000000}%
\pgfsetstrokecolor{textcolor}%
\pgfsetfillcolor{textcolor}%
\pgftext[x=2.455943in,y=0.870015in,left,base]{\color{textcolor}\rmfamily\fontsize{16.000000}{19.200000}\selectfont relational attribute}%
\end{pgfscope}%
\begin{pgfscope}%
\pgfsetbuttcap%
\pgfsetmiterjoin%
\definecolor{currentfill}{rgb}{1.000000,0.498039,0.054902}%
\pgfsetfillcolor{currentfill}%
\pgfsetlinewidth{0.000000pt}%
\definecolor{currentstroke}{rgb}{0.000000,0.000000,0.000000}%
\pgfsetstrokecolor{currentstroke}%
\pgfsetstrokeopacity{0.000000}%
\pgfsetdash{}{0pt}%
\pgfpathmoveto{\pgfqpoint{4.725599in}{0.870015in}}%
\pgfpathlineto{\pgfqpoint{5.170043in}{0.870015in}}%
\pgfpathlineto{\pgfqpoint{5.170043in}{1.025571in}}%
\pgfpathlineto{\pgfqpoint{4.725599in}{1.025571in}}%
\pgfpathclose%
\pgfusepath{fill}%
\end{pgfscope}%
\begin{pgfscope}%
\definecolor{textcolor}{rgb}{0.000000,0.000000,0.000000}%
\pgfsetstrokecolor{textcolor}%
\pgfsetfillcolor{textcolor}%
\pgftext[x=5.347821in,y=0.870015in,left,base]{\color{textcolor}\rmfamily\fontsize{16.000000}{19.200000}\selectfont textual attribute}%
\end{pgfscope}%
\end{pgfpicture}%
\makeatother%
\endgroup%

%% file: images/splits_reviews_all_5_shrinked.pgf
\begingroup%
\makeatletter%
\begin{pgfpicture}%
\pgfpathrectangle{\pgfpointorigin}{\pgfqpoint{8.000000in}{4.000000in}}%
\pgfusepath{use as bounding box, clip}%
\begin{pgfscope}%
\pgfsetbuttcap%
\pgfsetmiterjoin%
\definecolor{currentfill}{rgb}{1.000000,1.000000,1.000000}%
\pgfsetfillcolor{currentfill}%
\pgfsetlinewidth{0.000000pt}%
\definecolor{currentstroke}{rgb}{1.000000,1.000000,1.000000}%
\pgfsetstrokecolor{currentstroke}%
\pgfsetdash{}{0pt}%
\pgfpathmoveto{\pgfqpoint{0.000000in}{0.000000in}}%
\pgfpathlineto{\pgfqpoint{8.000000in}{0.000000in}}%
\pgfpathlineto{\pgfqpoint{8.000000in}{4.000000in}}%
\pgfpathlineto{\pgfqpoint{0.000000in}{4.000000in}}%
\pgfpathclose%
\pgfusepath{fill}%
\end{pgfscope}%
\begin{pgfscope}%
\pgfsetbuttcap%
\pgfsetmiterjoin%
\definecolor{currentfill}{rgb}{1.000000,1.000000,1.000000}%
\pgfsetfillcolor{currentfill}%
\pgfsetlinewidth{0.000000pt}%
\definecolor{currentstroke}{rgb}{0.000000,0.000000,0.000000}%
\pgfsetstrokecolor{currentstroke}%
\pgfsetstrokeopacity{0.000000}%
\pgfsetdash{}{0pt}%
\pgfpathmoveto{\pgfqpoint{0.914736in}{1.668889in}}%
\pgfpathlineto{\pgfqpoint{7.760000in}{1.668889in}}%
\pgfpathlineto{\pgfqpoint{7.760000in}{3.760000in}}%
\pgfpathlineto{\pgfqpoint{0.914736in}{3.760000in}}%
\pgfpathclose%
\pgfusepath{fill}%
\end{pgfscope}%
\begin{pgfscope}%
\pgfpathrectangle{\pgfqpoint{0.914736in}{1.668889in}}{\pgfqpoint{6.845264in}{2.091111in}}%
\pgfusepath{clip}%
\pgfsetbuttcap%
\pgfsetmiterjoin%
\definecolor{currentfill}{rgb}{0.121569,0.466667,0.705882}%
\pgfsetfillcolor{currentfill}%
\pgfsetlinewidth{0.000000pt}%
\definecolor{currentstroke}{rgb}{0.000000,0.000000,0.000000}%
\pgfsetstrokecolor{currentstroke}%
\pgfsetstrokeopacity{0.000000}%
\pgfsetdash{}{0pt}%
\pgfpathmoveto{\pgfqpoint{1.070311in}{1.668889in}}%
\pgfpathlineto{\pgfqpoint{1.381459in}{1.668889in}}%
\pgfpathlineto{\pgfqpoint{1.381459in}{1.668889in}}%
\pgfpathlineto{\pgfqpoint{1.070311in}{1.668889in}}%
\pgfpathclose%
\pgfusepath{fill}%
\end{pgfscope}%
\begin{pgfscope}%
\pgfpathrectangle{\pgfqpoint{0.914736in}{1.668889in}}{\pgfqpoint{6.845264in}{2.091111in}}%
\pgfusepath{clip}%
\pgfsetbuttcap%
\pgfsetmiterjoin%
\definecolor{currentfill}{rgb}{0.121569,0.466667,0.705882}%
\pgfsetfillcolor{currentfill}%
\pgfsetlinewidth{0.000000pt}%
\definecolor{currentstroke}{rgb}{0.000000,0.000000,0.000000}%
\pgfsetstrokecolor{currentstroke}%
\pgfsetstrokeopacity{0.000000}%
\pgfsetdash{}{0pt}%
\pgfpathmoveto{\pgfqpoint{1.692607in}{1.668889in}}%
\pgfpathlineto{\pgfqpoint{2.003756in}{1.668889in}}%
\pgfpathlineto{\pgfqpoint{2.003756in}{2.194215in}}%
\pgfpathlineto{\pgfqpoint{1.692607in}{2.194215in}}%
\pgfpathclose%
\pgfusepath{fill}%
\end{pgfscope}%
\begin{pgfscope}%
\pgfpathrectangle{\pgfqpoint{0.914736in}{1.668889in}}{\pgfqpoint{6.845264in}{2.091111in}}%
\pgfusepath{clip}%
\pgfsetbuttcap%
\pgfsetmiterjoin%
\definecolor{currentfill}{rgb}{0.121569,0.466667,0.705882}%
\pgfsetfillcolor{currentfill}%
\pgfsetlinewidth{0.000000pt}%
\definecolor{currentstroke}{rgb}{0.000000,0.000000,0.000000}%
\pgfsetstrokecolor{currentstroke}%
\pgfsetstrokeopacity{0.000000}%
\pgfsetdash{}{0pt}%
\pgfpathmoveto{\pgfqpoint{2.314904in}{1.668889in}}%
\pgfpathlineto{\pgfqpoint{2.626052in}{1.668889in}}%
\pgfpathlineto{\pgfqpoint{2.626052in}{2.382391in}}%
\pgfpathlineto{\pgfqpoint{2.314904in}{2.382391in}}%
\pgfpathclose%
\pgfusepath{fill}%
\end{pgfscope}%
\begin{pgfscope}%
\pgfpathrectangle{\pgfqpoint{0.914736in}{1.668889in}}{\pgfqpoint{6.845264in}{2.091111in}}%
\pgfusepath{clip}%
\pgfsetbuttcap%
\pgfsetmiterjoin%
\definecolor{currentfill}{rgb}{0.121569,0.466667,0.705882}%
\pgfsetfillcolor{currentfill}%
\pgfsetlinewidth{0.000000pt}%
\definecolor{currentstroke}{rgb}{0.000000,0.000000,0.000000}%
\pgfsetstrokecolor{currentstroke}%
\pgfsetstrokeopacity{0.000000}%
\pgfsetdash{}{0pt}%
\pgfpathmoveto{\pgfqpoint{2.937201in}{1.668889in}}%
\pgfpathlineto{\pgfqpoint{3.248349in}{1.668889in}}%
\pgfpathlineto{\pgfqpoint{3.248349in}{2.672497in}}%
\pgfpathlineto{\pgfqpoint{2.937201in}{2.672497in}}%
\pgfpathclose%
\pgfusepath{fill}%
\end{pgfscope}%
\begin{pgfscope}%
\pgfpathrectangle{\pgfqpoint{0.914736in}{1.668889in}}{\pgfqpoint{6.845264in}{2.091111in}}%
\pgfusepath{clip}%
\pgfsetbuttcap%
\pgfsetmiterjoin%
\definecolor{currentfill}{rgb}{0.121569,0.466667,0.705882}%
\pgfsetfillcolor{currentfill}%
\pgfsetlinewidth{0.000000pt}%
\definecolor{currentstroke}{rgb}{0.000000,0.000000,0.000000}%
\pgfsetstrokecolor{currentstroke}%
\pgfsetstrokeopacity{0.000000}%
\pgfsetdash{}{0pt}%
\pgfpathmoveto{\pgfqpoint{3.559497in}{1.668889in}}%
\pgfpathlineto{\pgfqpoint{3.870646in}{1.668889in}}%
\pgfpathlineto{\pgfqpoint{3.870646in}{3.213504in}}%
\pgfpathlineto{\pgfqpoint{3.559497in}{3.213504in}}%
\pgfpathclose%
\pgfusepath{fill}%
\end{pgfscope}%
\begin{pgfscope}%
\pgfpathrectangle{\pgfqpoint{0.914736in}{1.668889in}}{\pgfqpoint{6.845264in}{2.091111in}}%
\pgfusepath{clip}%
\pgfsetbuttcap%
\pgfsetmiterjoin%
\definecolor{currentfill}{rgb}{0.121569,0.466667,0.705882}%
\pgfsetfillcolor{currentfill}%
\pgfsetlinewidth{0.000000pt}%
\definecolor{currentstroke}{rgb}{0.000000,0.000000,0.000000}%
\pgfsetstrokecolor{currentstroke}%
\pgfsetstrokeopacity{0.000000}%
\pgfsetdash{}{0pt}%
\pgfpathmoveto{\pgfqpoint{4.181794in}{1.668889in}}%
\pgfpathlineto{\pgfqpoint{4.492942in}{1.668889in}}%
\pgfpathlineto{\pgfqpoint{4.492942in}{3.652583in}}%
\pgfpathlineto{\pgfqpoint{4.181794in}{3.652583in}}%
\pgfpathclose%
\pgfusepath{fill}%
\end{pgfscope}%
\begin{pgfscope}%
\pgfpathrectangle{\pgfqpoint{0.914736in}{1.668889in}}{\pgfqpoint{6.845264in}{2.091111in}}%
\pgfusepath{clip}%
\pgfsetbuttcap%
\pgfsetmiterjoin%
\definecolor{currentfill}{rgb}{0.121569,0.466667,0.705882}%
\pgfsetfillcolor{currentfill}%
\pgfsetlinewidth{0.000000pt}%
\definecolor{currentstroke}{rgb}{0.000000,0.000000,0.000000}%
\pgfsetstrokecolor{currentstroke}%
\pgfsetstrokeopacity{0.000000}%
\pgfsetdash{}{0pt}%
\pgfpathmoveto{\pgfqpoint{4.804091in}{1.668889in}}%
\pgfpathlineto{\pgfqpoint{5.115239in}{1.668889in}}%
\pgfpathlineto{\pgfqpoint{5.115239in}{3.660423in}}%
\pgfpathlineto{\pgfqpoint{4.804091in}{3.660423in}}%
\pgfpathclose%
\pgfusepath{fill}%
\end{pgfscope}%
\begin{pgfscope}%
\pgfpathrectangle{\pgfqpoint{0.914736in}{1.668889in}}{\pgfqpoint{6.845264in}{2.091111in}}%
\pgfusepath{clip}%
\pgfsetbuttcap%
\pgfsetmiterjoin%
\definecolor{currentfill}{rgb}{0.121569,0.466667,0.705882}%
\pgfsetfillcolor{currentfill}%
\pgfsetlinewidth{0.000000pt}%
\definecolor{currentstroke}{rgb}{0.000000,0.000000,0.000000}%
\pgfsetstrokecolor{currentstroke}%
\pgfsetstrokeopacity{0.000000}%
\pgfsetdash{}{0pt}%
\pgfpathmoveto{\pgfqpoint{5.426387in}{1.668889in}}%
\pgfpathlineto{\pgfqpoint{5.737536in}{1.668889in}}%
\pgfpathlineto{\pgfqpoint{5.737536in}{3.660423in}}%
\pgfpathlineto{\pgfqpoint{5.426387in}{3.660423in}}%
\pgfpathclose%
\pgfusepath{fill}%
\end{pgfscope}%
\begin{pgfscope}%
\pgfpathrectangle{\pgfqpoint{0.914736in}{1.668889in}}{\pgfqpoint{6.845264in}{2.091111in}}%
\pgfusepath{clip}%
\pgfsetbuttcap%
\pgfsetmiterjoin%
\definecolor{currentfill}{rgb}{0.121569,0.466667,0.705882}%
\pgfsetfillcolor{currentfill}%
\pgfsetlinewidth{0.000000pt}%
\definecolor{currentstroke}{rgb}{0.000000,0.000000,0.000000}%
\pgfsetstrokecolor{currentstroke}%
\pgfsetstrokeopacity{0.000000}%
\pgfsetdash{}{0pt}%
\pgfpathmoveto{\pgfqpoint{6.048684in}{1.668889in}}%
\pgfpathlineto{\pgfqpoint{6.359832in}{1.668889in}}%
\pgfpathlineto{\pgfqpoint{6.359832in}{3.660423in}}%
\pgfpathlineto{\pgfqpoint{6.048684in}{3.660423in}}%
\pgfpathclose%
\pgfusepath{fill}%
\end{pgfscope}%
\begin{pgfscope}%
\pgfpathrectangle{\pgfqpoint{0.914736in}{1.668889in}}{\pgfqpoint{6.845264in}{2.091111in}}%
\pgfusepath{clip}%
\pgfsetbuttcap%
\pgfsetmiterjoin%
\definecolor{currentfill}{rgb}{0.121569,0.466667,0.705882}%
\pgfsetfillcolor{currentfill}%
\pgfsetlinewidth{0.000000pt}%
\definecolor{currentstroke}{rgb}{0.000000,0.000000,0.000000}%
\pgfsetstrokecolor{currentstroke}%
\pgfsetstrokeopacity{0.000000}%
\pgfsetdash{}{0pt}%
\pgfpathmoveto{\pgfqpoint{6.670981in}{1.668889in}}%
\pgfpathlineto{\pgfqpoint{6.982129in}{1.668889in}}%
\pgfpathlineto{\pgfqpoint{6.982129in}{3.660423in}}%
\pgfpathlineto{\pgfqpoint{6.670981in}{3.660423in}}%
\pgfpathclose%
\pgfusepath{fill}%
\end{pgfscope}%
\begin{pgfscope}%
\pgfpathrectangle{\pgfqpoint{0.914736in}{1.668889in}}{\pgfqpoint{6.845264in}{2.091111in}}%
\pgfusepath{clip}%
\pgfsetbuttcap%
\pgfsetmiterjoin%
\definecolor{currentfill}{rgb}{0.121569,0.466667,0.705882}%
\pgfsetfillcolor{currentfill}%
\pgfsetlinewidth{0.000000pt}%
\definecolor{currentstroke}{rgb}{0.000000,0.000000,0.000000}%
\pgfsetstrokecolor{currentstroke}%
\pgfsetstrokeopacity{0.000000}%
\pgfsetdash{}{0pt}%
\pgfpathmoveto{\pgfqpoint{7.293277in}{1.668889in}}%
\pgfpathlineto{\pgfqpoint{7.604426in}{1.668889in}}%
\pgfpathlineto{\pgfqpoint{7.604426in}{3.660423in}}%
\pgfpathlineto{\pgfqpoint{7.293277in}{3.660423in}}%
\pgfpathclose%
\pgfusepath{fill}%
\end{pgfscope}%
\begin{pgfscope}%
\pgfpathrectangle{\pgfqpoint{0.914736in}{1.668889in}}{\pgfqpoint{6.845264in}{2.091111in}}%
\pgfusepath{clip}%
\pgfsetbuttcap%
\pgfsetmiterjoin%
\definecolor{currentfill}{rgb}{1.000000,0.498039,0.054902}%
\pgfsetfillcolor{currentfill}%
\pgfsetlinewidth{0.000000pt}%
\definecolor{currentstroke}{rgb}{0.000000,0.000000,0.000000}%
\pgfsetstrokecolor{currentstroke}%
\pgfsetstrokeopacity{0.000000}%
\pgfsetdash{}{0pt}%
\pgfpathmoveto{\pgfqpoint{1.070311in}{1.668889in}}%
\pgfpathlineto{\pgfqpoint{1.381459in}{1.668889in}}%
\pgfpathlineto{\pgfqpoint{1.381459in}{3.433043in}}%
\pgfpathlineto{\pgfqpoint{1.070311in}{3.433043in}}%
\pgfpathclose%
\pgfusepath{fill}%
\end{pgfscope}%
\begin{pgfscope}%
\pgfpathrectangle{\pgfqpoint{0.914736in}{1.668889in}}{\pgfqpoint{6.845264in}{2.091111in}}%
\pgfusepath{clip}%
\pgfsetbuttcap%
\pgfsetmiterjoin%
\definecolor{currentfill}{rgb}{1.000000,0.498039,0.054902}%
\pgfsetfillcolor{currentfill}%
\pgfsetlinewidth{0.000000pt}%
\definecolor{currentstroke}{rgb}{0.000000,0.000000,0.000000}%
\pgfsetstrokecolor{currentstroke}%
\pgfsetstrokeopacity{0.000000}%
\pgfsetdash{}{0pt}%
\pgfpathmoveto{\pgfqpoint{1.692607in}{2.194215in}}%
\pgfpathlineto{\pgfqpoint{2.003756in}{2.194215in}}%
\pgfpathlineto{\pgfqpoint{2.003756in}{3.472247in}}%
\pgfpathlineto{\pgfqpoint{1.692607in}{3.472247in}}%
\pgfpathclose%
\pgfusepath{fill}%
\end{pgfscope}%
\begin{pgfscope}%
\pgfpathrectangle{\pgfqpoint{0.914736in}{1.668889in}}{\pgfqpoint{6.845264in}{2.091111in}}%
\pgfusepath{clip}%
\pgfsetbuttcap%
\pgfsetmiterjoin%
\definecolor{currentfill}{rgb}{1.000000,0.498039,0.054902}%
\pgfsetfillcolor{currentfill}%
\pgfsetlinewidth{0.000000pt}%
\definecolor{currentstroke}{rgb}{0.000000,0.000000,0.000000}%
\pgfsetstrokecolor{currentstroke}%
\pgfsetstrokeopacity{0.000000}%
\pgfsetdash{}{0pt}%
\pgfpathmoveto{\pgfqpoint{2.314904in}{2.382391in}}%
\pgfpathlineto{\pgfqpoint{2.626052in}{2.382391in}}%
\pgfpathlineto{\pgfqpoint{2.626052in}{3.417362in}}%
\pgfpathlineto{\pgfqpoint{2.314904in}{3.417362in}}%
\pgfpathclose%
\pgfusepath{fill}%
\end{pgfscope}%
\begin{pgfscope}%
\pgfpathrectangle{\pgfqpoint{0.914736in}{1.668889in}}{\pgfqpoint{6.845264in}{2.091111in}}%
\pgfusepath{clip}%
\pgfsetbuttcap%
\pgfsetmiterjoin%
\definecolor{currentfill}{rgb}{1.000000,0.498039,0.054902}%
\pgfsetfillcolor{currentfill}%
\pgfsetlinewidth{0.000000pt}%
\definecolor{currentstroke}{rgb}{0.000000,0.000000,0.000000}%
\pgfsetstrokecolor{currentstroke}%
\pgfsetstrokeopacity{0.000000}%
\pgfsetdash{}{0pt}%
\pgfpathmoveto{\pgfqpoint{2.937201in}{2.672497in}}%
\pgfpathlineto{\pgfqpoint{3.248349in}{2.672497in}}%
\pgfpathlineto{\pgfqpoint{3.248349in}{3.323274in}}%
\pgfpathlineto{\pgfqpoint{2.937201in}{3.323274in}}%
\pgfpathclose%
\pgfusepath{fill}%
\end{pgfscope}%
\begin{pgfscope}%
\pgfpathrectangle{\pgfqpoint{0.914736in}{1.668889in}}{\pgfqpoint{6.845264in}{2.091111in}}%
\pgfusepath{clip}%
\pgfsetbuttcap%
\pgfsetmiterjoin%
\definecolor{currentfill}{rgb}{1.000000,0.498039,0.054902}%
\pgfsetfillcolor{currentfill}%
\pgfsetlinewidth{0.000000pt}%
\definecolor{currentstroke}{rgb}{0.000000,0.000000,0.000000}%
\pgfsetstrokecolor{currentstroke}%
\pgfsetstrokeopacity{0.000000}%
\pgfsetdash{}{0pt}%
\pgfpathmoveto{\pgfqpoint{3.559497in}{3.213504in}}%
\pgfpathlineto{\pgfqpoint{3.870646in}{3.213504in}}%
\pgfpathlineto{\pgfqpoint{3.870646in}{3.299752in}}%
\pgfpathlineto{\pgfqpoint{3.559497in}{3.299752in}}%
\pgfpathclose%
\pgfusepath{fill}%
\end{pgfscope}%
\begin{pgfscope}%
\pgfpathrectangle{\pgfqpoint{0.914736in}{1.668889in}}{\pgfqpoint{6.845264in}{2.091111in}}%
\pgfusepath{clip}%
\pgfsetbuttcap%
\pgfsetmiterjoin%
\definecolor{currentfill}{rgb}{1.000000,0.498039,0.054902}%
\pgfsetfillcolor{currentfill}%
\pgfsetlinewidth{0.000000pt}%
\definecolor{currentstroke}{rgb}{0.000000,0.000000,0.000000}%
\pgfsetstrokecolor{currentstroke}%
\pgfsetstrokeopacity{0.000000}%
\pgfsetdash{}{0pt}%
\pgfpathmoveto{\pgfqpoint{4.181794in}{3.652583in}}%
\pgfpathlineto{\pgfqpoint{4.492942in}{3.652583in}}%
\pgfpathlineto{\pgfqpoint{4.492942in}{3.660423in}}%
\pgfpathlineto{\pgfqpoint{4.181794in}{3.660423in}}%
\pgfpathclose%
\pgfusepath{fill}%
\end{pgfscope}%
\begin{pgfscope}%
\pgfpathrectangle{\pgfqpoint{0.914736in}{1.668889in}}{\pgfqpoint{6.845264in}{2.091111in}}%
\pgfusepath{clip}%
\pgfsetbuttcap%
\pgfsetmiterjoin%
\definecolor{currentfill}{rgb}{1.000000,0.498039,0.054902}%
\pgfsetfillcolor{currentfill}%
\pgfsetlinewidth{0.000000pt}%
\definecolor{currentstroke}{rgb}{0.000000,0.000000,0.000000}%
\pgfsetstrokecolor{currentstroke}%
\pgfsetstrokeopacity{0.000000}%
\pgfsetdash{}{0pt}%
\pgfpathmoveto{\pgfqpoint{4.804091in}{1.668889in}}%
\pgfpathlineto{\pgfqpoint{5.115239in}{1.668889in}}%
\pgfpathlineto{\pgfqpoint{5.115239in}{1.668889in}}%
\pgfpathlineto{\pgfqpoint{4.804091in}{1.668889in}}%
\pgfpathclose%
\pgfusepath{fill}%
\end{pgfscope}%
\begin{pgfscope}%
\pgfpathrectangle{\pgfqpoint{0.914736in}{1.668889in}}{\pgfqpoint{6.845264in}{2.091111in}}%
\pgfusepath{clip}%
\pgfsetbuttcap%
\pgfsetmiterjoin%
\definecolor{currentfill}{rgb}{1.000000,0.498039,0.054902}%
\pgfsetfillcolor{currentfill}%
\pgfsetlinewidth{0.000000pt}%
\definecolor{currentstroke}{rgb}{0.000000,0.000000,0.000000}%
\pgfsetstrokecolor{currentstroke}%
\pgfsetstrokeopacity{0.000000}%
\pgfsetdash{}{0pt}%
\pgfpathmoveto{\pgfqpoint{5.426387in}{1.668889in}}%
\pgfpathlineto{\pgfqpoint{5.737536in}{1.668889in}}%
\pgfpathlineto{\pgfqpoint{5.737536in}{1.668889in}}%
\pgfpathlineto{\pgfqpoint{5.426387in}{1.668889in}}%
\pgfpathclose%
\pgfusepath{fill}%
\end{pgfscope}%
\begin{pgfscope}%
\pgfpathrectangle{\pgfqpoint{0.914736in}{1.668889in}}{\pgfqpoint{6.845264in}{2.091111in}}%
\pgfusepath{clip}%
\pgfsetbuttcap%
\pgfsetmiterjoin%
\definecolor{currentfill}{rgb}{1.000000,0.498039,0.054902}%
\pgfsetfillcolor{currentfill}%
\pgfsetlinewidth{0.000000pt}%
\definecolor{currentstroke}{rgb}{0.000000,0.000000,0.000000}%
\pgfsetstrokecolor{currentstroke}%
\pgfsetstrokeopacity{0.000000}%
\pgfsetdash{}{0pt}%
\pgfpathmoveto{\pgfqpoint{6.048684in}{1.668889in}}%
\pgfpathlineto{\pgfqpoint{6.359832in}{1.668889in}}%
\pgfpathlineto{\pgfqpoint{6.359832in}{1.668889in}}%
\pgfpathlineto{\pgfqpoint{6.048684in}{1.668889in}}%
\pgfpathclose%
\pgfusepath{fill}%
\end{pgfscope}%
\begin{pgfscope}%
\pgfpathrectangle{\pgfqpoint{0.914736in}{1.668889in}}{\pgfqpoint{6.845264in}{2.091111in}}%
\pgfusepath{clip}%
\pgfsetbuttcap%
\pgfsetmiterjoin%
\definecolor{currentfill}{rgb}{1.000000,0.498039,0.054902}%
\pgfsetfillcolor{currentfill}%
\pgfsetlinewidth{0.000000pt}%
\definecolor{currentstroke}{rgb}{0.000000,0.000000,0.000000}%
\pgfsetstrokecolor{currentstroke}%
\pgfsetstrokeopacity{0.000000}%
\pgfsetdash{}{0pt}%
\pgfpathmoveto{\pgfqpoint{6.670981in}{1.668889in}}%
\pgfpathlineto{\pgfqpoint{6.982129in}{1.668889in}}%
\pgfpathlineto{\pgfqpoint{6.982129in}{1.668889in}}%
\pgfpathlineto{\pgfqpoint{6.670981in}{1.668889in}}%
\pgfpathclose%
\pgfusepath{fill}%
\end{pgfscope}%
\begin{pgfscope}%
\pgfpathrectangle{\pgfqpoint{0.914736in}{1.668889in}}{\pgfqpoint{6.845264in}{2.091111in}}%
\pgfusepath{clip}%
\pgfsetbuttcap%
\pgfsetmiterjoin%
\definecolor{currentfill}{rgb}{1.000000,0.498039,0.054902}%
\pgfsetfillcolor{currentfill}%
\pgfsetlinewidth{0.000000pt}%
\definecolor{currentstroke}{rgb}{0.000000,0.000000,0.000000}%
\pgfsetstrokecolor{currentstroke}%
\pgfsetstrokeopacity{0.000000}%
\pgfsetdash{}{0pt}%
\pgfpathmoveto{\pgfqpoint{7.293277in}{1.668889in}}%
\pgfpathlineto{\pgfqpoint{7.604426in}{1.668889in}}%
\pgfpathlineto{\pgfqpoint{7.604426in}{1.668889in}}%
\pgfpathlineto{\pgfqpoint{7.293277in}{1.668889in}}%
\pgfpathclose%
\pgfusepath{fill}%
\end{pgfscope}%
\begin{pgfscope}%
\pgfsetbuttcap%
\pgfsetroundjoin%
\definecolor{currentfill}{rgb}{0.000000,0.000000,0.000000}%
\pgfsetfillcolor{currentfill}%
\pgfsetlinewidth{0.803000pt}%
\definecolor{currentstroke}{rgb}{0.000000,0.000000,0.000000}%
\pgfsetstrokecolor{currentstroke}%
\pgfsetdash{}{0pt}%
\pgfsys@defobject{currentmarker}{\pgfqpoint{0.000000in}{-0.048611in}}{\pgfqpoint{0.000000in}{0.000000in}}{%
\pgfpathmoveto{\pgfqpoint{0.000000in}{0.000000in}}%
\pgfpathlineto{\pgfqpoint{0.000000in}{-0.048611in}}%
\pgfusepath{stroke,fill}%
}%
\begin{pgfscope}%
\pgfsys@transformshift{1.225885in}{1.668889in}%
\pgfsys@useobject{currentmarker}{}%
\end{pgfscope}%
\end{pgfscope}%
\begin{pgfscope}%
\definecolor{textcolor}{rgb}{0.000000,0.000000,0.000000}%
\pgfsetstrokecolor{textcolor}%
\pgfsetfillcolor{textcolor}%
\pgftext[x=1.225885in,y=1.571667in,,top]{\color{textcolor}\rmfamily\fontsize{16.000000}{19.200000}\selectfont 0.0}%
\end{pgfscope}%
\begin{pgfscope}%
\pgfsetbuttcap%
\pgfsetroundjoin%
\definecolor{currentfill}{rgb}{0.000000,0.000000,0.000000}%
\pgfsetfillcolor{currentfill}%
\pgfsetlinewidth{0.803000pt}%
\definecolor{currentstroke}{rgb}{0.000000,0.000000,0.000000}%
\pgfsetstrokecolor{currentstroke}%
\pgfsetdash{}{0pt}%
\pgfsys@defobject{currentmarker}{\pgfqpoint{0.000000in}{-0.048611in}}{\pgfqpoint{0.000000in}{0.000000in}}{%
\pgfpathmoveto{\pgfqpoint{0.000000in}{0.000000in}}%
\pgfpathlineto{\pgfqpoint{0.000000in}{-0.048611in}}%
\pgfusepath{stroke,fill}%
}%
\begin{pgfscope}%
\pgfsys@transformshift{1.848181in}{1.668889in}%
\pgfsys@useobject{currentmarker}{}%
\end{pgfscope}%
\end{pgfscope}%
\begin{pgfscope}%
\definecolor{textcolor}{rgb}{0.000000,0.000000,0.000000}%
\pgfsetstrokecolor{textcolor}%
\pgfsetfillcolor{textcolor}%
\pgftext[x=1.848181in,y=1.571667in,,top]{\color{textcolor}\rmfamily\fontsize{16.000000}{19.200000}\selectfont 0.1}%
\end{pgfscope}%
\begin{pgfscope}%
\pgfsetbuttcap%
\pgfsetroundjoin%
\definecolor{currentfill}{rgb}{0.000000,0.000000,0.000000}%
\pgfsetfillcolor{currentfill}%
\pgfsetlinewidth{0.803000pt}%
\definecolor{currentstroke}{rgb}{0.000000,0.000000,0.000000}%
\pgfsetstrokecolor{currentstroke}%
\pgfsetdash{}{0pt}%
\pgfsys@defobject{currentmarker}{\pgfqpoint{0.000000in}{-0.048611in}}{\pgfqpoint{0.000000in}{0.000000in}}{%
\pgfpathmoveto{\pgfqpoint{0.000000in}{0.000000in}}%
\pgfpathlineto{\pgfqpoint{0.000000in}{-0.048611in}}%
\pgfusepath{stroke,fill}%
}%
\begin{pgfscope}%
\pgfsys@transformshift{2.470478in}{1.668889in}%
\pgfsys@useobject{currentmarker}{}%
\end{pgfscope}%
\end{pgfscope}%
\begin{pgfscope}%
\definecolor{textcolor}{rgb}{0.000000,0.000000,0.000000}%
\pgfsetstrokecolor{textcolor}%
\pgfsetfillcolor{textcolor}%
\pgftext[x=2.470478in,y=1.571667in,,top]{\color{textcolor}\rmfamily\fontsize{16.000000}{19.200000}\selectfont 0.2}%
\end{pgfscope}%
\begin{pgfscope}%
\pgfsetbuttcap%
\pgfsetroundjoin%
\definecolor{currentfill}{rgb}{0.000000,0.000000,0.000000}%
\pgfsetfillcolor{currentfill}%
\pgfsetlinewidth{0.803000pt}%
\definecolor{currentstroke}{rgb}{0.000000,0.000000,0.000000}%
\pgfsetstrokecolor{currentstroke}%
\pgfsetdash{}{0pt}%
\pgfsys@defobject{currentmarker}{\pgfqpoint{0.000000in}{-0.048611in}}{\pgfqpoint{0.000000in}{0.000000in}}{%
\pgfpathmoveto{\pgfqpoint{0.000000in}{0.000000in}}%
\pgfpathlineto{\pgfqpoint{0.000000in}{-0.048611in}}%
\pgfusepath{stroke,fill}%
}%
\begin{pgfscope}%
\pgfsys@transformshift{3.092775in}{1.668889in}%
\pgfsys@useobject{currentmarker}{}%
\end{pgfscope}%
\end{pgfscope}%
\begin{pgfscope}%
\definecolor{textcolor}{rgb}{0.000000,0.000000,0.000000}%
\pgfsetstrokecolor{textcolor}%
\pgfsetfillcolor{textcolor}%
\pgftext[x=3.092775in,y=1.571667in,,top]{\color{textcolor}\rmfamily\fontsize{16.000000}{19.200000}\selectfont 0.3}%
\end{pgfscope}%
\begin{pgfscope}%
\pgfsetbuttcap%
\pgfsetroundjoin%
\definecolor{currentfill}{rgb}{0.000000,0.000000,0.000000}%
\pgfsetfillcolor{currentfill}%
\pgfsetlinewidth{0.803000pt}%
\definecolor{currentstroke}{rgb}{0.000000,0.000000,0.000000}%
\pgfsetstrokecolor{currentstroke}%
\pgfsetdash{}{0pt}%
\pgfsys@defobject{currentmarker}{\pgfqpoint{0.000000in}{-0.048611in}}{\pgfqpoint{0.000000in}{0.000000in}}{%
\pgfpathmoveto{\pgfqpoint{0.000000in}{0.000000in}}%
\pgfpathlineto{\pgfqpoint{0.000000in}{-0.048611in}}%
\pgfusepath{stroke,fill}%
}%
\begin{pgfscope}%
\pgfsys@transformshift{3.715072in}{1.668889in}%
\pgfsys@useobject{currentmarker}{}%
\end{pgfscope}%
\end{pgfscope}%
\begin{pgfscope}%
\definecolor{textcolor}{rgb}{0.000000,0.000000,0.000000}%
\pgfsetstrokecolor{textcolor}%
\pgfsetfillcolor{textcolor}%
\pgftext[x=3.715072in,y=1.571667in,,top]{\color{textcolor}\rmfamily\fontsize{16.000000}{19.200000}\selectfont 0.4}%
\end{pgfscope}%
\begin{pgfscope}%
\pgfsetbuttcap%
\pgfsetroundjoin%
\definecolor{currentfill}{rgb}{0.000000,0.000000,0.000000}%
\pgfsetfillcolor{currentfill}%
\pgfsetlinewidth{0.803000pt}%
\definecolor{currentstroke}{rgb}{0.000000,0.000000,0.000000}%
\pgfsetstrokecolor{currentstroke}%
\pgfsetdash{}{0pt}%
\pgfsys@defobject{currentmarker}{\pgfqpoint{0.000000in}{-0.048611in}}{\pgfqpoint{0.000000in}{0.000000in}}{%
\pgfpathmoveto{\pgfqpoint{0.000000in}{0.000000in}}%
\pgfpathlineto{\pgfqpoint{0.000000in}{-0.048611in}}%
\pgfusepath{stroke,fill}%
}%
\begin{pgfscope}%
\pgfsys@transformshift{4.337368in}{1.668889in}%
\pgfsys@useobject{currentmarker}{}%
\end{pgfscope}%
\end{pgfscope}%
\begin{pgfscope}%
\definecolor{textcolor}{rgb}{0.000000,0.000000,0.000000}%
\pgfsetstrokecolor{textcolor}%
\pgfsetfillcolor{textcolor}%
\pgftext[x=4.337368in,y=1.571667in,,top]{\color{textcolor}\rmfamily\fontsize{16.000000}{19.200000}\selectfont 0.5}%
\end{pgfscope}%
\begin{pgfscope}%
\pgfsetbuttcap%
\pgfsetroundjoin%
\definecolor{currentfill}{rgb}{0.000000,0.000000,0.000000}%
\pgfsetfillcolor{currentfill}%
\pgfsetlinewidth{0.803000pt}%
\definecolor{currentstroke}{rgb}{0.000000,0.000000,0.000000}%
\pgfsetstrokecolor{currentstroke}%
\pgfsetdash{}{0pt}%
\pgfsys@defobject{currentmarker}{\pgfqpoint{0.000000in}{-0.048611in}}{\pgfqpoint{0.000000in}{0.000000in}}{%
\pgfpathmoveto{\pgfqpoint{0.000000in}{0.000000in}}%
\pgfpathlineto{\pgfqpoint{0.000000in}{-0.048611in}}%
\pgfusepath{stroke,fill}%
}%
\begin{pgfscope}%
\pgfsys@transformshift{4.959665in}{1.668889in}%
\pgfsys@useobject{currentmarker}{}%
\end{pgfscope}%
\end{pgfscope}%
\begin{pgfscope}%
\definecolor{textcolor}{rgb}{0.000000,0.000000,0.000000}%
\pgfsetstrokecolor{textcolor}%
\pgfsetfillcolor{textcolor}%
\pgftext[x=4.959665in,y=1.571667in,,top]{\color{textcolor}\rmfamily\fontsize{16.000000}{19.200000}\selectfont 0.6}%
\end{pgfscope}%
\begin{pgfscope}%
\pgfsetbuttcap%
\pgfsetroundjoin%
\definecolor{currentfill}{rgb}{0.000000,0.000000,0.000000}%
\pgfsetfillcolor{currentfill}%
\pgfsetlinewidth{0.803000pt}%
\definecolor{currentstroke}{rgb}{0.000000,0.000000,0.000000}%
\pgfsetstrokecolor{currentstroke}%
\pgfsetdash{}{0pt}%
\pgfsys@defobject{currentmarker}{\pgfqpoint{0.000000in}{-0.048611in}}{\pgfqpoint{0.000000in}{0.000000in}}{%
\pgfpathmoveto{\pgfqpoint{0.000000in}{0.000000in}}%
\pgfpathlineto{\pgfqpoint{0.000000in}{-0.048611in}}%
\pgfusepath{stroke,fill}%
}%
\begin{pgfscope}%
\pgfsys@transformshift{5.581962in}{1.668889in}%
\pgfsys@useobject{currentmarker}{}%
\end{pgfscope}%
\end{pgfscope}%
\begin{pgfscope}%
\definecolor{textcolor}{rgb}{0.000000,0.000000,0.000000}%
\pgfsetstrokecolor{textcolor}%
\pgfsetfillcolor{textcolor}%
\pgftext[x=5.581962in,y=1.571667in,,top]{\color{textcolor}\rmfamily\fontsize{16.000000}{19.200000}\selectfont 0.7}%
\end{pgfscope}%
\begin{pgfscope}%
\pgfsetbuttcap%
\pgfsetroundjoin%
\definecolor{currentfill}{rgb}{0.000000,0.000000,0.000000}%
\pgfsetfillcolor{currentfill}%
\pgfsetlinewidth{0.803000pt}%
\definecolor{currentstroke}{rgb}{0.000000,0.000000,0.000000}%
\pgfsetstrokecolor{currentstroke}%
\pgfsetdash{}{0pt}%
\pgfsys@defobject{currentmarker}{\pgfqpoint{0.000000in}{-0.048611in}}{\pgfqpoint{0.000000in}{0.000000in}}{%
\pgfpathmoveto{\pgfqpoint{0.000000in}{0.000000in}}%
\pgfpathlineto{\pgfqpoint{0.000000in}{-0.048611in}}%
\pgfusepath{stroke,fill}%
}%
\begin{pgfscope}%
\pgfsys@transformshift{6.204258in}{1.668889in}%
\pgfsys@useobject{currentmarker}{}%
\end{pgfscope}%
\end{pgfscope}%
\begin{pgfscope}%
\definecolor{textcolor}{rgb}{0.000000,0.000000,0.000000}%
\pgfsetstrokecolor{textcolor}%
\pgfsetfillcolor{textcolor}%
\pgftext[x=6.204258in,y=1.571667in,,top]{\color{textcolor}\rmfamily\fontsize{16.000000}{19.200000}\selectfont 0.8}%
\end{pgfscope}%
\begin{pgfscope}%
\pgfsetbuttcap%
\pgfsetroundjoin%
\definecolor{currentfill}{rgb}{0.000000,0.000000,0.000000}%
\pgfsetfillcolor{currentfill}%
\pgfsetlinewidth{0.803000pt}%
\definecolor{currentstroke}{rgb}{0.000000,0.000000,0.000000}%
\pgfsetstrokecolor{currentstroke}%
\pgfsetdash{}{0pt}%
\pgfsys@defobject{currentmarker}{\pgfqpoint{0.000000in}{-0.048611in}}{\pgfqpoint{0.000000in}{0.000000in}}{%
\pgfpathmoveto{\pgfqpoint{0.000000in}{0.000000in}}%
\pgfpathlineto{\pgfqpoint{0.000000in}{-0.048611in}}%
\pgfusepath{stroke,fill}%
}%
\begin{pgfscope}%
\pgfsys@transformshift{6.826555in}{1.668889in}%
\pgfsys@useobject{currentmarker}{}%
\end{pgfscope}%
\end{pgfscope}%
\begin{pgfscope}%
\definecolor{textcolor}{rgb}{0.000000,0.000000,0.000000}%
\pgfsetstrokecolor{textcolor}%
\pgfsetfillcolor{textcolor}%
\pgftext[x=6.826555in,y=1.571667in,,top]{\color{textcolor}\rmfamily\fontsize{16.000000}{19.200000}\selectfont 0.9}%
\end{pgfscope}%
\begin{pgfscope}%
\pgfsetbuttcap%
\pgfsetroundjoin%
\definecolor{currentfill}{rgb}{0.000000,0.000000,0.000000}%
\pgfsetfillcolor{currentfill}%
\pgfsetlinewidth{0.803000pt}%
\definecolor{currentstroke}{rgb}{0.000000,0.000000,0.000000}%
\pgfsetstrokecolor{currentstroke}%
\pgfsetdash{}{0pt}%
\pgfsys@defobject{currentmarker}{\pgfqpoint{0.000000in}{-0.048611in}}{\pgfqpoint{0.000000in}{0.000000in}}{%
\pgfpathmoveto{\pgfqpoint{0.000000in}{0.000000in}}%
\pgfpathlineto{\pgfqpoint{0.000000in}{-0.048611in}}%
\pgfusepath{stroke,fill}%
}%
\begin{pgfscope}%
\pgfsys@transformshift{7.448852in}{1.668889in}%
\pgfsys@useobject{currentmarker}{}%
\end{pgfscope}%
\end{pgfscope}%
\begin{pgfscope}%
\definecolor{textcolor}{rgb}{0.000000,0.000000,0.000000}%
\pgfsetstrokecolor{textcolor}%
\pgfsetfillcolor{textcolor}%
\pgftext[x=7.448852in,y=1.571667in,,top]{\color{textcolor}\rmfamily\fontsize{16.000000}{19.200000}\selectfont 1.0}%
\end{pgfscope}%
\begin{pgfscope}%
\definecolor{textcolor}{rgb}{0.000000,0.000000,0.000000}%
\pgfsetstrokecolor{textcolor}%
\pgfsetfillcolor{textcolor}%
\pgftext[x=4.337368in,y=1.302762in,,top]{\color{textcolor}\rmfamily\fontsize{16.000000}{19.200000}\selectfont \(\displaystyle \lambda\)}%
\end{pgfscope}%
\begin{pgfscope}%
\pgfsetbuttcap%
\pgfsetroundjoin%
\definecolor{currentfill}{rgb}{0.000000,0.000000,0.000000}%
\pgfsetfillcolor{currentfill}%
\pgfsetlinewidth{0.803000pt}%
\definecolor{currentstroke}{rgb}{0.000000,0.000000,0.000000}%
\pgfsetstrokecolor{currentstroke}%
\pgfsetdash{}{0pt}%
\pgfsys@defobject{currentmarker}{\pgfqpoint{-0.048611in}{0.000000in}}{\pgfqpoint{-0.000000in}{0.000000in}}{%
\pgfpathmoveto{\pgfqpoint{-0.000000in}{0.000000in}}%
\pgfpathlineto{\pgfqpoint{-0.048611in}{0.000000in}}%
\pgfusepath{stroke,fill}%
}%
\begin{pgfscope}%
\pgfsys@transformshift{0.914736in}{1.668889in}%
\pgfsys@useobject{currentmarker}{}%
\end{pgfscope}%
\end{pgfscope}%
\begin{pgfscope}%
\definecolor{textcolor}{rgb}{0.000000,0.000000,0.000000}%
\pgfsetstrokecolor{textcolor}%
\pgfsetfillcolor{textcolor}%
\pgftext[x=0.707446in, y=1.585556in, left, base]{\color{textcolor}\rmfamily\fontsize{16.000000}{19.200000}\selectfont \(\displaystyle {0}\)}%
\end{pgfscope}%
\begin{pgfscope}%
\pgfsetbuttcap%
\pgfsetroundjoin%
\definecolor{currentfill}{rgb}{0.000000,0.000000,0.000000}%
\pgfsetfillcolor{currentfill}%
\pgfsetlinewidth{0.803000pt}%
\definecolor{currentstroke}{rgb}{0.000000,0.000000,0.000000}%
\pgfsetstrokecolor{currentstroke}%
\pgfsetdash{}{0pt}%
\pgfsys@defobject{currentmarker}{\pgfqpoint{-0.048611in}{0.000000in}}{\pgfqpoint{-0.000000in}{0.000000in}}{%
\pgfpathmoveto{\pgfqpoint{-0.000000in}{0.000000in}}%
\pgfpathlineto{\pgfqpoint{-0.048611in}{0.000000in}}%
\pgfusepath{stroke,fill}%
}%
\begin{pgfscope}%
\pgfsys@transformshift{0.914736in}{2.452958in}%
\pgfsys@useobject{currentmarker}{}%
\end{pgfscope}%
\end{pgfscope}%
\begin{pgfscope}%
\definecolor{textcolor}{rgb}{0.000000,0.000000,0.000000}%
\pgfsetstrokecolor{textcolor}%
\pgfsetfillcolor{textcolor}%
\pgftext[x=0.487310in, y=2.369624in, left, base]{\color{textcolor}\rmfamily\fontsize{16.000000}{19.200000}\selectfont \(\displaystyle {100}\)}%
\end{pgfscope}%
\begin{pgfscope}%
\pgfsetbuttcap%
\pgfsetroundjoin%
\definecolor{currentfill}{rgb}{0.000000,0.000000,0.000000}%
\pgfsetfillcolor{currentfill}%
\pgfsetlinewidth{0.803000pt}%
\definecolor{currentstroke}{rgb}{0.000000,0.000000,0.000000}%
\pgfsetstrokecolor{currentstroke}%
\pgfsetdash{}{0pt}%
\pgfsys@defobject{currentmarker}{\pgfqpoint{-0.048611in}{0.000000in}}{\pgfqpoint{-0.000000in}{0.000000in}}{%
\pgfpathmoveto{\pgfqpoint{-0.000000in}{0.000000in}}%
\pgfpathlineto{\pgfqpoint{-0.048611in}{0.000000in}}%
\pgfusepath{stroke,fill}%
}%
\begin{pgfscope}%
\pgfsys@transformshift{0.914736in}{3.237026in}%
\pgfsys@useobject{currentmarker}{}%
\end{pgfscope}%
\end{pgfscope}%
\begin{pgfscope}%
\definecolor{textcolor}{rgb}{0.000000,0.000000,0.000000}%
\pgfsetstrokecolor{textcolor}%
\pgfsetfillcolor{textcolor}%
\pgftext[x=0.487310in, y=3.153693in, left, base]{\color{textcolor}\rmfamily\fontsize{16.000000}{19.200000}\selectfont \(\displaystyle {200}\)}%
\end{pgfscope}%
\begin{pgfscope}%
\definecolor{textcolor}{rgb}{0.000000,0.000000,0.000000}%
\pgfsetstrokecolor{textcolor}%
\pgfsetfillcolor{textcolor}%
\pgftext[x=0.431754in,y=2.714444in,,bottom,rotate=90.000000]{\color{textcolor}\rmfamily\fontsize{16.000000}{19.200000}\selectfont number of splits}%
\end{pgfscope}%
\begin{pgfscope}%
\pgfsetrectcap%
\pgfsetmiterjoin%
\pgfsetlinewidth{0.803000pt}%
\definecolor{currentstroke}{rgb}{0.000000,0.000000,0.000000}%
\pgfsetstrokecolor{currentstroke}%
\pgfsetdash{}{0pt}%
\pgfpathmoveto{\pgfqpoint{0.914736in}{1.668889in}}%
\pgfpathlineto{\pgfqpoint{0.914736in}{3.760000in}}%
\pgfusepath{stroke}%
\end{pgfscope}%
\begin{pgfscope}%
\pgfsetrectcap%
\pgfsetmiterjoin%
\pgfsetlinewidth{0.803000pt}%
\definecolor{currentstroke}{rgb}{0.000000,0.000000,0.000000}%
\pgfsetstrokecolor{currentstroke}%
\pgfsetdash{}{0pt}%
\pgfpathmoveto{\pgfqpoint{7.760000in}{1.668889in}}%
\pgfpathlineto{\pgfqpoint{7.760000in}{3.760000in}}%
\pgfusepath{stroke}%
\end{pgfscope}%
\begin{pgfscope}%
\pgfsetrectcap%
\pgfsetmiterjoin%
\pgfsetlinewidth{0.803000pt}%
\definecolor{currentstroke}{rgb}{0.000000,0.000000,0.000000}%
\pgfsetstrokecolor{currentstroke}%
\pgfsetdash{}{0pt}%
\pgfpathmoveto{\pgfqpoint{0.914736in}{1.668889in}}%
\pgfpathlineto{\pgfqpoint{7.760000in}{1.668889in}}%
\pgfusepath{stroke}%
\end{pgfscope}%
\begin{pgfscope}%
\pgfsetrectcap%
\pgfsetmiterjoin%
\pgfsetlinewidth{0.803000pt}%
\definecolor{currentstroke}{rgb}{0.000000,0.000000,0.000000}%
\pgfsetstrokecolor{currentstroke}%
\pgfsetdash{}{0pt}%
\pgfpathmoveto{\pgfqpoint{0.914736in}{3.760000in}}%
\pgfpathlineto{\pgfqpoint{7.760000in}{3.760000in}}%
\pgfusepath{stroke}%
\end{pgfscope}%
\begin{pgfscope}%
\pgfsetbuttcap%
\pgfsetmiterjoin%
\definecolor{currentfill}{rgb}{1.000000,1.000000,1.000000}%
\pgfsetfillcolor{currentfill}%
\pgfsetfillopacity{0.800000}%
\pgfsetlinewidth{1.003750pt}%
\definecolor{currentstroke}{rgb}{0.800000,0.800000,0.800000}%
\pgfsetstrokecolor{currentstroke}%
\pgfsetstrokeopacity{0.800000}%
\pgfsetdash{}{0pt}%
\pgfpathmoveto{\pgfqpoint{1.735082in}{0.734444in}}%
\pgfpathlineto{\pgfqpoint{6.939654in}{0.734444in}}%
\pgfpathquadraticcurveto{\pgfqpoint{6.984099in}{0.734444in}}{\pgfqpoint{6.984099in}{0.778889in}}%
\pgfpathlineto{\pgfqpoint{6.984099in}{1.081126in}}%
\pgfpathquadraticcurveto{\pgfqpoint{6.984099in}{1.125571in}}{\pgfqpoint{6.939654in}{1.125571in}}%
\pgfpathlineto{\pgfqpoint{1.735082in}{1.125571in}}%
\pgfpathquadraticcurveto{\pgfqpoint{1.690638in}{1.125571in}}{\pgfqpoint{1.690638in}{1.081126in}}%
\pgfpathlineto{\pgfqpoint{1.690638in}{0.778889in}}%
\pgfpathquadraticcurveto{\pgfqpoint{1.690638in}{0.734444in}}{\pgfqpoint{1.735082in}{0.734444in}}%
\pgfpathclose%
\pgfusepath{stroke,fill}%
\end{pgfscope}%
\begin{pgfscope}%
\pgfsetbuttcap%
\pgfsetmiterjoin%
\definecolor{currentfill}{rgb}{0.121569,0.466667,0.705882}%
\pgfsetfillcolor{currentfill}%
\pgfsetlinewidth{0.000000pt}%
\definecolor{currentstroke}{rgb}{0.000000,0.000000,0.000000}%
\pgfsetstrokecolor{currentstroke}%
\pgfsetstrokeopacity{0.000000}%
\pgfsetdash{}{0pt}%
\pgfpathmoveto{\pgfqpoint{1.779527in}{0.870015in}}%
\pgfpathlineto{\pgfqpoint{2.223971in}{0.870015in}}%
\pgfpathlineto{\pgfqpoint{2.223971in}{1.025571in}}%
\pgfpathlineto{\pgfqpoint{1.779527in}{1.025571in}}%
\pgfpathclose%
\pgfusepath{fill}%
\end{pgfscope}%
\begin{pgfscope}%
\definecolor{textcolor}{rgb}{0.000000,0.000000,0.000000}%
\pgfsetstrokecolor{textcolor}%
\pgfsetfillcolor{textcolor}%
\pgftext[x=2.401749in,y=0.870015in,left,base]{\color{textcolor}\rmfamily\fontsize{16.000000}{19.200000}\selectfont relational attribute}%
\end{pgfscope}%
\begin{pgfscope}%
\pgfsetbuttcap%
\pgfsetmiterjoin%
\definecolor{currentfill}{rgb}{1.000000,0.498039,0.054902}%
\pgfsetfillcolor{currentfill}%
\pgfsetlinewidth{0.000000pt}%
\definecolor{currentstroke}{rgb}{0.000000,0.000000,0.000000}%
\pgfsetstrokecolor{currentstroke}%
\pgfsetstrokeopacity{0.000000}%
\pgfsetdash{}{0pt}%
\pgfpathmoveto{\pgfqpoint{4.671404in}{0.870015in}}%
\pgfpathlineto{\pgfqpoint{5.115849in}{0.870015in}}%
\pgfpathlineto{\pgfqpoint{5.115849in}{1.025571in}}%
\pgfpathlineto{\pgfqpoint{4.671404in}{1.025571in}}%
\pgfpathclose%
\pgfusepath{fill}%
\end{pgfscope}%
\begin{pgfscope}%
\definecolor{textcolor}{rgb}{0.000000,0.000000,0.000000}%
\pgfsetstrokecolor{textcolor}%
\pgfsetfillcolor{textcolor}%
\pgftext[x=5.293626in,y=0.870015in,left,base]{\color{textcolor}\rmfamily\fontsize{16.000000}{19.200000}\selectfont textual attribute}%
\end{pgfscope}%
\end{pgfpicture}%
\makeatother%
\endgroup%

%% file: main-frame.bbl

\begin{thebibliography}{63}


\ifx \showCODEN    \undefined \def \showCODEN     #1{\unskip}     \fi
\ifx \showDOI      \undefined \def \showDOI       #1{#1}\fi
\ifx \showISBNx    \undefined \def \showISBNx     #1{\unskip}     \fi
\ifx \showISBNxiii \undefined \def \showISBNxiii  #1{\unskip}     \fi
\ifx \showISSN     \undefined \def \showISSN      #1{\unskip}     \fi
\ifx \showLCCN     \undefined \def \showLCCN      #1{\unskip}     \fi
\ifx \shownote     \undefined \def \shownote      #1{#1}          \fi
\ifx \showarticletitle \undefined \def \showarticletitle #1{#1}   \fi
\ifx \showURL      \undefined \def \showURL       {\relax}        \fi
\providecommand\bibfield[2]{#2}
\providecommand\bibinfo[2]{#2}
\providecommand\natexlab[1]{#1}
\providecommand\showeprint[2][]{arXiv:#2}

\bibitem[\protect\citeauthoryear{Abu-Khzam, Bazgan, Casel, and
  Fernau}{Abu-Khzam et~al\mbox{.}}{2018}]%
        {Abu-Khzam2018}
\bibfield{author}{\bibinfo{person}{Faisal~N. Abu-Khzam},
  \bibinfo{person}{Cristina Bazgan}, \bibinfo{person}{Katrin Casel}, {and}
  \bibinfo{person}{Henning Fernau}.} \bibinfo{year}{2018}\natexlab{}.
\newblock \showarticletitle{{Clustering with Lower-Bounded Sizes: A General
  Graph-Theoretic Framework}}.
\newblock \bibinfo{journal}{\emph{Algorithmica}} \bibinfo{volume}{80},
  \bibinfo{number}{9} (\bibinfo{year}{2018}), \bibinfo{pages}{2517--2550}.
\newblock
\showISSN{14320541}


\bibitem[\protect\citeauthoryear{Agrawal and Narayanan}{Agrawal and
  Narayanan}{2011}]%
        {Agrawal2011}
\bibfield{author}{\bibinfo{person}{Prachi Agrawal} {and} \bibinfo{person}{P.~J.
  Narayanan}.} \bibinfo{year}{2011}\natexlab{}.
\newblock \showarticletitle{{Person De-Identification in Videos}}.
\newblock \bibinfo{journal}{\emph{IEEE Trans. on Circuits and Systems for Video
  Technology}} \bibinfo{volume}{21}, \bibinfo{number}{3}
  (\bibinfo{year}{2011}), \bibinfo{pages}{299--310}.
\newblock
\showISSN{1051-8215}


\bibitem[\protect\citeauthoryear{Anandan, Clifton, Jiang, Murugesan,
  Pastrana-Camacho, and Si}{Anandan et~al\mbox{.}}{2012}]%
        {Anandan2012}
\bibfield{author}{\bibinfo{person}{Balamurugan Anandan}, \bibinfo{person}{Chris
  Clifton}, \bibinfo{person}{Wei Jiang}, \bibinfo{person}{Mummoorthy
  Murugesan}, \bibinfo{person}{Pedro Pastrana-Camacho}, {and}
  \bibinfo{person}{Luo Si}.} \bibinfo{year}{2012}\natexlab{}.
\newblock \showarticletitle{{t-Plausibility: Generalizing words to desensitize
  text}}.
\newblock \bibinfo{journal}{\emph{Trans. on Data Privacy}} \bibinfo{volume}{5},
  \bibinfo{number}{3} (\bibinfo{year}{2012}), \bibinfo{pages}{505--534}.
\newblock
\showISSN{18885063}


\bibitem[\protect\citeauthoryear{Bayardo and Agrawal}{Bayardo and
  Agrawal}{2005}]%
        {Bayardo2005}
\bibfield{author}{\bibinfo{person}{Roberto~J. Bayardo} {and}
  \bibinfo{person}{Rakesh Agrawal}.} \bibinfo{year}{2005}\natexlab{}.
\newblock \showarticletitle{{Data Privacy through Optimal k-Anonymization}}. In
  \bibinfo{booktitle}{\emph{ICDE}}. \bibinfo{publisher}{IEEE},
  \bibinfo{pages}{217--228}.
\newblock
\showISBNx{0-7695-2285-8}
\showISSN{10844627}


\bibitem[\protect\citeauthoryear{Chakaravarthy, Gupta, Roy, and
  Mohania}{Chakaravarthy et~al\mbox{.}}{2008}]%
        {Chakaravarthy2008}
\bibfield{author}{\bibinfo{person}{Venkatesan~T. Chakaravarthy},
  \bibinfo{person}{Himanshu Gupta}, \bibinfo{person}{Prasan Roy}, {and}
  \bibinfo{person}{Mukesh~K. Mohania}.} \bibinfo{year}{2008}\natexlab{}.
\newblock \showarticletitle{{Efficient techniques for document sanitization}}.
  In \bibinfo{booktitle}{\emph{Int. Conf. on Information and Knowledge Mining
  (CIKM)}}. \bibinfo{publisher}{ACM}, \bibinfo{pages}{843--852}.
\newblock
\showISBNx{9781595939913}


\bibitem[\protect\citeauthoryear{Cohn, Laish, Beryozkin, Li, Shafran, Szpektor,
  Hartman, Hassidim, and Matias}{Cohn et~al\mbox{.}}{2019}]%
        {Cohn2019}
\bibfield{author}{\bibinfo{person}{Ido Cohn}, \bibinfo{person}{Itay Laish},
  \bibinfo{person}{Genady Beryozkin}, \bibinfo{person}{Gang Li},
  \bibinfo{person}{Izhak Shafran}, \bibinfo{person}{Idan Szpektor},
  \bibinfo{person}{Tzvika Hartman}, \bibinfo{person}{Avinatan Hassidim}, {and}
  \bibinfo{person}{Yossi Matias}.} \bibinfo{year}{2019}\natexlab{}.
\newblock \showarticletitle{{Audio De-identification - a New Entity Recognition
  Task}}. In \bibinfo{booktitle}{\emph{Conf. of the North American Chapter of
  the Association for Computational Linguistics}}. \bibinfo{publisher}{ACL},
  \bibinfo{pages}{197--204}.
\newblock
\showISBNx{9781950737147}
\showeprint{1903.07037}


\bibitem[\protect\citeauthoryear{{Council of European Union}}{{Council of
  European Union}}{2016}]%
        {GDPR}
\bibfield{author}{\bibinfo{person}{{Council of European Union}}.}
  \bibinfo{year}{2016}\natexlab{}.
\newblock \bibinfo{title}{{EU General Data Protection Regulation (GDPR)}}.
\newblock
\newblock
\showISBNx{9781510866126}
\showISSN{2042-6852}
\urldef\tempurl%
\url{https://eur-lex.europa.eu/eli/reg/2016/679/oj}
\showURL{%
\tempurl}


\bibitem[\protect\citeauthoryear{Dernoncourt, Lee, Uzuner, and
  Szolovits}{Dernoncourt et~al\mbox{.}}{2017}]%
        {Dernoncourt2017}
\bibfield{author}{\bibinfo{person}{Franck Dernoncourt},
  \bibinfo{person}{Ji~Young Lee}, \bibinfo{person}{Ozlem Uzuner}, {and}
  \bibinfo{person}{Peter Szolovits}.} \bibinfo{year}{2017}\natexlab{}.
\newblock \showarticletitle{{De-identification of patient notes with recurrent
  neural networks}}.
\newblock \bibinfo{journal}{\emph{J. of the American Medical Informatics
  Association}} \bibinfo{volume}{24}, \bibinfo{number}{3}
  (\bibinfo{year}{2017}), \bibinfo{pages}{596--606}.
\newblock
\showISSN{1067-5027}
\showeprint{1606.03475}


\bibitem[\protect\citeauthoryear{Devlin, Chang, Lee, and Toutanova}{Devlin
  et~al\mbox{.}}{2019}]%
        {bert}
\bibfield{author}{\bibinfo{person}{Jacob Devlin}, \bibinfo{person}{Ming{-}Wei
  Chang}, \bibinfo{person}{Kenton Lee}, {and} \bibinfo{person}{Kristina
  Toutanova}.} \bibinfo{year}{2019}\natexlab{}.
\newblock \showarticletitle{{BERT:} Pre-training of Deep Bidirectional
  Transformers for Language Understanding}. In
  \bibinfo{booktitle}{\emph{NAACL-HLT}}. \bibinfo{publisher}{ACL},
  \bibinfo{pages}{4171--4186}.
\newblock


\bibitem[\protect\citeauthoryear{Dwork}{Dwork}{2006}]%
        {Dwork2006}
\bibfield{author}{\bibinfo{person}{Cynthia Dwork}.}
  \bibinfo{year}{2006}\natexlab{}.
\newblock \showarticletitle{{Differential Privacy}}. In
  \bibinfo{booktitle}{\emph{Automata, Languages and Programming}},
  Vol.~\bibinfo{volume}{4052}. \bibinfo{publisher}{Springer},
  \bibinfo{pages}{1--12}.
\newblock
\showISBNx{3540359079}
\showISSN{16113349}


\bibitem[\protect\citeauthoryear{Eder, Krieg-Holz, and Hahn}{Eder
  et~al\mbox{.}}{2019}]%
        {Eder2019}
\bibfield{author}{\bibinfo{person}{Elisabeth Eder}, \bibinfo{person}{Ulrike
  Krieg-Holz}, {and} \bibinfo{person}{Udo Hahn}.}
  \bibinfo{year}{2019}\natexlab{}.
\newblock \showarticletitle{{De-Identification of Emails: Pseudonymizing
  Privacy-Sensitive Data in a German Email Corpus}}. In
  \bibinfo{booktitle}{\emph{Int. Conf. on Recent Advances in Natural Language
  Processing}}. \bibinfo{publisher}{Incoma Ltd.}, \bibinfo{pages}{259--269}.
\newblock
\showISBNx{9789544520564}
\showISSN{13138502}


\bibitem[\protect\citeauthoryear{{El Emam}, Dankar, Issa, and {others}}{{El
  Emam} et~al\mbox{.}}{2009}]%
        {ElEmam2009}
\bibfield{author}{\bibinfo{person}{Khaled {El Emam}},
  \bibinfo{person}{Fida~Kamal Dankar}, \bibinfo{person}{Romeo Issa}, {and}
  \bibinfo{person}{{others}}.} \bibinfo{year}{2009}\natexlab{}.
\newblock \showarticletitle{{A Globally Optimal k-Anonymity Method for the
  De-Identification of Health Data}}.
\newblock \bibinfo{journal}{\emph{J. of the American Medical Informatics
  Association}} \bibinfo{volume}{16}, \bibinfo{number}{5}
  (\bibinfo{year}{2009}), \bibinfo{pages}{670--682}.
\newblock
\showISSN{1067-5027}


\bibitem[\protect\citeauthoryear{Fernandes, Dras, and McIver}{Fernandes
  et~al\mbox{.}}{2019}]%
        {Fernandes2018}
\bibfield{author}{\bibinfo{person}{Natasha Fernandes}, \bibinfo{person}{Mark
  Dras}, {and} \bibinfo{person}{Annabelle McIver}.}
  \bibinfo{year}{2019}\natexlab{}.
\newblock \showarticletitle{{Generalised Differential Privacy for Text Document
  Processing}}. In \bibinfo{booktitle}{\emph{Principles of Security and
  Trust}}. \bibinfo{publisher}{Springer}, \bibinfo{pages}{123--148}.
\newblock
\showISBNx{978-3-030-17137-7}
\showISSN{23318422}


\bibitem[\protect\citeauthoryear{Fung, Wang, Chen, and Yu}{Fung
  et~al\mbox{.}}{2010}]%
        {Fung2010}
\bibfield{author}{\bibinfo{person}{Benjamin C.~M. Fung}, \bibinfo{person}{Ke
  Wang}, \bibinfo{person}{Rui Chen}, {and} \bibinfo{person}{Philip~S. Yu}.}
  \bibinfo{year}{2010}\natexlab{}.
\newblock \showarticletitle{{Privacy-preserving data publishing: A survey of
  recent developments}}.
\newblock \bibinfo{journal}{\emph{Comput. Surveys}} \bibinfo{volume}{42},
  \bibinfo{number}{4} (\bibinfo{year}{2010}), \bibinfo{pages}{1--53}.
\newblock
\showISSN{0360-0300}


\bibitem[\protect\citeauthoryear{Gafni, Wolf, and Taigman}{Gafni
  et~al\mbox{.}}{2019}]%
        {Gafni2019}
\bibfield{author}{\bibinfo{person}{Oran Gafni}, \bibinfo{person}{Lior Wolf},
  {and} \bibinfo{person}{Yaniv Taigman}.} \bibinfo{year}{2019}\natexlab{}.
\newblock \showarticletitle{{Live Face De-Identification in Video}}. In
  \bibinfo{booktitle}{\emph{Int. Conf. on Computer Vision (ICCV)}}.
  \bibinfo{publisher}{IEEE}, \bibinfo{pages}{9377--9386}.
\newblock
\showISBNx{978-1-7281-4803-8}
\showISSN{15505499}


\bibitem[\protect\citeauthoryear{Gardner and Xiong}{Gardner and Xiong}{2008}]%
        {Gardner2008}
\bibfield{author}{\bibinfo{person}{James Gardner} {and} \bibinfo{person}{Li
  Xiong}.} \bibinfo{year}{2008}\natexlab{}.
\newblock \showarticletitle{{HIDE: An Integrated System for Health Information
  DE-identification}}. In \bibinfo{booktitle}{\emph{Int. Symposium on
  Computer-Based Medical Systems}}. \bibinfo{publisher}{IEEE},
  \bibinfo{pages}{254--259}.
\newblock
\showISBNx{978-0-7695-3165-6}
\showISSN{10637125}


\bibitem[\protect\citeauthoryear{Ghinita, Karras, Kalnis, and Mamoulis}{Ghinita
  et~al\mbox{.}}{2007}]%
        {Ghinita2007}
\bibfield{author}{\bibinfo{person}{Gabriel Ghinita},
  \bibinfo{person}{Panagiotis Karras}, \bibinfo{person}{Panos Kalnis}, {and}
  \bibinfo{person}{Nikos Mamoulis}.} \bibinfo{year}{2007}\natexlab{}.
\newblock \showarticletitle{{Fast data anonymization with low information
  loss}}. In \bibinfo{booktitle}{\emph{VLDB}}. \bibinfo{publisher}{ACM},
  \bibinfo{pages}{758--769}.
\newblock
\showISBNx{9781595936493}


\bibitem[\protect\citeauthoryear{Gkountouna and Terrovitis}{Gkountouna and
  Terrovitis}{2015}]%
        {DBLP:journals/tkde/GkountounaT15}
\bibfield{author}{\bibinfo{person}{Olga Gkountouna} {and}
  \bibinfo{person}{Manolis Terrovitis}.} \bibinfo{year}{2015}\natexlab{}.
\newblock \showarticletitle{Anonymizing Collections of Tree-Structured Data}.
\newblock \bibinfo{journal}{\emph{Trans. Knowl. Data Eng.}}
  \bibinfo{volume}{27}, \bibinfo{number}{8} (\bibinfo{year}{2015}),
  \bibinfo{pages}{2034--2048}.
\newblock


\bibitem[\protect\citeauthoryear{Gong, Luo, Yang, Ni, and Li}{Gong
  et~al\mbox{.}}{2017}]%
        {Gong2017}
\bibfield{author}{\bibinfo{person}{Qiyuan Gong}, \bibinfo{person}{Junzhou Luo},
  \bibinfo{person}{Ming Yang}, \bibinfo{person}{Weiwei Ni}, {and}
  \bibinfo{person}{Xiao~Bai Li}.} \bibinfo{year}{2017}\natexlab{}.
\newblock \showarticletitle{{Anonymizing 1:M microdata with high utility}}.
\newblock \bibinfo{journal}{\emph{Knowledge-Based Systems}}
  \bibinfo{volume}{115} (\bibinfo{year}{2017}), \bibinfo{pages}{15--26}.
\newblock
\showISSN{09507051}


\bibitem[\protect\citeauthoryear{Gross, Sweeney, de~la Torre, and Baker}{Gross
  et~al\mbox{.}}{2006}]%
        {Gross2006}
\bibfield{author}{\bibinfo{person}{Ralph Gross}, \bibinfo{person}{Latanya
  Sweeney}, \bibinfo{person}{F. de~la Torre}, {and} \bibinfo{person}{Simon
  Baker}.} \bibinfo{year}{2006}\natexlab{}.
\newblock \showarticletitle{{Model-Based Face De-Identification}}. In
  \bibinfo{booktitle}{\emph{Computer Vision and Pattern Recognition}}.
  \bibinfo{publisher}{IEEE}, \bibinfo{pages}{161--161}.
\newblock
\showISBNx{0-7695-2646-2}


\bibitem[\protect\citeauthoryear{Hassanzadeh, Lim, Kementsietsidis, and
  Wang}{Hassanzadeh et~al\mbox{.}}{2009}]%
        {Hassanzadeh2009}
\bibfield{author}{\bibinfo{person}{Oktie Hassanzadeh}, \bibinfo{person}{Lipyeow
  Lim}, \bibinfo{person}{Anastasios Kementsietsidis}, {and}
  \bibinfo{person}{Min Wang}.} \bibinfo{year}{2009}\natexlab{}.
\newblock \showarticletitle{{A declarative framework for semantic link
  discovery over relational data}}. In \bibinfo{booktitle}{\emph{Int. World
  Wide Web Conference}}. \bibinfo{publisher}{ACM}, \bibinfo{pages}{1101--1102}.
\newblock
\showISBNx{9781605584874}


\bibitem[\protect\citeauthoryear{He and Naughton}{He and Naughton}{2009}]%
        {He2009}
\bibfield{author}{\bibinfo{person}{Yeye He} {and} \bibinfo{person}{Jeffrey~F.
  Naughton}.} \bibinfo{year}{2009}\natexlab{}.
\newblock \showarticletitle{{Anonymization of Set-Valued Data via Top-Down,
  Local Generalization}}.
\newblock \bibinfo{journal}{\emph{VLDB}} \bibinfo{volume}{2},
  \bibinfo{number}{1} (\bibinfo{year}{2009}), \bibinfo{pages}{934--945}.
\newblock
\showISBNx{9781605589480}
\showISSN{2150-8097}


\bibitem[\protect\citeauthoryear{Hukkel{\aa}s, Mester, and
  Lindseth}{Hukkel{\aa}s et~al\mbox{.}}{2019}]%
        {Embedding2019}
\bibfield{author}{\bibinfo{person}{H{\aa}kon Hukkel{\aa}s},
  \bibinfo{person}{Rudolf Mester}, {and} \bibinfo{person}{Frank Lindseth}.}
  \bibinfo{year}{2019}\natexlab{}.
\newblock \showarticletitle{{DeepPrivacy: A Generative Adversarial Network for
  Face Anonymization}}. In \bibinfo{booktitle}{\emph{Advances in Visual
  Computing}}, Vol.~\bibinfo{volume}{11844}. \bibinfo{publisher}{Springer},
  \bibinfo{pages}{565--578}.
\newblock
\showISBNx{978-3-030-33719-3}


\bibitem[\protect\citeauthoryear{{Information and Privacy Commissioner of
  Ontario}}{{Information and Privacy Commissioner of Ontario}}{2016}]%
        {InformationandPrivacyCommissionerofOntario2016}
\bibfield{author}{\bibinfo{person}{{Information and Privacy Commissioner of
  Ontario}}.} \bibinfo{year}{2016}\natexlab{}.
\newblock \bibinfo{booktitle}{\emph{{De-identification Guidelines for
  Structured Data}}}.
\newblock \bibinfo{type}{{T}echnical {R}eport} June.
  \bibinfo{institution}{Information and Privacy Commissioner of Ontario}.
  \bibinfo{pages}{1--28} pages.
\newblock
\urldef\tempurl%
\url{https://www.ipc.on.ca/wp-content/uploads/2016/08/Deidentification-Guidelines-for-Structured-Data.pdf}
\showURL{%
\tempurl}


\bibitem[\protect\citeauthoryear{Johnson, Bulgarelli, and Pollard}{Johnson
  et~al\mbox{.}}{2020}]%
        {Johnson2020}
\bibfield{author}{\bibinfo{person}{Alistair E.~W. Johnson},
  \bibinfo{person}{Lucas Bulgarelli}, {and} \bibinfo{person}{Tom~J. Pollard}.}
  \bibinfo{year}{2020}\natexlab{}.
\newblock \showarticletitle{{Deidentification of free-text medical records
  using pre-trained bidirectional transformers}}. In
  \bibinfo{booktitle}{\emph{ACM Conf. on Health, Inference, and Learning}}.
  \bibinfo{publisher}{ACM}, \bibinfo{pages}{214--221}.
\newblock
\showISBNx{9781450370462}


\bibitem[\protect\citeauthoryear{Justin, Struc, Dobrisek, Vesnicer, Ipsic, and
  Mihelic}{Justin et~al\mbox{.}}{2015}]%
        {Justin2015}
\bibfield{author}{\bibinfo{person}{Tadej Justin}, \bibinfo{person}{Vitomir
  Struc}, \bibinfo{person}{Simon Dobrisek}, \bibinfo{person}{Bostjan Vesnicer},
  \bibinfo{person}{Ivo Ipsic}, {and} \bibinfo{person}{France Mihelic}.}
  \bibinfo{year}{2015}\natexlab{}.
\newblock \showarticletitle{{Speaker de-identification using diphone
  recognition and speech synthesis}}. In \bibinfo{booktitle}{\emph{11th IEEE
  Int. Conf. and Workshops on Automatic Face and Gesture Recognition}}.
  \bibinfo{publisher}{IEEE}, \bibinfo{pages}{1--7}.
\newblock
\showISBNx{978-1-4799-6026-2}


\bibitem[\protect\citeauthoryear{Kayaalp, Browne, Dodd, Sagan, and
  McDonald}{Kayaalp et~al\mbox{.}}{2014}]%
        {Kayaalp2014}
\bibfield{author}{\bibinfo{person}{Mehmet Kayaalp}, \bibinfo{person}{Allen~C.
  Browne}, \bibinfo{person}{Zeyno~A. Dodd}, \bibinfo{person}{Pamela Sagan},
  {and} \bibinfo{person}{Clement~J. McDonald}.}
  \bibinfo{year}{2014}\natexlab{}.
\newblock \showarticletitle{{De-identification of Address, Date, and
  Alphanumeric Identifiers in Narrative Clinical Reports}}. In
  \bibinfo{booktitle}{\emph{American Medical Informatics Association Annual
  Symposium}}. \bibinfo{publisher}{AMIA}, \bibinfo{pages}{767--776}.
\newblock
\showISSN{1942-597X}


\bibitem[\protect\citeauthoryear{Khan, Ziyadi, and AbdelHady}{Khan
  et~al\mbox{.}}{2020}]%
        {Khan2020}
\bibfield{author}{\bibinfo{person}{Muhammad~Raza Khan},
  \bibinfo{person}{Morteza Ziyadi}, {and} \bibinfo{person}{Mohamed AbdelHady}.}
  \bibinfo{year}{2020}\natexlab{}.
\newblock \bibinfo{title}{{MT-BioNER: Multi-task Learning for Biomedical Named
  Entity Recognition using Deep Bidirectional Transformers}}.
  (\bibinfo{year}{2020}).
\newblock
\showeprint[arxiv]{2001.08904}


\bibitem[\protect\citeauthoryear{Koppel, Schler, Argamon, and Messeri}{Koppel
  et~al\mbox{.}}{2006}]%
        {Koppel2006}
\bibfield{author}{\bibinfo{person}{Moshe Koppel}, \bibinfo{person}{Jonathan
  Schler}, \bibinfo{person}{Shlomo Argamon}, {and} \bibinfo{person}{Eran
  Messeri}.} \bibinfo{year}{2006}\natexlab{}.
\newblock \showarticletitle{{Authorship attribution with thousands of candidate
  authors}}. In \bibinfo{booktitle}{\emph{Int. Conf. on Research and
  Development in Information Retrieval (SIGIR)}}, Vol.~\bibinfo{volume}{2006}.
  \bibinfo{publisher}{ACM}, \bibinfo{pages}{659--660}.
\newblock
\showISBNx{1595933697}


\bibitem[\protect\citeauthoryear{Lee, Huh, and McNiel}{Lee
  et~al\mbox{.}}{2008}]%
        {Lee2008}
\bibfield{author}{\bibinfo{person}{Sangno Lee}, \bibinfo{person}{Soon~Young
  Huh}, {and} \bibinfo{person}{Ronald~D. McNiel}.}
  \bibinfo{year}{2008}\natexlab{}.
\newblock \showarticletitle{{Automatic generation of concept hierarchies using
  WordNet}}.
\newblock \bibinfo{journal}{\emph{Expert Systems with Applications}}
  \bibinfo{volume}{35}, \bibinfo{number}{3} (\bibinfo{year}{2008}),
  \bibinfo{pages}{1132--1144}.
\newblock
\showISSN{09574174}


\bibitem[\protect\citeauthoryear{LeFevre, DeWitt, and Ramakrishnan}{LeFevre
  et~al\mbox{.}}{2005}]%
        {LeFevre2005}
\bibfield{author}{\bibinfo{person}{Kristen LeFevre}, \bibinfo{person}{David~J.
  DeWitt}, {and} \bibinfo{person}{Raghu Ramakrishnan}.}
  \bibinfo{year}{2005}\natexlab{}.
\newblock \showarticletitle{{Incognito: efficient full-domain K-anonymity}}. In
  \bibinfo{booktitle}{\emph{2005 ACM SIGMOD Int. Conf. on Management of data}}.
  \bibinfo{publisher}{ACM}, \bibinfo{pages}{49--60}.
\newblock
\showISBNx{1595930604}
\showISSN{07308078}


\bibitem[\protect\citeauthoryear{LeFevre, DeWitt, and Ramakrishnan}{LeFevre
  et~al\mbox{.}}{2006}]%
        {LeFevre2006}
\bibfield{author}{\bibinfo{person}{Kristen LeFevre}, \bibinfo{person}{David~J.
  DeWitt}, {and} \bibinfo{person}{Raghu Ramakrishnan}.}
  \bibinfo{year}{2006}\natexlab{}.
\newblock \showarticletitle{{Mondrian Multidimensional K-Anonymity}}. In
  \bibinfo{booktitle}{\emph{ICDE}}. \bibinfo{publisher}{IEEE},
  \bibinfo{pages}{25--25}.
\newblock
\showISBNx{0-7695-2570-9}
\showISSN{10844627}


\bibitem[\protect\citeauthoryear{Li, Li, and Venkatasubramanian}{Li
  et~al\mbox{.}}{2007}]%
        {Li2007}
\bibfield{author}{\bibinfo{person}{Ninghui Li}, \bibinfo{person}{Tiancheng Li},
  {and} \bibinfo{person}{Suresh Venkatasubramanian}.}
  \bibinfo{year}{2007}\natexlab{}.
\newblock \showarticletitle{{t-Closeness: Privacy Beyond k-Anonymity and
  l-Diversity}}. In \bibinfo{booktitle}{\emph{ICDE}}.
  \bibinfo{publisher}{IEEE}, \bibinfo{pages}{106--115}.
\newblock
\showISBNx{1-4244-0802-4}


\bibitem[\protect\citeauthoryear{Ling and Weld}{Ling and Weld}{2012}]%
        {Ling2012}
\bibfield{author}{\bibinfo{person}{Xiao Ling} {and} \bibinfo{person}{Daniel~S.
  Weld}.} \bibinfo{year}{2012}\natexlab{}.
\newblock \showarticletitle{{Fine-grained entity recognition}}.
\newblock \bibinfo{journal}{\emph{National Conf. on Artificial Intelligence}}
  \bibinfo{volume}{1} (\bibinfo{year}{2012}), \bibinfo{pages}{94--100}.
\newblock
\showISBNx{9781577355687}


\bibitem[\protect\citeauthoryear{Liu, Tang, Wang, and Chen}{Liu
  et~al\mbox{.}}{2017}]%
        {Liu2017}
\bibfield{author}{\bibinfo{person}{Zengjian Liu}, \bibinfo{person}{Buzhou
  Tang}, \bibinfo{person}{Xiaolong Wang}, {and} \bibinfo{person}{Qingcai
  Chen}.} \bibinfo{year}{2017}\natexlab{}.
\newblock \showarticletitle{{De-identification of clinical notes via recurrent
  neural network and conditional random field}}.
\newblock \bibinfo{journal}{\emph{J. of Biomedical Informatics}}
  \bibinfo{volume}{75} (\bibinfo{year}{2017}), \bibinfo{pages}{S34--S42}.
\newblock
\showISSN{15320464}


\bibitem[\protect\citeauthoryear{Machanavajjhala, Gehrke, Kifer, and
  Venkitasubramaniam}{Machanavajjhala et~al\mbox{.}}{2006}]%
        {Machanavajjhala2006}
\bibfield{author}{\bibinfo{person}{Ashwin Machanavajjhala},
  \bibinfo{person}{Johannes Gehrke}, \bibinfo{person}{Daniel Kifer}, {and}
  \bibinfo{person}{Muthuramakrishnan Venkitasubramaniam}.}
  \bibinfo{year}{2006}\natexlab{}.
\newblock \showarticletitle{{L-diversity: privacy beyond k-anonymity}}. In
  \bibinfo{booktitle}{\emph{ICDE}}. \bibinfo{publisher}{IEEE},
  \bibinfo{pages}{24--24}.
\newblock
\showISBNx{0-7695-2570-9}
\showISSN{10844627}


\bibitem[\protect\citeauthoryear{McCallister, Grance, and Scarfone}{McCallister
  et~al\mbox{.}}{2010}]%
        {McCallister2010}
\bibfield{author}{\bibinfo{person}{Erika McCallister}, \bibinfo{person}{Timothy
  Grance}, {and} \bibinfo{person}{Karen~A. Scarfone}.}
  \bibinfo{year}{2010}\natexlab{}.
\newblock \bibinfo{booktitle}{\emph{{Guide to protecting the confidentiality of
  Personally Identifiable Information (PII)}}}.
\newblock \bibinfo{type}{{T}echnical {R}eport}. \bibinfo{institution}{National
  Institute of Standards and Technology}. \bibinfo{pages}{1--59} pages.
\newblock
\showISBNx{9781437934885}
\urldef\tempurl%
\url{https://nvlpubs.nist.gov/nistpubs/Legacy/SP/nistspecialpublication800-122.pdf}
\showURL{%
\tempurl}


\bibitem[\protect\citeauthoryear{McDonald, Afroz, Caliskan, Stolerman, and
  Greenstadt}{McDonald et~al\mbox{.}}{2012}]%
        {McDonald2012}
\bibfield{author}{\bibinfo{person}{Andrew W.~E. McDonald},
  \bibinfo{person}{Sadia Afroz}, \bibinfo{person}{Aylin Caliskan},
  \bibinfo{person}{Ariel Stolerman}, {and} \bibinfo{person}{Rachel
  Greenstadt}.} \bibinfo{year}{2012}\natexlab{}.
\newblock \showarticletitle{{Use Fewer Instances of the Letter “i”: Toward
  Writing Style Anonymization}}. In \bibinfo{booktitle}{\emph{Privacy Enhancing
  Technologies}}, Vol.~\bibinfo{volume}{7384}. \bibinfo{publisher}{Springer},
  \bibinfo{pages}{299--318}.
\newblock
\showISBNx{9783642316791}
\showISSN{03029743}


\bibitem[\protect\citeauthoryear{Meyerson and Williams}{Meyerson and
  Williams}{2004}]%
        {Meyerson2004}
\bibfield{author}{\bibinfo{person}{Adam Meyerson} {and} \bibinfo{person}{Ryan
  Williams}.} \bibinfo{year}{2004}\natexlab{}.
\newblock \showarticletitle{{On the complexity of optimal k-anonymity}}. In
  \bibinfo{booktitle}{\emph{Symposium on Principles of Database Systems}}.
  \bibinfo{publisher}{ACM}, \bibinfo{pages}{223--228}.
\newblock


\bibitem[\protect\citeauthoryear{Mikolov, Sutskever, Chen, Corrado, and
  Dean}{Mikolov et~al\mbox{.}}{2013}]%
        {DBLP:conf/nips/MikolovSCCD13}
\bibfield{author}{\bibinfo{person}{Tom{\'{a}}s Mikolov}, \bibinfo{person}{Ilya
  Sutskever}, \bibinfo{person}{Kai Chen}, \bibinfo{person}{Gregory~S. Corrado},
  {and} \bibinfo{person}{Jeffrey Dean}.} \bibinfo{year}{2013}\natexlab{}.
\newblock \showarticletitle{Distributed Representations of Words and Phrases
  and their Compositionality}. In \bibinfo{booktitle}{\emph{NIPS}},
  \bibfield{editor}{\bibinfo{person}{Christopher J.~C. Burges},
  \bibinfo{person}{L{\'{e}}on Bottou}, \bibinfo{person}{Zoubin Ghahramani},
  {and} \bibinfo{person}{Kilian~Q. Weinberger}} (Eds.).
  \bibinfo{pages}{3111--3119}.
\newblock


\bibitem[\protect\citeauthoryear{Neamatullah, Douglass, Lehman, and
  {others}}{Neamatullah et~al\mbox{.}}{2008}]%
        {Neamatullah2008}
\bibfield{author}{\bibinfo{person}{Ishna Neamatullah},
  \bibinfo{person}{Margaret~M. Douglass}, \bibinfo{person}{Li-wei~H Lehman},
  {and} \bibinfo{person}{{others}}.} \bibinfo{year}{2008}\natexlab{}.
\newblock \showarticletitle{{Automated de-identification of free-text medical
  records}}.
\newblock \bibinfo{journal}{\emph{BMC Medical Informatics and Decision Making}}
   \bibinfo{volume}{8} (\bibinfo{year}{2008}), \bibinfo{pages}{32}.
\newblock
\showISBNx{1472694783}
\showISSN{1472-6947}


\bibitem[\protect\citeauthoryear{Nergiz, Clifton, and Nergiz}{Nergiz
  et~al\mbox{.}}{2007}]%
        {Nergiz2007}
\bibfield{author}{\bibinfo{person}{Mehmet~Ercan Nergiz},
  \bibinfo{person}{Christopher Clifton}, {and} \bibinfo{person}{Ahmet~Erhan
  Nergiz}.} \bibinfo{year}{2007}\natexlab{}.
\newblock \showarticletitle{{MultiRelational k-Anonymity}}. In
  \bibinfo{booktitle}{\emph{ICDE}}, Vol.~\bibinfo{volume}{21}.
  \bibinfo{publisher}{IEEE}, \bibinfo{pages}{1417--1421}.
\newblock
\showISBNx{1-4244-0802-4}


\bibitem[\protect\citeauthoryear{Poulis, Loukides, Gkoulalas-Divanis, and
  Skiadopoulos}{Poulis et~al\mbox{.}}{2013}]%
        {Poulis2013}
\bibfield{author}{\bibinfo{person}{Giorgos Poulis}, \bibinfo{person}{Grigorios
  Loukides}, \bibinfo{person}{Aris Gkoulalas-Divanis}, {and}
  \bibinfo{person}{Spiros Skiadopoulos}.} \bibinfo{year}{2013}\natexlab{}.
\newblock \showarticletitle{{Anonymizing Data with Relational and Transaction
  Attributes}}. In \bibinfo{booktitle}{\emph{Machine Learning and Knowledge
  Discovery in Databases}}. \bibinfo{publisher}{Springer},
  \bibinfo{pages}{353--369}.
\newblock
\showISBNx{9783642409936}
\showISSN{03029743}


\bibitem[\protect\citeauthoryear{Prasser, Eicher, Spengler, Bild, and
  Kuhn}{Prasser et~al\mbox{.}}{2020}]%
        {Prasser2020}
\bibfield{author}{\bibinfo{person}{Fabian Prasser}, \bibinfo{person}{Johanna
  Eicher}, \bibinfo{person}{Helmut Spengler}, \bibinfo{person}{Raffael Bild},
  {and} \bibinfo{person}{Klaus~A. Kuhn}.} \bibinfo{year}{2020}\natexlab{}.
\newblock \showarticletitle{{Flexible data anonymization using ARX—Current
  status and challenges ahead}}.
\newblock \bibinfo{journal}{\emph{Software: Practice and Experience}}
  \bibinfo{volume}{50}, \bibinfo{number}{7} (\bibinfo{year}{2020}),
  \bibinfo{pages}{1277--1304}.
\newblock
\showISSN{0038-0644}


\bibitem[\protect\citeauthoryear{Prasser, Kohlmayer, Lautenschl{\"{a}}ger, and
  Kuhn}{Prasser et~al\mbox{.}}{2014}]%
        {Prasser2014}
\bibfield{author}{\bibinfo{person}{Fabian Prasser}, \bibinfo{person}{Florian
  Kohlmayer}, \bibinfo{person}{Ronald Lautenschl{\"{a}}ger}, {and}
  \bibinfo{person}{Klaus~A. Kuhn}.} \bibinfo{year}{2014}\natexlab{}.
\newblock \showarticletitle{{ARX--A Comprehensive Tool for Anonymizing
  Biomedical Data}}. In \bibinfo{booktitle}{\emph{American Medical Informatics
  Association Annual Symposium}}. \bibinfo{publisher}{AMIA},
  \bibinfo{pages}{984--993}.
\newblock
\showISSN{1942-597X}


\bibitem[\protect\citeauthoryear{Ruch, Baud, Rassinoux, Bouillon, and
  Robert}{Ruch et~al\mbox{.}}{2000}]%
        {Ruch2000}
\bibfield{author}{\bibinfo{person}{Patrick Ruch}, \bibinfo{person}{Robert~H.
  Baud}, \bibinfo{person}{Anne-Marie Rassinoux}, \bibinfo{person}{Pierrette
  Bouillon}, {and} \bibinfo{person}{Gilbert Robert}.}
  \bibinfo{year}{2000}\natexlab{}.
\newblock \showarticletitle{{Medical document anonymization with a semantic
  lexicon}}. In \bibinfo{booktitle}{\emph{American Medical Informatics
  Association Annual Symposium}}. \bibinfo{publisher}{AMIA},
  \bibinfo{pages}{729--733}.
\newblock
\showISSN{1531-605X}


\bibitem[\protect\citeauthoryear{Samarati}{Samarati}{2001}]%
        {Samarati2001}
\bibfield{author}{\bibinfo{person}{Pierangela Samarati}.}
  \bibinfo{year}{2001}\natexlab{}.
\newblock \showarticletitle{{Protecting respondents identities in microdata
  release}}.
\newblock \bibinfo{journal}{\emph{IEEE Trans. on Knowledge and Data
  Engineering}} \bibinfo{volume}{13}, \bibinfo{number}{6}
  (\bibinfo{year}{2001}), \bibinfo{pages}{1010--1027}.
\newblock
\showISSN{10414347}


\bibitem[\protect\citeauthoryear{S{\'{a}}nchez, Batet, and Viejo}{S{\'{a}}nchez
  et~al\mbox{.}}{2013}]%
        {Sanchez2013}
\bibfield{author}{\bibinfo{person}{David S{\'{a}}nchez},
  \bibinfo{person}{Montserrat Batet}, {and} \bibinfo{person}{Alexandre Viejo}.}
  \bibinfo{year}{2013}\natexlab{}.
\newblock \showarticletitle{{Automatic general-purpose sanitization of textual
  documents}}.
\newblock \bibinfo{journal}{\emph{IEEE Trans. on Information Forensics and
  Security}} \bibinfo{volume}{8}, \bibinfo{number}{6} (\bibinfo{year}{2013}),
  \bibinfo{pages}{853--862}.
\newblock
\showISBNx{4300002010}
\showISSN{15566013}


\bibitem[\protect\citeauthoryear{Saygin, Hakkani-T{\"{u}}r, and
  T{\"{u}}r}{Saygin et~al\mbox{.}}{2009}]%
        {SayginHT09}
\bibfield{author}{\bibinfo{person}{Y{\"{u}}cel Saygin}, \bibinfo{person}{Dilek
  Hakkani-T{\"{u}}r}, {and} \bibinfo{person}{G{\"{o}}khan T{\"{u}}r}.}
  \bibinfo{year}{2009}\natexlab{}.
\newblock \showarticletitle{{Sanitization and Anonymization of Document
  Repositories}}.
\newblock In \bibinfo{booktitle}{\emph{Database Technologies: Concepts,
  Methodologies, Tools, and Applications}}. \bibinfo{publisher}{IGI Global},
  \bibinfo{pages}{2129--2139}.
\newblock


\bibitem[\protect\citeauthoryear{Schler, Koppel, Argamon, and
  Pennebaker}{Schler et~al\mbox{.}}{2006}]%
        {Schler2006}
\bibfield{author}{\bibinfo{person}{Jonathan Schler}, \bibinfo{person}{Moshe
  Koppel}, \bibinfo{person}{Shlomo Argamon}, {and} \bibinfo{person}{James
  Pennebaker}.} \bibinfo{year}{2006}\natexlab{}.
\newblock \showarticletitle{{Effects of age and gender on blogging}}. In
  \bibinfo{booktitle}{\emph{Computational Approaches to Analyzing Weblogs}}.
  \bibinfo{publisher}{AAAI}, \bibinfo{pages}{199--205}.
\newblock
\showISBNx{1577352645}


\bibitem[\protect\citeauthoryear{Sweeney}{Sweeney}{1996}]%
        {Sweeney1996}
\bibfield{author}{\bibinfo{person}{Latanya Sweeney}.}
  \bibinfo{year}{1996}\natexlab{}.
\newblock \showarticletitle{{Replacing personally-identifying information in
  medical records, the Scrub system}}. In \bibinfo{booktitle}{\emph{Proceedings
  : a Conf. of the American Medical Informatics Association. AMIA Fall
  Symposium}}. \bibinfo{publisher}{AMIA}, \bibinfo{pages}{333--337}.
\newblock
\showISSN{1091-8280}


\bibitem[\protect\citeauthoryear{Sweeney}{Sweeney}{2000}]%
        {Sweeney2000}
\bibfield{author}{\bibinfo{person}{Latanya Sweeney}.}
  \bibinfo{year}{2000}\natexlab{}.
\newblock \showarticletitle{Simple Demographics Often Identify People
  Uniquely}. In \bibinfo{booktitle}{\emph{Data Privacy Working Paper 3}}.
  \bibinfo{publisher}{Carnegie Mellon U.}
\newblock


\bibitem[\protect\citeauthoryear{Sweeney}{Sweeney}{2002a}]%
        {Sweeney2002}
\bibfield{author}{\bibinfo{person}{Latanya Sweeney}.}
  \bibinfo{year}{2002}\natexlab{a}.
\newblock \showarticletitle{{Achieving k-Anonymity Privacy Protection Using
  Generalization and Suppression}}.
\newblock \bibinfo{journal}{\emph{Int. J. of Uncertainty, Fuzziness and
  Knowledge-Based Systems}} \bibinfo{volume}{10}, \bibinfo{number}{5}
  (\bibinfo{year}{2002}), \bibinfo{pages}{571--588}.
\newblock
\showISSN{0218-4885}


\bibitem[\protect\citeauthoryear{Sweeney}{Sweeney}{2002b}]%
        {Sweeney2002k}
\bibfield{author}{\bibinfo{person}{Latanya Sweeney}.}
  \bibinfo{year}{2002}\natexlab{b}.
\newblock \showarticletitle{{k-Anonymity: A Model for Protecting Privacy}}.
\newblock \bibinfo{journal}{\emph{Int. J. of Uncertainty, Fuzziness and
  Knowledge-Based Systems}} \bibinfo{volume}{10}, \bibinfo{number}{5}
  (\bibinfo{year}{2002}), \bibinfo{pages}{557--570}.
\newblock
\showISSN{0218-4885}


\bibitem[\protect\citeauthoryear{Teinemaa, Dumas, Maggi, and {Di
  Francescomarino}}{Teinemaa et~al\mbox{.}}{2016}]%
        {Teinemaa2016}
\bibfield{author}{\bibinfo{person}{Irene Teinemaa}, \bibinfo{person}{Marlon
  Dumas}, \bibinfo{person}{Fabrizio~Maria Maggi}, {and} \bibinfo{person}{Chiara
  {Di Francescomarino}}.} \bibinfo{year}{2016}\natexlab{}.
\newblock \showarticletitle{{Predictive business process monitoring with
  structured and unstructured data}}. In \bibinfo{booktitle}{\emph{Business
  Process Management}}, Vol.~\bibinfo{volume}{9850}.
  \bibinfo{publisher}{Springer}, \bibinfo{pages}{401--417}.
\newblock
\showISSN{16113349}


\bibitem[\protect\citeauthoryear{Terrovitis, Mamoulis, and Kalnis}{Terrovitis
  et~al\mbox{.}}{2008}]%
        {Terrovitis2008}
\bibfield{author}{\bibinfo{person}{Manolis Terrovitis}, \bibinfo{person}{Nikos
  Mamoulis}, {and} \bibinfo{person}{Panos Kalnis}.}
  \bibinfo{year}{2008}\natexlab{}.
\newblock \showarticletitle{{Privacy-preserving anonymization of set-valued
  data}}.
\newblock \bibinfo{journal}{\emph{VLDB}} \bibinfo{volume}{1},
  \bibinfo{number}{1} (\bibinfo{year}{2008}), \bibinfo{pages}{115--125}.
\newblock
\showISBNx{0000000000000}
\showISSN{2150-8097}


\bibitem[\protect\citeauthoryear{{The Office for Civil Rights (OCR)} and
  Malin}{{The Office for Civil Rights (OCR)} and Malin}{2012}]%
        {HIPAA}
\bibfield{author}{\bibinfo{person}{{The Office for Civil Rights (OCR)}} {and}
  \bibinfo{person}{Bradley Malin}.} \bibinfo{year}{2012}\natexlab{}.
\newblock \bibinfo{booktitle}{\emph{{Guidance Regarding Methods for
  de-identification of protected health information in accordance with the
  Health Insurance Portability and Accountability Act (HIPAA) Privacy Rule}}}.
\newblock \bibinfo{type}{{T}echnical {R}eport}. \bibinfo{institution}{U.S.
  Department of Health {\&} Human Services}. \bibinfo{pages}{1--32} pages.
\newblock
\urldef\tempurl%
\url{https://www.hhs.gov/sites/default/files/ocr/privacy/hipaa/understanding/coveredentities/De-identification/hhs\_deid\_guidance.pdf}
\showURL{%
\tempurl}


\bibitem[\protect\citeauthoryear{Trienes, Trieschnigg, Seifert, and
  Hiemstra}{Trienes et~al\mbox{.}}{2020}]%
        {Trienes2020}
\bibfield{author}{\bibinfo{person}{Jan Trienes}, \bibinfo{person}{Dolf
  Trieschnigg}, \bibinfo{person}{Christin Seifert}, {and}
  \bibinfo{person}{Djoerd Hiemstra}.} \bibinfo{year}{2020}\natexlab{}.
\newblock \showarticletitle{{Comparing Rule-based, Feature-based and Deep
  Neural Methods for De-identification of Dutch Medical Records}}. In
  \bibinfo{booktitle}{\emph{Health Search and Data Mining}},
  Vol.~\bibinfo{volume}{2551}. \bibinfo{publisher}{CEUR},
  \bibinfo{pages}{3--11}.
\newblock
\showISSN{16130073}


\bibitem[\protect\citeauthoryear{Vaswani, Shazeer, Parmar, Uszkoreit, Jones,
  Gomez, Kaiser, and Polosukhin}{Vaswani et~al\mbox{.}}{2017}]%
        {Vaswani2017}
\bibfield{author}{\bibinfo{person}{Ashish Vaswani}, \bibinfo{person}{Noam
  Shazeer}, \bibinfo{person}{Niki Parmar}, \bibinfo{person}{Jakob Uszkoreit},
  \bibinfo{person}{Llion Jones}, \bibinfo{person}{Aidan~N. Gomez},
  \bibinfo{person}{Lukasz Kaiser}, {and} \bibinfo{person}{Illia Polosukhin}.}
  \bibinfo{year}{2017}\natexlab{}.
\newblock \showarticletitle{{Attention Is All You Need}}. In
  \bibinfo{booktitle}{\emph{Conf. on Neural Information Processing Systems}}.
  \bibinfo{publisher}{Curran}, \bibinfo{pages}{5998--6008}.
\newblock


\bibitem[\protect\citeauthoryear{Xu, Wang, Pei, Wang, Shi, and Fu}{Xu
  et~al\mbox{.}}{2006}]%
        {Xu2006}
\bibfield{author}{\bibinfo{person}{Jian Xu}, \bibinfo{person}{Wei Wang},
  \bibinfo{person}{Jian Pei}, \bibinfo{person}{Xiaoyuan Wang},
  \bibinfo{person}{Baile Shi}, {and} \bibinfo{person}{Ada~Wai{-}Chee Fu}.}
  \bibinfo{year}{2006}\natexlab{}.
\newblock \showarticletitle{{Utility-based anonymization using local
  recoding}}. In \bibinfo{booktitle}{\emph{SIGKDD}}. \bibinfo{publisher}{ACM},
  \bibinfo{pages}{785--790}.
\newblock
\showISBNx{1595933395}


\bibitem[\protect\citeauthoryear{Xu, Aggarwal, Feyisetan, and Teissier}{Xu
  et~al\mbox{.}}{2020}]%
        {Xu2020}
\bibfield{author}{\bibinfo{person}{Zekun Xu}, \bibinfo{person}{Abhinav
  Aggarwal}, \bibinfo{person}{Oluwaseyi Feyisetan}, {and}
  \bibinfo{person}{Nathanael Teissier}.} \bibinfo{year}{2020}\natexlab{}.
\newblock \bibinfo{title}{{A Differentially Private Text Perturbation Method
  Using Regularized Mahalanobis Metric}}.  (\bibinfo{year}{2020}),
  \bibinfo{numpages}{7--17}~pages.
\newblock
\showeprint[arxiv]{2010.11947}


\bibitem[\protect\citeauthoryear{Yan, Deng, Li, and Qiu}{Yan
  et~al\mbox{.}}{2019}]%
        {Yan2019}
\bibfield{author}{\bibinfo{person}{Hang Yan}, \bibinfo{person}{Bocao Deng},
  \bibinfo{person}{Xiaonan Li}, {and} \bibinfo{person}{Xipeng Qiu}.}
  \bibinfo{year}{2019}\natexlab{}.
\newblock \bibinfo{title}{{TENER: Adapting Transformer Encoder for Named Entity
  Recognition}}.  (\bibinfo{year}{2019}).
\newblock
\showeprint[arxiv]{1911.04474}


\bibitem[\protect\citeauthoryear{Zhao and Zhou}{Zhao and Zhou}{2020}]%
        {Zhao2020}
\bibfield{author}{\bibinfo{person}{Ying Zhao} {and} \bibinfo{person}{Charles~C.
  Zhou}.} \bibinfo{year}{2020}\natexlab{}.
\newblock \showarticletitle{{Link Analysis to Discover Insights from Structured
  and Unstructured Data on COVID-19}}. In
  \bibinfo{booktitle}{\emph{Bioinformatics, Computational Biology and Health
  Informatics}}. \bibinfo{publisher}{ACM}, \bibinfo{pages}{1--8}.
\newblock
\showISBNx{9781450379649}


\end{thebibliography}
